\documentclass{article}

\usepackage[utf8]{inputenc} % allow utf-8 input
\usepackage[T1]{fontenc}    % use 8-bit T1 fonts
\usepackage[urlcolor=blue]{hyperref}
\usepackage{url}            % simple URL typesetting
\usepackage{nicefrac}       % compact symbols for 1/2, etc.
\usepackage{microtype}
\usepackage{graphicx}
\usepackage{tabularx}
\usepackage{amsmath, amsfonts}
\usepackage{amssymb}
\usepackage{bbm}
\usepackage{wrapfig}
\usepackage{paralist}
\usepackage{multirow}
\usepackage[table,dvipsnames]{xcolor}
\usepackage{subfigure}
\usepackage[ruled,vlined]{algorithm2e}
\usepackage{algpseudocode}
\usepackage{booktabs} % for professional tables
\usepackage{adjustbox}
\usepackage{dashbox}
\usepackage{siunitx}
\usepackage{colortbl}
\usepackage{xspace}
\usepackage{xifthen}

\newtheorem{manualtheoreminner}{Theorem}
\newenvironment{manualtheorem}[1]{%
  \renewcommand\themanualtheoreminner{#1}%
  \manualtheoreminner
}{\endmanualtheoreminner}

\definecolor{light-gray}{gray}{0.9}

\hypersetup{
    colorlinks,
    linkcolor={red!60!black},
    citecolor={green!50!black},
    urlcolor={blue!80!black}
}

\newcommand{\name}{TAAC\xspace}

\newcommand{\expect}{\mathop{\mathbb{E}}}
\newcommand{\Min}{\mathop{\min}}
\newcommand{\Max}{\mathop{\max}}

\newcommand{\eg}{\textit{e.g.},}
\newcommand{\etc}{\textit{etc.}}
\newcommand{\ta}{\tilde{a}}
\newcommand{\tb}{\tilde{b}}

\newcommand{\thetaA}{\phi}
\newcommand{\thetaQ}{\theta}
\newcommand{\task}{\textit}
\newcommand{\vs}{\textit{vs.}}
\makeatletter
\DeclareRobustCommand
  \myvdots{\vbox{\baselineskip4\p@ \lineskiplimit\z@
    \hbox{.}\hbox{.}\hbox{.}}}
\makeatother

\DeclareMathOperator*{\argmax}{arg\,max}

\usepackage{pifont}% http://ctan.org/pkg/pifont
\newcommand{\cmark}{\ding{51}}%
\newcommand{\xmark}{\ding{55}}%

\usepackage[final]{neurips_2021}

\title{\name: Temporally Abstract Actor-Critic for Continuous Control}

\author{%
  Haonan Yu, \ Wei Xu, \ Haichao Zhang 
  \\
  Horizon Robotics\\
  Cupertino, CA 95014 \\
  \texttt{\{haonan.yu,wei.xu,haichao.zhang\}@horizon.ai} \\
}

\begin{document}

\maketitle

\begin{abstract}
We present temporally abstract actor-critic (\name), a simple but effective off-policy RL algorithm that incorporates 
closed-loop temporal abstraction into the actor-critic framework.
\name adds a second-stage binary policy to choose between the previous action and a new action output by an actor.
Crucially, its ``act-or-repeat'' decision hinges on the actually sampled action instead of the expected behavior of the actor.
This post-acting switching scheme let the overall policy make more informed decisions.
\name has two important features: a) persistent exploration, and b) a new compare-through Q operator for multi-step TD backup, specially tailored to the action repetition scenario.
We demonstrate \name's advantages over several strong baselines across 14 continuous control tasks.
Our surprising finding reveals that while achieving top performance, \name is able to ``mine'' a significant number of repeated actions with the trained policy 
even on continuous tasks whose problem structures on the surface seem to repel action repetition.
This suggests that aside from encouraging persistent exploration, action repetition can find its place in a good policy behavior.
%
%This is perhaps due to that the action frequency of a task can be difficult to be set exactly as the minimum value that doesn't comprise optimal control while leaving no room 
%for temporal abstraction.
%
%with its improvement of convergence speed and final performance over the top-performing baseline as ?? and ??, respectively.
%
\ifcsname usepreprint\endcsname
Code is available at \url{https://github.com/hnyu/taac}.
\else
Code will be made public.
\fi
\end{abstract}

\section{Introduction}
Deep reinforcement learning (RL) has achieved great success in various continuous action domains such as locomotion and manipulation \citep{schulman2017trust,Lillicrap2016,duan2016benchmarking,schulman2017proximal,fujimoto2018,Haarnoja2018}.
%p
Despite promising empirical results, these widely applicable continuous RL algorithms execute a newly computed action at every step of the finest time scale of a problem.
With no decision making at higher levels, they attempt to solve the challenging credit assignment problem over a long horizon.
As a result, considerable sample efficiency improvements have yet to be made by them in complex task structures \citep{Riedmiller2018,Li2020,Lee2020Learning} and extremely sparse reward settings \citep{andrychowicz2018hindsight,plappert2018multigoal,zhang2021hierarchical}.

On the other hand, it is argued that temporal abstraction \citep{Parr1998,Dietterich98themaxq,SUTTON1999,Precup2000temporalabstraction} is one of the crucial keys to solving 
control problems with complex structures.
%
%The action hierarchy formed by temporal abstraction could serve as a useful prior aligned with the complex task structure, or encourage exploration for discovering sparse rewards, as 
%
Larger steps are taken at higher levels of abstraction while lower-level actions only need to focus on solving isolated subtasks \citep{Dayan1993,Vezhnevets2017,bacon2017}.
However, most hierarchical RL (HRL) methods are task specific and nontrivial to adapt. 
For example, the options framework \citep{SUTTON1999,Precup2000temporalabstraction,bacon2017} requires pre-defining an option space,
while the feudal RL framework \cite{Vezhnevets2017,nachum2018dataefficient,zhang2021hierarchical} requires tuning the hyperparameters of dimensionality and domain 
range of the goal space.
In practice, their final performance usually hinges on these choices.
%
\iffalse
Moreover, learning sub-policies and their termination conditions from scratch could be expensive in terms of data and computation.
%
As a result, some HRL methods \citep{Riedmiller2018,Lee2020Learning} use pre-trained skills or sub-policies to facilitate training, 
assuming a known task structure.
%
This somewhat limits their generalization to new scenarios.
\fi

Perhaps the simplest form of an option or sub-policy would be just repeating an action for a certain number of steps, a straightforward 
idea that has been widely explored \citep{Lakshminarayanan17,sharma2017learning,dabney2020,metelli2020control,LLK2020,Biedenkapp2021}.
This line of works can be regarded as a middle ground between ``flat'' RL and HRL.
They assume a fixed candidate set of action durations, and repeat actions in an \emph{open-loop} manner.
Open-loop control forces an agent to commit to the same action over a predicted duration with no opportunity of early terminations.
It weakens the agent's ability of handling emergency situations and correcting wrong durations predicted earlier.
To address this inflexibility, a handful of prior works~\citep{Neunert2020,Chen2021} propose to output an ``act-or-repeat'' binary decision 
to decide if the action at the previous step should be repeated.
Because this ``act-or-repeat'' decision will be examined at every step depending on the current environment state, this results 
in \emph{closed-loop} action repetition.

All these action-repetition methods are well justified by the need of action \emph{persistence} \citep{dabney2020,amin2021,zhang2021,grigsby2021} 
for designing a good exploration strategy, when action \emph{diversity} should be traded for it properly.
This trade-off is important because when reward is sparse or short-term reward is deceptive, action diversity alone only makes the agent wandering around 
its local neighborhood since any persistent trajectory has an exponentially small probability. 
In such a case, a sub-optimal solution is likely to be found.
In contrast, persistence via action repetition makes the policy explore deeper (while sacrificing action diversity to some extent).

This paper further explores in the direction of closed-loop action repetition, striving to discover a novel algorithm that instantiates this idea better.
The key question we ask is, how can we exploit the special structure of closed-loop repetition, so that our algorithm yields better sample 
efficiency and final performance compared to existing methods? 
As an answer to this question, we propose 
\textbf{t}emporally \textbf{a}bstract \textbf{a}ctor-\textbf{c}ritic 
(\name), a simple but effective off-policy RL algorithm that incorporates closed-loop action repetition into an actor-critic framework.
Generally, we add a second stage that chooses between a candidate action 
output by an actor and the action from the previous step (Figure~\ref{fig:two-stage}).
Crucially, its ``act-or-repeat'' decision hinges on the actually sampled individual action instead of the expected behavior of the actor unlike recent works~\citep{Neunert2020,Chen2021}.
This post-acting switching scheme let the overall policy make more informed decisions.
Moreover,
\begin{compactenum}[i)]
\item for policy evaluation, we propose a new compare-through Q operator for multi-step TD backup tailored to the action repetition scenario, instead of replying on generic importance correction;
\item for policy improvement, we compute the actor gradient by multiplying a scaling factor to the $\frac{\partial Q}{\partial a}$ term
from DDPG~\citep{Lillicrap2016} and SAC~\citep{Haarnoja2018}, where the scaling factor is the optimal probability of choosing the actor's action in the second stage.
\end{compactenum}
\name is much easier to train compared to sophisticated HRL methods, while it has two important features compared to ``flat'' RL algorithms, namely, persistent exploration and native multi-step TD backup support without the need of off-policyness correction.

We evaluate \name on 14 continuous control tasks, covering simple control, locomotion, terrain walking~\citep{Brockman2016}, manipulation~\citep{plappert2018multigoal}, and self-driving~\citep{Dosovitskiy17}.
Averaged over these tasks, \name largely outperforms 6 strong baselines.
Importantly, our results show that it is our concrete instantiation of closed-loop action repetition that is vital to the final performance.
The mere idea of repeating actions in a closed-loop manner doesn't guarantee better results than the compared open-loop methods.
Moreover, our surprising finding reveals that while achieving top performance, TAAC is able to ``mine'' a significant number of repeated actions with the trained policy 
even on continuous tasks whose problem structures on the surface seem to repel action repetition (Section~\ref{sec:repeat_behavior}).
This suggests that aside from encouraging persistent exploration, action repetition can find its place in a good policy behavior.
This is perhaps due to that the action frequency of a task can be difficult to be set exactly as the minimum value that doesn't comprise optimal control while leaving no room 
for temporal abstraction \citep{grigsby2021}.

\section{Related work}
%
%Many works have been proposed to handle a hybrid continuous-discrete action space \citep{Xiong2018,Fan2019,Delalleau2019}.
%
%However, they mainly focus on environment-defined hybrid actions instead of temporal abstraction.
%
%In contrast, \name defines a binary switching action as a \emph{latent} variable and does not directly compute its Q value, as it is not from the environment.

Under the category of temporal abstraction via action repetition, there have been various formulations.
\citet{dabney2020} proposes temporally extended $\epsilon$-greedy exploration where a duration for repeating actions is sampled from a pre-defined truncated zeta distribution.
This strategy only affects the exploration behavior for generating off-policy data but does not change the training objective.
\citet{sharma2017learning} and \citet{Biedenkapp2021} learn a hybrid action space and treat the discrete action as a latent variable of action repeating steps, but the introduced temporal abstraction is open-loop and lacks flexibility.
One recent work close to \name is PIC \citep{Chen2021} which also learns to repeat the last action to address the action oscillation issue within consecutive
steps.
However, PIC was proposed for discrete control and its extension to continuous control is unclear yet.
Also, PIC predicts whether to repeat the last action \emph{independent of} a newly sampled action, which requires its switching policy to make a decision regarding the core policy's \emph{expected} behavior.
In an application section, H-MPO~\citep{Neunert2020} explored how continuous control can benefit from a meta binary action that modifies the overall system behavior.
Again, like PIC their binary decision is made \emph{in parallel with} a newly sampled action.
%
%Our method has a much simpler actor training step compared to H-MPO.
%
Different from PIC and H-MPO, TAAC only decides ``act-or-repeat'' after comparing the previous action with a newly sampled action.
Moreover, \name employs a new compare-through Q operator to exploit repeated actions for multi-step TD backup, and is trained by a much simpler actor gradient by absorbing the closed-form solution of the binary policy into the continuous action objective to avoid parameterizing a discrete policy unlike H-MPO.

Our experiment design (Section~\ref{sec:comparison_methods}) has covered most methods that consider action repetition. 
Table~\ref{tab:methods} provides a checklist of the differences between TAAC and these methods.

\section{Preliminaries}

We consider the RL problem as policy search in a Markov Decision
Process (MDP). Let $s\in \mathbb{R}^M$ denote a state, where
a continuous action $a\in \mathbb{R}^N$ is taken. Let
$\pi(a|s)$ be the action policy, and
$\mathcal{P}(s_{t+1}|s_t,a_t)$ the
probability of the environment transitioning to
$s_{t+1}$ after an action $a_t$ is taken at $s_t$. Upon reaching $s_{t+1}$, the agent 
receives a scalar reward $r(s_t, a_t, s_{t+1})$. 
%Assuming an initial state $s_0\sim\mathcal{P}(s_0)$, t
%
The RL objective is to find a policy $\pi^*$ that maximizes the expected discounted return: 
$\expect_{\pi,\mathcal{P}} \left[ \sum_{t=0}^{\infty} \gamma^t r(s_t, a_t, s_{t+1})\right]$,
where $\gamma \in (0, 1)$ is a discount factor. 
%
%In this paper, $\mathcal{P}$ will be  omitted in the expectation operator for notational simplicity. 
%
We also define $Q^{\pi}(s_t,a_t)=\expect_{\pi}[\sum_{t'=t}^{\infty}
\gamma^{t'-t}r(s_{t'}, a_{t'}, s_{t'+1})]$ as the discounted 
return starting from $s_t$ given that $a_{t}$ is taken and then $\pi$ is followed,
and $V^{\pi}(s_t)=\expect_{a_t\sim \pi}Q^{\pi}(s_t,a_t)$ as the discounted return starting from $s_t$ following $\pi$. 

%We denote the discounted state visitation distribution by $\rho^{\pi}(s)$,
%then the RL objective can also be written as $\expect_{s\sim \rho^{\pi}, a\sim \pi} Q^{\pi}(s,a)$.
%%% Haonan: Technically this is not correct, because the objective includes additional future r by averaging Q values over state distribution.
%%%         We can only do this for the policy gradient, because PG introduces another sum over time when decomposing the trajectory log prob. 
%%%.        See http://joschu.net/docs/thesis.pdf Eq 1 for policy gradient and TRPO paper for computing state distribution 

In an off-policy actor-critic 
setting with $\pi$ and $Q$ parameterized by $\thetaA$ and $\thetaQ$, a surrogate objective is usually 
used \citep{Lillicrap2016,Haarnoja2018}
\begin{equation}
\label{eq:rl_objective}
   \displaystyle \Max_{\thetaA}\expect_{s\sim\mathcal{D}} V^{\pi_{\thetaA}}_{\thetaQ}(s) \triangleq \Max_{\thetaA}\expect_{s\sim \mathcal{D},a\sim \pi_{\thetaA}} Q_{\thetaQ}(s,a).
\end{equation}
This objective maximizes the expected state value over some state distribution, 
assuming that 1) $s$ is sampled from a replay buffer $\mathcal{D}$ instead of the current policy, and 2) the 
dependency of the critic $Q_{\thetaQ}(s,a)$ on the policy $\pi_{\thetaA}$ is dropped when computing the gradient of $\thetaA$.
Meanwhile, $\thetaQ$ is learned separately via policy evaluation with typical TD backup.

\section{Temporally abstract actor-critic}
\label{submission}

\begin{wrapfigure}[8]{r}{0.43\textwidth}
\vspace{-9ex}
    \centering
    \includegraphics[width=0.43\textwidth]{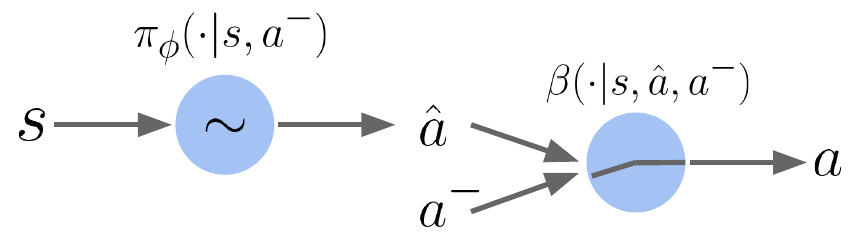}
    \caption{\name's two-stage policy during inference. 
    In the first stage, an action policy $\pi_{\thetaA}$ samples a candidate action $\hat{a}$.
    In the second stage, a binary switching policy $\beta$ chooses between this candidate and the previous action $a^-$.
    }
    \label{fig:two-stage}
\end{wrapfigure}

\iffalse
\begin{figure}[!t]
    \centering
    \includegraphics[width=0.5\textwidth]{images/two-stage.pdf}
    \caption{\name's two-stage policy during inference. 
    %
    In the first stage, an action policy $\pi_{\thetaA}$ samples a candidate action $\hat{a}$.
    %
    In the second stage, a binary switching policy $\beta$ chooses between this candidate and the previous action $a^-$.
    }
    \label{fig:two-stage}
\end{figure}
\fi

To enable temporal abstraction, we decompose the agent's action decision into 
two stages (Figure~\ref{fig:two-stage}): 1) sampling a new candidate action $\hat{a}\sim \pi_{\thetaA}(\cdot|s,a^-)$ conditioned on the action 
$a^-$ at the previous time step, and 2) choosing between $a^-$ and $\hat{a}$ as the actual output at the current step. 
The overall \name algorithm is summarized in Algorithm~\ref{alg:taac} Appendix~\ref{app:algorithm}.

\subsection{Two-stage policy}
\label{sec:policy}
Formally, let $\beta(b|s,\hat{a},a^-)$ be the binary
switching policy, where $b=0$/$1$ means choosing $a^-$/$\hat{a}$. For simplicity, 
in the following we will denote $\beta_b=\beta(b|s,\hat{a},a^-)$ (always assuming
its dependency on $s$, $\hat{a}$, and $a^-$). Then our two-stage policy $\pi^{\text{ta}}$ for temporal abstraction is defined as 
\begin{equation}
\label{eq:two-stage}
\begin{array}{@{}l}
    \pi^{\text{ta}}(a|s,a^-) %\triangleq \displaystyle\int_{\hat{a}} \pi_{\thetaA}(\hat{a}|s,a^-) P(a|s,\hat{a},a^-) \text{d}\hat{a}
    \triangleq \displaystyle\int_{\hat{a}} \pi_{\thetaA}(\hat{a}|s,a^-)\left[\beta_0\delta(a-a^-) + \beta_1\delta(a-\hat{a}) \right] \text{d}\hat{a},\\
\end{array}
\end{equation}
which can be shown to be a proper probability distribution of $a$.  
This two-stage policy repeats previous actions through a binary policy $\beta$, a decision maker that compares $a^-$ and $\hat{a}$ side by side
given the current state $s$. Repeatedly favoring $b=0$ results in temporal abstraction of executing the same action for multiple
steps. Moreover, this control is closed-loop, as it does not commit to a pre-determined
time window; instead it can stop repetition whenever necessary. As a 
special case, when $\beta_1=1$, $\pi^{\text{ta}}$ reduces to $\pi_{\thetaA}$; when $\beta_0=1$, $\pi^{\text{ta}}(a|s,a^-)=\delta(a-a^-)$.

%This temporal abstraction introduced by $\beta$ results in persistent exploration (Figure~\ref{fig:kde} Section~\ref{sec:visualization}) and unbiased multi-step Q operator for TD learning
%(Section~\ref{sec:policy_eval}).

%
\iffalse
Following SAC \citep{Haarnoja2018}, we can add the joint entropy $\mathcal{H}(\mathbf{\hat{a}}, \mathbf{b})$ as a bonus reward:
%
\begin{equation}
\label{eq:ent_v}
    V^{\pi^{\text{ta}}}_{\thetaQ}(s|a^-)=\displaystyle\expect_{\hat{a}\sim \pi_{\thetaA}, b\sim\beta}
    \left[ \begin{array}{l}
            Q_{\thetaQ}(s,a^-,\hat{a},b) \\
            - \alpha(\log \beta_b + \log \pi_{\thetaA}(\hat{a}|s,a^-))\\
           \end{array}
    \right],
\end{equation}
%
where $\alpha$ is a temperature that controls the importance of the entropy terms. 
\fi
%
%With our two-stage policy defined, next we describe how to perform policy evaluation for learning $Q_{\thetaQ}$ (Section~\ref{sec:policy_eval}) and policy improvement for learning $\pi^{\text{ta}}$ (Section~\ref{sec:policy_improve}).
%

\subsection{Policy evaluation with the compare-through operator}
\label{sec:policy_eval}
The typical one-step TD learning objective for a policy $\pi$ is
\begin{equation}
\label{eq:policy_eval}
    \begin{array}{l}
        \displaystyle \Min_{\thetaQ} \expect_{(s,a,s')\sim \mathcal{D}} \left[Q_{\thetaQ}(s,a)- \mathcal{B}^{\pi}Q_{\bar{\thetaQ}}(s,a)\right]^2,
        \text{with}\ \displaystyle\mathcal{B}^{\pi}Q_{\bar{\thetaQ}}(s,a) = r(s,a,s') + \gamma V^{\pi}_{\bar{\thetaQ}}(s'),\\
    \end{array}
\vspace{-1.5ex}
\end{equation}
where $\mathcal{B}^{\pi}$ is the Bellman operator, and $\bar{\thetaQ}$ slowly tracks $\thetaQ$ to stabilize the learning \citep{mnih2015humanlevel}. 
For multi-step bootstrapping, importance correction is usually needed, for example, Retrace~\citep{Munos2016} relies on a function of $\frac{\pi(a|s)}{\mu(a|s)}$ ($\mu$ is a behavior policy) to correct for the off-policyness of a trajectory.
Unfortunately, our $\pi^{\text{ta}}$ makes probability density computation challenging because of the marginalization over $\hat{a}$.
Thus importance correction methods including Retrace cannot be applied to our case.
To address this issue, below we propose a new multi-step Q operator, called \emph{compare-through}.
Then we explain how $\pi^{\text{ta}}$ can exploit this operator for efficient policy evaluation.

For a learning policy $\pi$, given a trajectory $(s_0,a_0,s_1,a_1,\ldots, s_N,a_N)$ from a behavior policy $\mu$,
let $\ta_n$ denote the actions sampled from $\pi$ at states $s_n (n\ge 1)$ and $\tilde{a}_0=a_0$ (we do not sample from $\pi$ at $s_0$).
We define (a point estimate of) the compare-through operator $\mathcal{T}^{\pi}$ as 

\vspace{-3ex}
\begin{equation}
\label{eq:operator-estimate}
\begin{array}{l}
\mathcal{T}^{\pi}Q_{\bar{\thetaQ}}(s_0,a_0) \approx \displaystyle\sum_{t=0}^{n-1} \gamma^t r(s_t,a_t,s_{t+1}) + \gamma^n Q_{\bar{\theta}}(s_n,\ta_n),\\
\end{array}
\end{equation}
where $n=\min \left(\{n: \tilde{a}_n \ne a_n\} \cup \{N\}\right)$.
Intuitively, given a sampled trajectory of length $N$ from the replay buffer, the compare-through operator takes an expectation, 
under the current policy at the sampled states (from $s_1$ to $s_N$), over all the sub-trajectoriess (up to length $N$) of actions that match the sampled actions.
Note that Eq.~\ref{eq:operator-estimate} is a \emph{point estimate} of this expectation.
%
%starting from $n=1$, we set $n\leftarrow n+1$ if $\ta_n=a_n$.
%
%We repeat this comparison until $\ta_n\ne a_n$ or $n=N$, and use $Q_{\bar{\thetaQ}}(s_n,\ta_n)$ as the bootstrapping Q value.
%
A formal definition of $\mathcal{T}^{\pi}$ is described by Eq.~\ref{eq:operator}, and its relation to Retrace~\citep{Munos2016} is shown in Appendix~\ref{app:policy_eval_proof}.

\begin{wrapfigure}[16]{r}{0.45\textwidth}
\vspace{-3ex}
    \centering
    \includegraphics[width=0.45\textwidth]{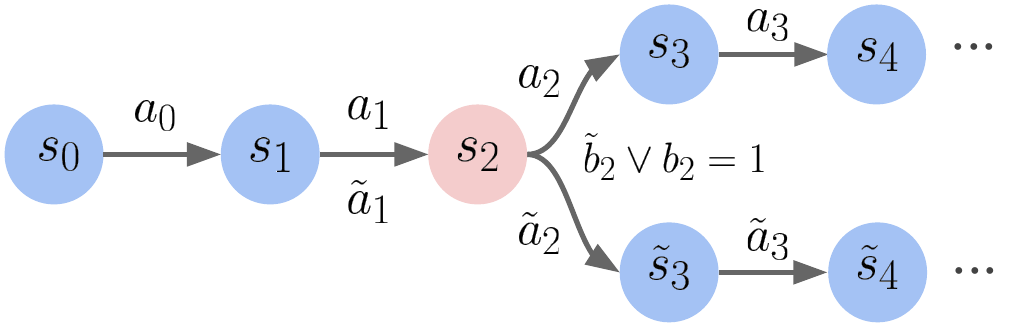}
    \caption{An illustration of the compare-through operator by exploiting action repetition of $\pi^{\text{ta}}$. The upper branch is the trajectory sampled by a rollout policy; the lower one is sampled by the current policy during training. We have $\ta_1=a_1=a_0$ due to $b_1=\tb_1=0$.
    The two trajectories diverge at $s_2$ because either $b_2=1$ or $\tb_2=1$. For 
    bootstrapping, we use $s_2$ as the target state in this example.
    }
    \label{fig:multi-step}
\end{wrapfigure}

\iffalse
\begin{figure}[!t]
    \centering
    \includegraphics[width=0.5\textwidth]{images/multi-step.pdf}
    \caption{An illustration of the compare-through operator by exploiting action repetition of $\pi^{\text{ta}}$. The upper branch is the trajectory sampled by a rollout policy; the lower one is sampled by the current policy during training. We have $\ta_1=a_1=a_0$ due to $b_1=\tb_1=0$.
    The two trajectories diverge at $s_2$ because either $b_2=1$ or $\tb_2=1$. For 
    bootstrapping, we use $s_2$ as the target state in this example.
    }
    \label{fig:multi-step}
\end{figure}
\fi

\begin{manualtheorem}{1}[Policy evaluation convergence]
    In a tabular setting, the compare-through operator $\mathcal{T}^{\pi}$, whose point estimate defined by Eq.~\ref{eq:operator-estimate} (without the parameters $\bar{\thetaQ}$) and expectation form defined by Eq.~\ref{eq:operator}, has a unique fixed point $Q^{\pi}$, where $\pi$ is the current (target) policy.
\end{manualtheorem}
For the detailed proof we refer the reader to Appendix~\ref{app:policy_eval_proof}.
Although the actual setting considered in this paper are continuous state and action domains with function approximators, Theorem 1 still provides some justification for our compare-through operator.
%
%This suggests that when applying $\mathcal{T}^{\pi}$ to Eq.~\ref{eq:policy_eval}, $Q_{\thetaQ}$ will be driven to fit $Q^{\pi}$.

Clearly, any discrete policy could exploit the compare-through operator since there can be a non-zero chance of two discrete actions being compared equal.
A typical stochastic continuous policy such as Gaussian used by SAC~\citep{Haarnoja2018} always has $\ta_n\ne a_n$ for $n\ge 1$ w.r.t. any behavior policy $\mu$.
In this case $\mathcal{T}^{\pi}$ is no more than just a Bellman operator $\mathcal{B}^{\pi}$.
However, if a continuous policy is specially structured to be ``action-reproducible'', it will enjoy the privilege of using $s_n$ for $n>1$ as the target state.
Our two-stage $\pi^{\text{ta}}$ is such a case, where each action $a_n$ ($\ta_n$) is accompanied by a repeating choice $b_n$ ($\tb_n$).
Starting from $n=1$ with a previous action $a_0$, if $\tb_m=b_m=0$ (both repeated) for all $1\le m \le n$, then we know that $\ta_n=a_n=\ldots=\ta_1=a_1=a_0$.
%\footnote{There is an assumption that the behavior policy repeats the previous action with a non-zero probability at every state. This can be satisfied if we add noise to its $\beta$ or enforce $\beta$'s entropy to be positive.}.
%
In other words, if two trajectories start with the same $(s_0,a_0)$, we can compare their discrete $\{b_n\}$ sequences in place of the continuous $\{a_n\}$ sequences.
See Figure~\ref{fig:multi-step} for an illustration.
Thus for multi-step TD learning we use $\mathcal{T}^{\pi^{\text{ta}}}$ to replace $\mathcal{B}^{\pi}$ in Eq.~\ref{eq:policy_eval}.

\textbf{Remark}\ The compare-through operator is not meant to replace multi-step TD learning with importance correction in a general scenario,
as it is only effective for ``action-reproducible'' policies. 
Its bootstrapping has a hard cutoff (by checking $\ta_n\ne a_n$) instead of a soft one as in \citet{Munos2016}.
%
%and the step is always smaller than the maximal bootstrapping step $N$.

\textbf{On the importance of reward normalization}\ Because $\mathcal{T}^{\pi^{\text{ta}}}$ computes bootstrapping targets based on dynamic step lengths and rewards are propagated faster with greater lengths, 
Q values bootstrapped by greater lengths might become overly optimistic/pessimistic (\eg\ imagine a task with rewards that are all positive/negative).
This could affect the policy selecting actions according to their Q values.
However, this side effect is only \emph{temporary} and will vanish if all Q values are well learned eventually.
In practice, we find it beneficial to normalize the immediate reward by its moving average statistics to roughly maintain a zero mean and unit standard deviation (detailed in Appendix~\ref{app:reward_norm}).

\subsection{Policy improvement with a closed-form $\beta^*$}
\label{sec:policy_improve}
It can be shown (Appendix~\ref{app:state_value}) that the parameterized state value of $\pi^{\text{ta}}$ is
\begin{equation}
\label{eq:v}
\begin{array}{c}
    V^{\pi^{\text{ta}}}_{\thetaQ}(s|a^-)=\displaystyle\expect_{\hat{a}\sim \pi_{\thetaA}, b\sim \beta}\left[(1-b)Q_{\thetaQ}(s,a^-) + bQ_{\thetaQ}(s,\hat{a})\right],
\end{array}
\end{equation}
Intuitively, the actual Q value of each $a\sim\pi^{\text{ta}}$
is an interpolation by $\beta$ between the Q values of $a^-$ and $\hat{a}$. 
\iffalse
This is consistent with 
our motivation: if $a^-$ is taken, by the Q definition the 
future discounted return is just $Q(s,a^-)$, otherwise $Q(s,\hat{a})$.
%
Note that unlike SAC~\citep{Haarnoja2018} here we choose not to add entropy as a bonus reward in $V_{\thetaQ}^{\pi^{\text{ta}}}$.
%
Instead, entropy will be used merely as a regularization term for policy improvement (Section~\ref{sec:policy_improve})~
\iffalse
\footnote{Strictly speaking, our two-stage policy $\pi^{\text{ta}}$ is no longer ``soft'' as its value does not include the entropy reward.
%
But we still call the proposed algorithm ``TASAC'' to remind that its overall training framework is very analogous to SAC's.
}.
\fi
%
This choice can be justified by a recent work~\citep{Wang2020} which shows that the major contribution of the entropy term in SAC
is to resolve the action squashing problem during actor optimization.
\fi
%p
Our policy improvement objective is then $\max_{\thetaA,\beta}\ \expect_{(s,a^-)\sim \mathcal{D}} V^{\pi^{\text{ta}}}_{\thetaQ}(s|a^-)$.
Note that each time we sample the previous action $a^-$ along with a state $s$ from the replay buffer.
To encourage exploration, following prior works~\citep{mnih2016asynchronous,Riedmiller2018} we augment this objective with a joint entropy 
$\mathcal{H}_{s,a^-}=\expect_{\pi_{\thetaA}(\hat{a}|s,a^-)\beta_b}\left[- \log \beta_b -\log \pi_{\thetaA}(\hat{a}|s,a^-) \right]$.
%
\iffalse
\begin{equation*}
\begin{array}{l}
    \mathcal{H}_{s,a^-} =-\displaystyle\expect_{P(\hat{a},b|s,a^-)}\log P(\hat{a},b|s,a^-)
    =-\displaystyle\expect_{\pi_{\thetaA}(\hat{a}|s,a^-)\beta_b}\left[\log \pi_{\thetaA}(\hat{a}|s,a^-) + \log \beta_b\right]\\
\end{array}
\end{equation*}
\fi
%
Thus the final policy improvement objective is
\begin{equation}
\label{eq:ent_v_opt}
    \begin{array}{l}
    \displaystyle\max_{\thetaA,\beta}\ \expect_{(s,a^-)\sim \mathcal{D}} \left[V^{\pi^{\text{ta}}}_{\thetaQ}(s|a^-)+\alpha \mathcal{H}_{s,a^-}\right]\\
    = \displaystyle\max_{\thetaA,\beta}\ \expect_{\substack{(s,a^-) \sim \mathcal{D}\\ \hat{a}\sim \pi_{\thetaA}, b\sim\beta}}
    \left[ 
        \begin{array}{l}
            (1-b)Q_{\thetaQ}(s,a^-) + bQ_{\thetaQ}(s,\hat{a})
            - \alpha\left(\log \beta_b + \log \pi_{\thetaA}(\hat{a}|s,a^-)\right)\\
        \end{array}
    \right],\\
    \end{array}\\
\end{equation}
where $\alpha$ is a temperature parameter.
Given any $(s,a^-,\hat{a})$, we can derive a closed-form solution of the (non-parametric) $\beta$ policy for the innermost expectation $b\sim\beta$ as 
\[\beta^*_1 = \displaystyle \left.\exp\left(\frac{Q_{\thetaQ}(s,\hat{a})}{\alpha}\right) \middle/ \left(\exp\left(\frac{Q_{\thetaQ}(s,\hat{a})}{\alpha}\right) + \exp\left(\frac{Q_{\thetaQ}(s,a^-)}{\alpha}\right)\right)\right. .\]
Then applying the re-parameterization trick $\hat{a}=f_{\thetaA}(\epsilon,s,a^-), \epsilon\sim \mathcal{N}(0,I)$, one can show that the estimated actor gradient is
\begin{equation}
\label{eq:policy_gradient}
\begin{array}{rcl}
%\triangleq&\displaystyle\frac{V_{\thetaQ}^{\pi^{\text{ta}}}(s|a^-)}{\partial \thetaA}\\
\displaystyle\Delta\thetaA & \triangleq &\displaystyle\left(\beta^*_1\frac{\partial Q_{\thetaQ}(s,f_{\thetaA})}{\partial \thetaA} - \alpha\frac{\partial \log\pi_{\thetaA}(f_{\thetaA}|s,a^-)}{\partial \thetaA} \right)\\
&=&\displaystyle\left(\beta^*_1\frac{\partial Q_{\thetaQ}(s,\hat{a})}{\partial \hat{a}} - \alpha\frac{\partial \log\pi_{\thetaA}(\hat{a}|s,a^-)}{\partial \hat{a}}\right)
\frac{\partial f_{\thetaA}}{\partial \thetaA} 
- \displaystyle\alpha \frac{\partial \log \pi_{\thetaA}(\hat{a}|s,a^-)}{\partial \thetaA}.\\
\end{array}
\end{equation}
This gradient has a very similar form with SAC's \citep{Haarnoja2018}, except that here
$\frac{\partial Q}{\partial \hat{a}}$ has a scaling factor $\beta_1^*$.
We refer the reader to a full derivation of the actor gradient in Appendix~\ref{app:actor-gradient}.

\textbf{Remark on $\beta^*$}\ This closed-form solution is only possible after $\hat{a}$ is sampled. 
It's essentially comparing the Q values between $a^-$ and $\hat{a}$.
This side-by-side comparison is absent in previous closed-loop repeating methods like PIC~\citep{Chen2021} and H-MPO~\citep{Neunert2020}.

\iffalse
\textbf{Discrete action}
%
For discrete action, we can avoid parameterizing/training the action policy $\pi_{\thetaA}$. Similar to the derivation of $\beta^*$, in the discrete
case, we derive the optimal first-stage policy $\pi^*(\hat{a}|s,a^-)$ given any sampled $(s,a^-)$, from Eq.~\ref{eq:ent_v_simple} (Appendix~\ref{app:closed-form}):
%
\begin{equation}
\label{eq:optimal_a}
\pi^*(\hat{a}|s,a^-)\propto \sum_{b=0}^1\exp(\frac{Q_{\thetaQ}(s,a^-,\hat{a},b)}{\alpha}).
\end{equation}
%
This optimal policy can be easily computed because the normalizer is tractable in the discrete case. Suppose that there are $N$ actions. Then
%
\[\displaystyle\pi^*(i|s,a^-)=\frac{\exp(\frac{Q_{\thetaQ}(s,a^-)}{\alpha}) + \exp(\frac{Q_{\thetaQ}(s,i)}{\alpha})}{N\cdot\exp(\frac{Q_{\thetaQ}(s,a^-)}{\alpha}) + \sum_{i'=0}^{N-1}\exp(\frac{Q_{\thetaQ}(s,i')}{\alpha})}.\]
%
As a special case, if $Q_{\thetaQ}(s,a^-)\gg Q_{\thetaQ}(s,i)$ for $i\ne a^-$, then 
%
\begin{equation*}
\pi^*(i)=\Big\{ 
\begin{array}{ll}
    \frac{2}{N+1}, & i=a^-\\
    \frac{1}{N+1}, & \text{else}\\
\end{array}
\end{equation*}
%
When $N$ is large, $\pi^*$ is basically uniform, meaning that if the previous action $a^-$ is very good, then at the second stage $\beta$ will mostly choose $a^-$ regardless of the sampled candidate $\hat{a}\sim\pi^*$.
%
On the other hand, if $Q_{\thetaQ}(s,a^-)\ll Q_{\theta}(s,i)$ for $i\ne a^-$, $\pi^*$ reduces to a typical Q-based Boltzmann distribution.
\fi

\textbf{Remark on multi-step actor gradient}\ According to Figure~\ref{fig:two-stage}, the newly generated action $\hat{a}_t$ at the current step $t$, if repeated as future $a^-_{t+1},\ldots,a^-_{t'}$ ($t'>t$), will also influence the maximization of future V values: $V_{\thetaQ}^{\pi^{\text{ta}}}(s_{t+1}|a^-_{t+1}),\ldots,V_{\thetaQ}^{\pi^{\text{ta}}}(s_{t'}|a^-_{t'})$.
As a result, in principle $\hat{a}_t$ has a multi-step gradient.
To exactly compute this multi-step gradient, a fresh rollout of the current $\pi^{\text{ta}}$ via interacting with the environment is necessary.
For reasons detailed in Appendix~\ref{app:multi-step-actor-grad}, we end up truncating this full gradient to the first step, by letting Eq.~\ref{eq:ent_v_opt} sample $(s,a^-)$ pairs at non-consecutive time steps from the replay buffer.
This one-step truncation is also (implicitly) adopted by H-MPO~\citep{Neunert2020} for their action repetition application case.
Interestingly, the truncation results in a simple implementation of actor gradient of \name similar to SAC's.

%
\iffalse
To exactly quantify this multi-step influence, starting from a replayed $s_t$, we need to perform a fresh rollout of $\pi^{\text{ta}}$ to estimate which future states $s_{t+k}$ will adopt $\hat{a}_t$ (via repeating).
%
Let $\beta_b^*(t+k)=\beta^*(b|s_{t+k},\hat{a}_{t+k},a^-_{t+k})$ be the optimal repeating policy at step $t+k$ and define $w_{t+k}=\beta_1^*(t)\prod_{k'=1}^k\beta^*_0(t+k')$ to be the probability of $s_{t+k}$ adopting $\hat{a}_t$.
%
Then the first term ($\beta^*_1\frac{\partial Q_{\theta}}{\partial \hat{a}_t}$) in Eq.~\ref{eq:policy_gradient} now becomes $\sum_{k=0}^{\infty}w_{t+k}\frac{\partial Q_{\theta}(s_{t+k},\hat{a}_t)}{\partial \hat{a}_t}$.
%
while the weight of $\hat{a}$ on future $V_{\thetaQ}^{\pi^{\text{ta}}}$ values decays exponentially.
while the form of $\pi^{\text{ta}}$ does not allow an easy computation of probability density for importance correction if behavior trajectories are used.
%
\item $\frac{\partial Q_{\thetaQ}(s_{t+k},\hat{a}_t)}{\partial \hat{a}_t}$ has a much higher sample variance as $k$ increases.
%
\item $w_{t+k}$ decreases exponentially so the importance of future $\frac{\partial Q_{\thetaQ}}{\partial \hat{a}_t}$ quickly decays.
%
\fi
    
\textbf{Automatically tuned temperatures}\ Given two entropy targets $\mathcal{H}'$ and $\mathcal{H}''$, we learn temperatures $\alpha'$ and $\alpha''$ from the objective
\begin{equation}
\label{eq:alpha}
\min_{\substack{\log(\alpha'),\\ \log(\alpha'')}} \left\{\expect_{(s,a^-)\sim\mathcal{D}, \hat{a}\sim \pi_{\thetaA}, b\sim \beta^*} \Big[\log(\alpha') (- \log \beta^*_b - \mathcal{H}') + \log(\alpha'')(-\log \pi_{\thetaA}(\hat{a}|s,a^-)  - \mathcal{H}'')\Big]\right\},
\end{equation}
by adjusting $\log(\alpha)$ instead of $\alpha$ as in SAC~\citep{Haarnoja2018}.
We learn temperatures for $\pi_{\thetaA}$ and $\beta$ separately, 
to enable a finer control of the two policies and their entropy terms, similar to the separate policy constraints \citep{Abdolmaleki2018,Neunert2020}.
Appendix~\ref{app:different-temps} shows how several equations slightly change if two temperatures are used ($\alpha$ is replaced by $\alpha'$ or $\alpha''$).

\section{Experiments}
\label{sec:experiments}

\subsection{Tasks}
\label{sec:tasks}
To test if the proposed algorithm is robust and can be readily applied to many tasks, we perform experiments over 14 continuous control tasks under different scenarios:
\begin{compactenum}[a)]
    \item \textbf{SimpleControl}: Three control tasks \citep{Brockman2016} with small action and observation spaces:
        \task{MountainCarContinuous}, \task{LunarLanderContinuous}, and \task{InvertedDoublePendulum} ;
    \item \textbf{Locomotion}: Four locomotion tasks \citep{Brockman2016} that feature complex physics and action spaces: 
        \task{Hopper}, \task{Ant}, \task{Walker2d}, and \task{HalfCheetah};
    \item \textbf{Terrain}: Two locomotion tasks that require adapting to randomly generated terrains: \task{BipedalWalker} and \task{BipedalWalkerHardcore};
    \item \textbf{Manipulation}: Four Fetch \citep{plappert2018multigoal} tasks with sparse rewards and hard exploration (reward given only upon success): 
        \task{FetchReach}, \task{FetchPush}, \task{FetchSlide}, and \task{FetchPickAndPlace};
    \item \textbf{Driving}: One CARLA autonomous-driving task \citep{Dosovitskiy17} that has complex high-dimensional multi-modal sensor inputs (camera, radar, IMU, collision, GPS, \etc): \task{Town01}. The goal is to reach a destination starting from a randomly spawned location in a small realistic town, while avoiding collisions and red light violations.
\end{compactenum}
Among the 5 categories, (d) and (e) might benefit greatly from temporal abstraction because of hard exploration or the problem structure (\eg\ driving naturally involves repeated actions). Categories (a)-(c) appear unrelated with temporal abstraction, but we test if seemingly unrelated tasks can also benefit from it.
By comparing \name against different methods across vastly different tasks, we hope to demonstrate its generality, because adaptation to this kind of task variety requires few assumptions
about the task structure and inputs/outputs.
%
%Some example task screenshots are shown in Figure~\ref{fig:task-examples}.
%
For more task details, we refer the reader to Appendix~\ref{app:tasks}.

\iffalse
\begin{figure}[!t]
    \centering
    \resizebox{0.48\textwidth}{!}{
    \begin{tabular}{@{}c@{}c@{}c@{}c@{}c@{}}
    \includegraphics[height=\textheight]{images/pendulum.png}
    &\includegraphics[height=\textheight]{images/locomotion.png}    
    &\includegraphics[height=\textheight]{images/bipedal_hard.png}
    &\includegraphics[height=\textheight]{images/fetch.png}
    &\includegraphics[height=\textheight]{images/carla.png}\\
    \end{tabular}
    }
    \caption{Example tasks (out of 14) evaluated in our experiments. From left to right: \task{InvertedDoublePendulum}, 
    \task{HalfCheetah}, \task{BipedalWalkerHardcore}, \task{FetchSlide}, and \task{Town01}.}
\label{fig:task-examples}
\end{figure}
\fi

\begin{table*}[!t]
    \centering
    \resizebox{\textwidth}{!}{
    \begin{tabular}{l|c|c|c|c|c|c|c|c}
         & SAC & \multirow{2}{*}{SAC-Ntd} & \multirow{2}{*}{SAC-Nrep} & SAC-Krep & SAC-EZ & SAC-Hybrid & \multirow{2}{*}{\name-1td} 
         & \multirow{2}{*}{\name} \\
         & \citep{Haarnoja2018} & & & \citep{sharma2017learning} & \citep{dabney2020} & \citep{Neunert2020} & & \\
         & & & & \citep{Biedenkapp2021} & & & & \\
         \hline
         Persistent exploration & \xmark & \xmark & \cmark & \cmark & \cmark & \cmark & \cmark & \cmark \\
         Multi-step TD & \xmark & \cmark & \cmark & \cmark & \xmark & \cmark & \xmark & \cmark \\
%         Unbiased Q operator & \cmark & \cmark & \cmark & \cmark & \cmark & \cmark & \cmark & \cmark \\
         Closed-loop repetition & \xmark & \xmark & \xmark & \xmark & \xmark & \cmark & \cmark & \cmark \\
         Learnable duration & \xmark & \xmark & \xmark & \cmark & \xmark & \cmark & \cmark & \cmark \\
    \end{tabular}
    }
    \caption{A summary of the 8 major comparison methods in our experiments. 
    %``Persistent exploration'': the ability of exploring along one action direction persistently;
    %``Multi-step TD'': using $N$-step ($N>1$) targets for TD learning;
%    ``Unbiased Q operator'': no off-policyness in the Q operator for TD learning;
    %``Closed-loop repetition'': the ability of terminating action repetition at any time without a duration commitment;
    %``Learnable duration'': learning an (implicit) action duration model that depends on the current state.
    }
    \label{tab:methods}
\vspace{-1ex}
\end{table*}

\subsection{Comparison methods}
\label{sec:comparison_methods}
While there exist many off-policy hierarchical RL 
methods that model temporal abstraction, for example \citet{nachum2018dataefficient,Riedmiller2018,Levy2019,Li2020,zhang2021hierarchical}, 
we did not find them readily scalable to our entire list of tasks (especially to high dimensional input space like CARLA), without considerable efforts of adaptation. 
Thus our primary focus is to compare \name with baselines of different formulations of action repetition: vanilla SAC \citep{Haarnoja2018}, SAC-Nrep, SAC-Krep~\citep{sharma2017learning,Biedenkapp2021}, SAC-EZ~\citep{dabney2020}, and SAC-Hybrid~\citep{Neunert2020}.
Although originally some baselines have their own RL algorithm backbones, in this experiment we choose SAC as the common backbone because: 1) SAC is state-of-the-art among  open-sourced actor-critic algorithms, 2) a common backbone facilitates reusing the same set of core hyperparameters for a fairer comparison, and 3) by reducing experimental variables, it gives us a better focus on the design choices of action repetition instead of other orthogonal algorithmic components.

Let $N$ be a parameter controlling the maximal number of action repeating steps.
SAC-Nrep simply repeats every action $N$ times. 
%
%We find this simple strategy surprisingly effective in some scenarios when there exists large room for temporal 
%abstraction.
SAC-Krep, inspired by FiGAR \citep{sharma2017learning} and TempoRL~\citep{Biedenkapp2021}, upgrades an action $a$ to a pair 
of $(a, K)$, indicating that the agent will repeat action $a$ for the next $K$ steps ($1\le K\le N$) without being interrupted
until finishing.
To implement SAC-Krep, following \citet{Delalleau2019} we extended the original SAC algorithm to support a mixture of 
continuous and discrete actions (Appendix~\ref{app:sac-mixed}).
Note that SAC-Krep's open-loop control is in contrast to \name's closed-loop control.
SAC-EZ incorporates the temporally extended $\epsilon$-greedy exploration \citep{dabney2020} into SAC. During rollout, if the agent 
decides to explore, then the action is uniformly sampled and the duration for repeating that action is sampled from a truncated zeta 
distribution $zeta(n)\propto n^{-\mu}$, $1\le n \le N$. 
This fixed duration model encourages persistent exploration depending on the value of 
the hyperparameter $\mu$. 
The training step of SAC-EZ is the same with SAC.
SAC-Hybrid shares a similar flavor of H-MPO~\citep{Neunert2020} for closed-loop action repetition.
It defines a hybrid policy to output the continuous action and binary switching action in parallel, assuming their conditional independence given the state.
This independence between hybrid actions and how the Q values are computed in SAC-Hybrid are the biggest differences with \name.
We also apply Retrace~\citep{Munos2016} to its policy evaluation with $N$-step TD as done by H-MPO.
We refer the reader to the algorithm details of SAC-Hybrid in Appendix~\ref{app:sac-hybrid}.

%While open-loop methods, like SAC-Nrep and SAC-Krep, introduce persistent exploration, they also have a built-in benefit 
%of multi-step TD learning, because their reward for training is accumulated over several steps. 
%
In order to analyze the benefit of persistent exploration independent of that of multi-step TD learning, we also compare two additional methods.
SAC-Ntd is a variant of SAC where a Q value is bootstrapped by an $N$-step value target with Retrace~\citep{Munos2016} to correct for off-policyness.
%
%Recall that our full method \name employs the compare-through operator $\mathcal{T}^{\pi^{\text{ta}}}$ proposed in Section~\ref{sec:policy_eval} depending on the output of $\beta$, up to $N$ steps.
%
For ablating \name, we evaluate \name-1td that employs a typical Bellman operator $\mathcal{B}^{\pi^{\text{ta}}}$ for one-step bootstrapping.
%
%Finally, \name-Ntd always bootstraps a critic with an $N$-step ($N>1$) target like SAC-Ntd, regardless of $\beta$'s outputs.
%
Thus we have 8 methods in total for comparison in each task.
See Table~\ref{tab:methods} for a summary and Appendix~\ref{app:experiments} for method details.

In our experiments, we set the repeating hyperparameter $N$ to 3 on \textbf{SimpleControl}, \textbf{Locomotion} and \textbf{Manipulation}, and 
to 5 on \textbf{Terrain} and \textbf{Driving}.
Here the consideration of $N$ value is mainly for open-loop methods like SAC-Nrep and SAC-Krep because they will yield poor performance with large values of $N$.
\name is not sensitive to $N$ for policy evaluation, as shown in Section~\ref{sec:off-policy-exp}.

\subsection{Evaluation protocol}
\begin{wrapfigure}[23]{r}{0.4\textwidth}
\vspace{-5ex}
    \centering
    \resizebox{0.4\textwidth}{!}{
        \includegraphics[width=\textwidth]{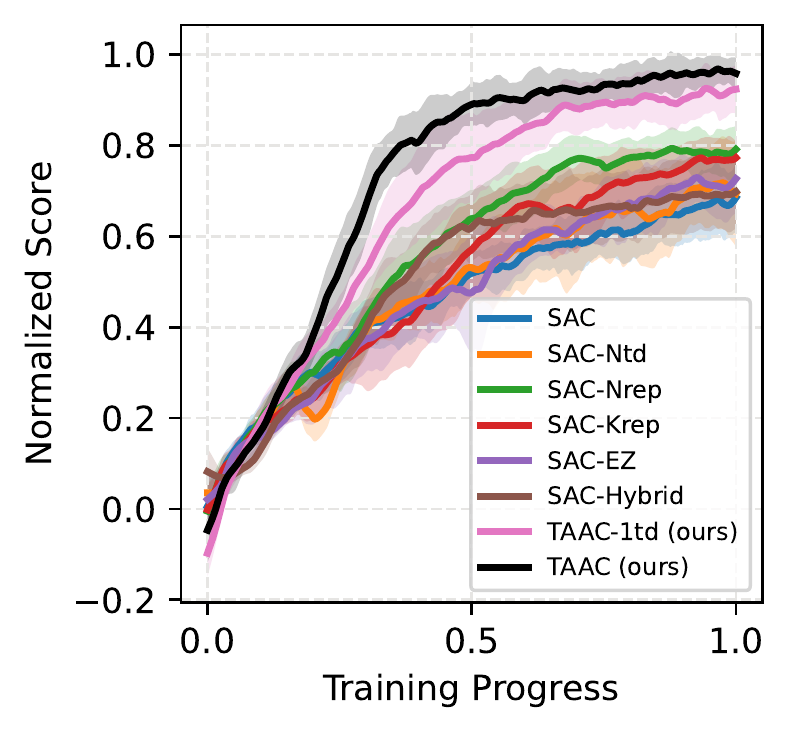}
    }
    \caption{Training curves (n-score \vs training progress) of the 8 comparison methods in one plot.
    Each curve is a mean of a method’s n-score curves over all the 14 tasks, where the method is run with 3 random seeds on each task.
    See Figure~\ref{fig:curves} and Figure~\ref{fig:unnormalized-curves} (Appendix~\ref{app:more-results}) for the complete set of individual training curves.
    }
\label{fig:overall_curves}
\end{wrapfigure}
To measure the final model quality, we define \textit{score} as the episodic return $\sum_{t=0}^T r(s_t,a_t,s_{t+1})$ of evaluating (by taking the approximate mode of 
a parameterized continuous policy; see Appendix~\ref{app:eval} for details) a method for a task episode, where $T$ is a pre-defined time limit or when the episode terminates early.
Different tasks, even within the same group, can have vastly different reward scales (\eg\ tens \vs\ thousands). 
So it is impractical to directly average their scores.
It is not uncommon in prior works \citep{Hessel2018} to set a performance range for each task separately, and normalize the score of that task to roughly $[0,1]$ 
before averaging scores of a method over multiple tasks. 
Similarly, to facilitate score aggregation across tasks, we adopt the metric of \textit{normalized score} (\emph{n-score}).
For each task, we obtain the score (averaged over 100 episodes) of a random policy and denote it by $Z_0$.
We also evaluate the best method on that task and obtain its average score $Z_1$.
Given a score $Z$, its normalized value is calculated as $\frac{Z-Z_0}{Z_1-Z_0}$.
With this definition, the n-score of each task category (a)-(e) can be computed as the averaged n-score across tasks within that category.
Additionally, to measure training convergence speed, we approximate \emph{n-AUC} (area under the n-score curve normalized by $x$ value range) by averaging n-scores on a n-score curve throughout the training.
A higher n-AUC value indicates a faster convergence speed.
n-AUC is a secondary metric to look at when two methods have similar final n-scores. 
Finally, by averaging n-score and n-AUC values over multiple tasks, we emphasize the robustness of an RL method. 
%
%It is not uncommon to see some RL methods perform well on a very narrow domain of tasks, but difficult to adapt to others. 
%
%This situation is what we try to avoid when designing the evaluation metrics.

Given a task, we train each method for the same number of environment frames. 
Crucially, for fair comparisons we also make each method train 1) for the same number of gradient steps, 2) with the same mini-batch size and learning rate, 3) using roughly the same number of weights, and 4) with a common set of hyperparameters (tuned with vanilla SAC) for the SAC backbone .
More details of the experimental settings are described in Appendix~\ref{app:experiments}.

\begin{table}[!t]
\setlength\tabcolsep{2pt}
\newcolumntype{?}{!{\vrule width 2pt}}
\newcolumntype{a}{>{\columncolor{light-gray}}c}
    \centering
    \resizebox{\textwidth}{!}{
    \begin{tabular}{@{}ll|c|c|c|c|c|c|a|a@{}}
        & & SAC & SAC-Ntd & SAC-Nrep & SAC-Krep & SAC-EZ & SAC-Hybrid & \name-1td & \name\\ 
        \hline
        \multirow{5}{*}{Final n-score}
        &\textbf{SimpleControl} & $0.64[0.03]$ & $0.54[0.07]$ & $0.84[0.01]$ & $0.55[0.03]$ & $0.65[0.03]$ & $0.88[0.14]$ & $0.83[0.13]$ & $\mathbf{0.99}[0.04]$\\
        &\textbf{Locomotion} & $0.89[0.06]$ & $0.88[0.11]$ & $0.70[0.08]$ & $0.82[0.07]$ & $\mathbf{0.92}[0.07]$ & $0.63[0.08]$ & $0.88[0.04]$ & $0.90[0.05]$\\
        &\textbf{Terrain} & $0.35[0.10]$ & $0.48[0.02]$ & $0.54[0.03]$ & $0.48[0.03]$ & $0.45[0.15]$ & $0.79[0.03]$ & $0.96[0.02]$ & $\mathbf{1.00}[0.03]$\\
        (model quality)&\textbf{Manipulation} & $0.60[0.12]$ & $0.75[0.17]$ & $0.91[0.07]$ & $0.98[0.05]$ & $0.68[0.07]$ & $0.49[0.07]$ & $\mathbf{0.99}[0.02]$ & $\mathbf{0.99}[0.01]$\\
        &\textbf{Driving} & $0.88[0.05]$ & $0.84[0.04]$ & $0.92[0.03]$ & $0.95[0.04]$ & $0.71[0.03]$ & $\mathbf{1.00}[0.04]$ & $0.92[0.02]$ & $0.97[0.04]$\\

        &\cellcolor{blue!15}\textbf{All} 
        &\cellcolor{blue!15} $0.68[0.08]$
        &\cellcolor{blue!15} $0.71[0.10]$
        &\cellcolor{blue!15} $0.78[0.05]$
        &\cellcolor{blue!15} $0.77[0.05]$
        &\cellcolor{blue!15} $0.71[0.07]$ 
        &\cellcolor{blue!15} $0.69[0.08]$
        &\cellcolor{blue!15} $0.92[0.05]$
        &\cellcolor{blue!15} $\mathbf{0.96}[0.03]$\\
        \hline
        \multirow{5}{*}{n-AUC}
        &\textbf{SimpleControl} & $0.45[0.01]$ & $0.41[0.01]$ & $0.51[0.02]$ & $0.28[0.03]$ & $0.45[0.02]$ & $0.62[0.08]$ & $0.60[0.10]$ & $\mathbf{0.72}[0.04]$\\
        &\textbf{Locomotion} & $0.69[0.03]$ & $0.64[0.07]$ & $0.55[0.05]$ & $0.59[0.04]$ & $0.64[0.05]$ & $0.50[0.06]$ & $0.72[0.02]$ & $\mathbf{0.74}[0.05]$\\
        &\textbf{Terrain} & $0.17[0.02]$ & $0.19[0.02]$ & $0.38[0.01]$ & $0.23[0.01]$ & $0.21[0.03]$ & $0.50[0.01]$ & $0.50[0.02]$ & $\mathbf{0.59}[0.02]$\\
        (convergence &\textbf{Manipulation} & $0.41[0.04]$ & $0.50[0.10]$ & $0.69[0.06]$ & $0.71[0.08]$ & $0.49[0.06]$ & $0.38[0.02]$ & $0.71[0.04]$ & $\mathbf{0.77}[0.02]$\\
        speed) &\textbf{Driving} & $0.38[0.02]$ & $0.41[0.04]$ & $0.52[0.02]$ & $0.51[0.03]$ & $0.32[0.02]$ & $0.61[0.02]$ & $0.60[0.03]$ & $\mathbf{0.65}[0.02]$\\

        &\cellcolor{blue!15}\textbf{All} 
        &\cellcolor{blue!15} $0.46[0.03]$
        &\cellcolor{blue!15} $0.47[0.06]$
        &\cellcolor{blue!15} $0.55[0.04]$
        &\cellcolor{blue!15} $0.50[0.05]$ 
        &\cellcolor{blue!15} $0.47[0.04]$ 
        &\cellcolor{blue!15} $0.50[0.04]$ 
        &\cellcolor{blue!15} $0.65[0.04]$ 
        &\cellcolor{blue!15} $\mathbf{0.72}[0.03]$\\
    \end{tabular}
    }
    \caption{n-score and n-AUC results. 
    Margins in brackets are computed by averaging the standard deviations (across 3 random seeds) of individual tasks.
    The last two shaded columns are our methods.
    }
    \label{tab:results}
\vspace{-5ex}
\end{table}

\subsection{Results and observations}
\label{exp:results}
The n-AUC and final n-score values are shown in Table~\ref{tab:results}, and the training curves of n-score are shown in Figure~\ref{fig:overall_curves}.
First of all, we conclude that the tasks are diverse enough to reduce the evaluation variance. 
The averaged standard deviations are small for most methods. 
Thus the comparison results are not coincidences and are likely to generalize to other scenarios. 
Overall, \name largely outperforms the 6 baselines regarding both final n-score ($0.96[0.03]$ \textit{vs.} second-best $0.78[0.05]$) and n-AUC ($0.72[0.03]$ \textit{vs.} second-best $0.55[0.04]$), with relatively small standard deviations.
Note that the n-AUC gap naturally tends to be smaller than the final n-score gap because the former reflects a convergence trend throughout the training.
Moreover, \name achieved top performance of n-AUC on each individual task category.
Although some baselines achieved best final n-scores by slightly dominating \name, their performance is not consistent over all task categories.
More observations can be made below.

\begin{compactenum}[-]
\item\emph{Persistent exploration and the compare-through operator are both crucial.}\ \ Even with one-step TD, \name-1td's performance ($0.92[0.05]$ and $0.65[0.04]$) already outperforms the baselines.
        This shows that persistent exploration alone helps much.
        Furthermore, \name is generally better than \name-1td ($0.96[0.03]$ \textit{vs.} $0.92[0.05]$  and $0.72[0.03]$ \textit{vs.} $0.65[0.04]$).
        This shows that efficient policy evaluation by the compare-through operator is also a key component of \name.

%
\iffalse
\item\textit{Repeating with predicted duration lengths}\ \ With predicted duration of variable lengths, SAC-Krep is sometimes (\textbf{Locomotion}, \textbf{Manipulation}, and \textbf{Driving}) better than 
        SAC-Nrep in terms of final n-scores thanks to its flexibility.
        %
        However, SAC-Krep usually converges slower than SAC-Nrep because of additional multiple Q value heads to be learned for predicting 
        repetition steps.
        %
        %As a result, with fewer environment frames such as in \textbf{SimpleControl} and \textbf{Terrain}, SAC-Krep hardly 
        %catches up SAC-Nrep at the end of training.
        %
        %While \name is also flexible in repeating, it does not learn additional value heads (Eq.~\ref{eq:optimal_beta}).
        %
        %Thus it is much more efficient than SAC-Krep. 
        Importantly, the closed-loop repetition of \name yields better final n-scores than the open-loop repetition of 
        SAC-Krep in most task categories.
\fi
%
\item\emph{A proper formulation of closed-loop action repetition is important.}\ \ The idea of closed-loop action repetition alone is not a magic ingredient, as SAC-Hybrid only has moderate performance among the baselines.
    Notably, it performs worst on \textbf{Locomotion} and \textbf{Manipulation}.
    Our analysis revealed that its ``act-or-repeat`` policy tends to repeat with very high probabilities on \textit{Ant}, \textit{FetchPickAndPlace}, and \textit{FetchPush} even towards the end of training, resulting in very poor n-scores.
    All these three tasks feature hard exploration.
    %\footnote{Although \textit{Ant} has densely shaped rewards, a policy lacking exploration could easily gets stuck in a local minimum where the ant stands stationarily to maximize the alive bonus rewards.}.
    %
    This result suggests that a good formulation of the idea is crucial.
    The two-stage decision $\pi(\hat{a}|s,a^-)\pi(b|s,a^-,\hat{a})$ of \name is clearly more informed than the parallel decisions $\pi(\hat{a}|s,a^-)\pi(b|s,a^-)$ of SAC-Hybrid. 
    Furthermore, even with a latent ``act-or-repeat`` action, \name manages to maintain the complexity of the Q function $Q(s,a)$ as in the original control problem, while SAC-Hybrid has a more complex form $Q((s,a^-),(\hat{a},b))$ (Appendix~\ref{app:sac-hybrid}).

    \iffalse
    \item \textbf{Unbiased multi-step Q operator}\ \ Our earlier observation of multi-step TD revealed that 
        \name-Ntd (with a biased Q operator) is almost always better than \name-1td.
        %
        We can further boost the performance (\name \vs\ \name-Ntd) using the unbiased multi-step operator $\mathcal{T}^{\pi}$ in Section~\ref{sec:policy_eval}.
        %
        The result shows that it helps in \textbf{SimpleControl} and \textbf{Manipulation}.
        %
        However, for the other three categories, \name-Ntd is on par with \name.
        %
        We hypothesize that this is due to $N$ being small (3 or 5) to accommodate SAC-\{K,N\}rep which will perform poorly otherwise (due to open-loop control),
        and expect the off-policyness in \name-Ntd to be more as $N$ increases.
        %
        To verify this, we increased $N$ to 10 and re-trained \name and \name-Ntd on the three categories.
        %
        The final n-scores are listed in Table~\ref{tab:n10} and training curves are in Figure~\ref{fig:n10} Appendix~\ref{app:more-results}.
        %
        We can see that as $N$ increases, \name-Ntd suffers more while \name does not.
    \fi
    
%
\item\emph{Naive action repetition works well.}\ \ Interestingly, in this particular experiment, SAC-Nrep is the overall top-performing baseline due to its relatively balanced results on all task categories.
        %    
        %It is surprisingly effective in some scenarios.
        %
        %When there is large room for temporal abstraction and the task does not require
        %fine-grained control, SAC-Nrep can be much better than SAC. 
        %
        However, when it fails, the performance could be very bad (\textbf{Locomotion} and \textbf{Terrain}).
        While its sample efficiency is good, it has a \emph{difficulty of approaching the optimal control} (final mean n-scores $\le 0.92$ in individual task categories) due to its lack of flexibility in the action repeating duration.
        This suggests that action repetition greatly helps, but a fixed duration is difficult to pick.

\item\emph{Limitation: action repetition hardly benefits tasks with densely shaped rewards and frame skipping.}\ \ We notice that \name is no better than SAC and SAC-EZ on \textbf{Locomotion} regarding the final n-score, although it has slight advantages of n-AUC.
    The locomotion tasks have densely shaped rewards to guide policy search.
    Thus action repetition hardly helps locomotion exploration, especially when the 4 tasks already have built-in frameskips
    (4 frames for \task{Hopper} and \task{Walker2d}; 5 frames for \task{Ant} and \task{HalfCheetah}).
    %
    %This observation confirms the finding by \citet{Neunert2020}.
    %
    We believe that more sophisticated temporal abstraction (\eg\ skills) is needed to improve the performance in this case.
    %
    %However, that \name is comparable to SAC demonstrates its robustness.
%
\end{compactenum}
%
%Combining all observations above together, the results have justified our choices of using a second-stage $\beta$ policy for 
%closed-loop action repetition, and using an unbiased Q operator conditioned on $\beta$'s output.
%
%Next, with visualization we will further analyze the importance of persistent exploration induced by action repetition.

%\begin{wrapfigure}{r}{0.6\textwidth}
\begin{table}[t!]
\newcolumntype{?}{!{\vrule width 2pt}}
\newcolumntype{a}{>{\columncolor{blue!15}}c}
\setlength\tabcolsep{2pt}
\centering
\resizebox{\textwidth}{!}{
\begin{tabular}{@{}l|c|c|c|c|c|a|c|c|c|c|c|a@{}}
    & \multicolumn{6}{c|}{Final n-score (model quality)} & \multicolumn{6}{c}{n-AUC (convergence speed)}\\
    & \textbf{S} & \textbf{L} & \textbf{T} & \textbf{M} & \textbf{D} & \textbf{All}
    & \textbf{S} & \textbf{L} & \textbf{T} & \textbf{M} & \textbf{D} & \textbf{All}\\
    \hline
    \name-Ntd ($N=10$) & $0.88[0.16]$ & $0.73[0.18]$ & $0.79[0.06]$ & $0.82[0.07]$ & $0.72[0.01]$ & $0.79[0.12]$ 
                       & $0.56[0.11]$ & $0.54[0.12]$ & $0.53[0.03]$ & $0.62[0.04]$ & $0.38[0.06]$ & $0.55[0.08]$ \\
    \name ($N=10$) & $1.00[0.02]$ & $0.85[0.11]$ & $0.97[0.04]$ & $0.94[0.02]$ & $0.94[0.03]$ & $0.93[0.05]$ 
                   & $0.73[0.01]$ & $0.67[0.06]$ & $0.59[0.02]$ & $0.73[0.03]$ & $0.60[0.00]$ & $0.68[0.03]$\\
    \hline
    SAC-Hybrid & $0.88[0.14]$ & $0.63[0.08]$ & $0.79[0.03]$ & $0.49[0.07]$ & $1.00[0.04]$ & $0.69[0.08]$ 
               & $0.62[0.08]$ & $0.50[0.06]$ & $0.50[0.01]$ & $0.38[0.02]$ & $0.61[0.02]$ & $0.50[0.04]$\\
    SAC-Hybrid-CompThr & $0.89[0.12]$ & $0.65[0.08]$ & $0.75[0.03]$ & $0.68[0.17]$ & $0.90[0.04]$ & $0.74[0.11]$ 
                      & $0.58[0.08]$ & $0.48[0.06]$ & $0.47[0.01]$ & $0.47[0.09]$ & $0.49[0.01]$ & $0.50[0.06]$\\
\end{tabular}}
\caption{Off-policyness experiments results.
Error margins inside brackets are computed by averaging the standard deviations (across 3 random seeds) of individual tasks in a category.
\textbf{S}: \textbf{SimpleControl}; \textbf{L}: \textbf{Locomotion}; \textbf{T}: \textbf{Terrain}; \textbf{M}: \textbf{Manipulation}; 
\textbf{D}: \textbf{Driving}.}
\label{tab:off-policyness}
\end{table}
%\end{wrapfigure}

\subsection{Off-policyness experiments}
\label{sec:off-policy-exp}
To verify that our compare-through operator is not affected by off-policyness in policy evaluation, we compare \name to a variant \name-Ntd which always bootstraps a Q value with an $N$-step target, regardless of $\beta$'s outputs and without importance correction\footnote{We cannot apply Retrace to \name-Ntd since the probability density of $\pi^{\text{ta}}$ is computationally intractable.}.
We choose a large trajectory length $N=10$ to amplify the effect of off-policyness.
Table~\ref{tab:off-policyness} shows that $N$-step TD without importance correction significantly degrades the performance ($0.79[0.12]$ \textit{vs.} $0.93[0.05]$ and $0.55[0.08]$ \textit{vs.} $0.68[0.03]$).
In contrast, the compare-through operator well addresses this issue for \name.

Furthermore, we implement a variant of SAC-Hybrid by replacing the Retrace operator with our compare-through operator, to see if the result difference between \name and SAC-Hybrid is mainly due to different ways of handling off-policyness.
Table~\ref{tab:off-policyness} shows that SAC-Hybrid-CompThr performs similarly to SAC-Hybrid ($0.74[0.11]$ \textit{vs.} $0.69[0.08]$ and $0.50[0.06]$ \textit{vs.} $0.50[0.04]$), suggesting that it is indeed the formulation of SAC-Hybrid that creates its performance gap with \name.

\begin{wrapfigure}[14]{r}{0.45\textwidth}
\vspace{-6ex}
    \centering
    \begin{tabular}{@{}c@{}c@{}c@{}}
    \includegraphics[width=0.2\textwidth]{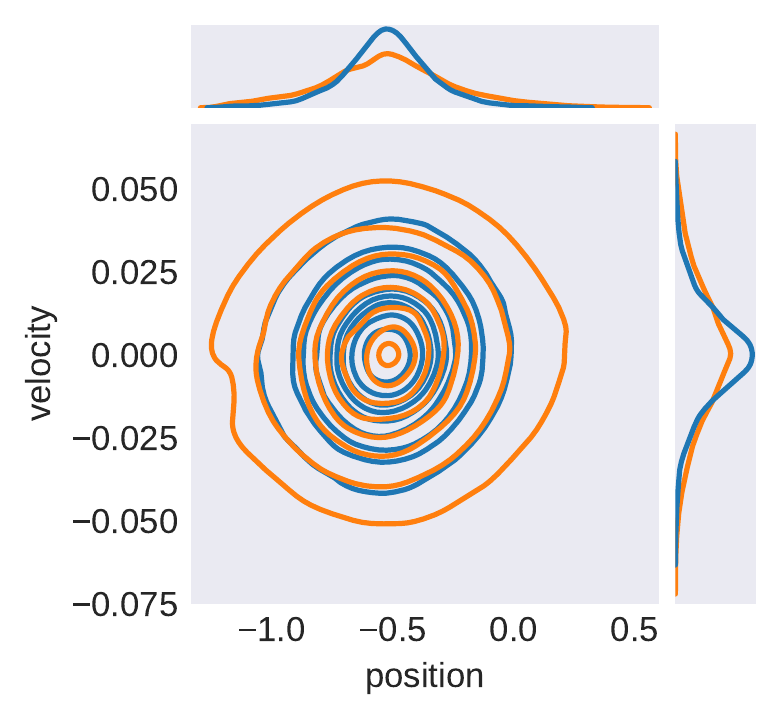}
    &\includegraphics[width=0.2\textwidth]{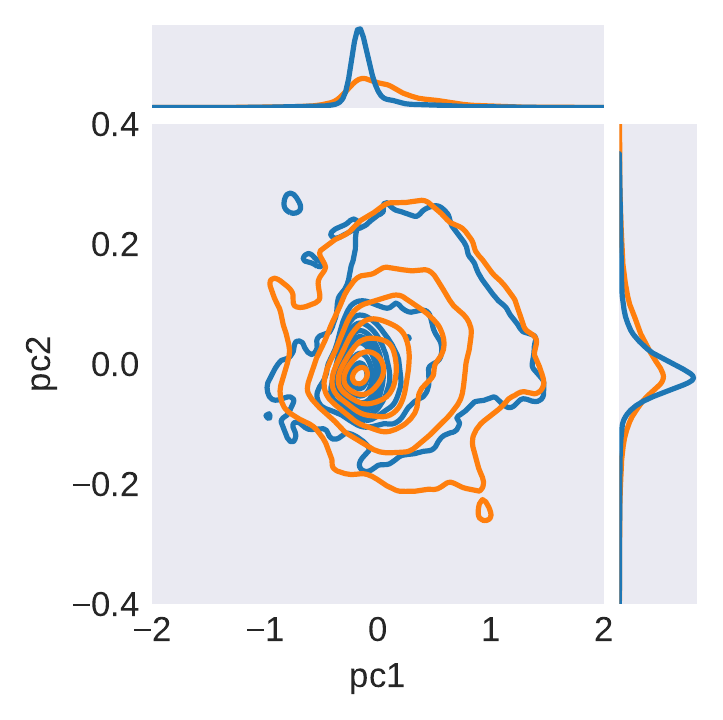}    
    &\includegraphics[width=0.05\textwidth]{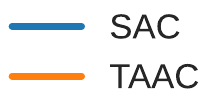}\\
    \end{tabular}
    \caption{KDE plots of $100$K state vectors visited by \name and SAC with their randomized policies.
    The side plots on top and right represent 1D marginal density functions.
    Left: \task{MountainCarContinuous}: $x$ is the car's position and $y$ is the car's velocity; 
    Right: \task{BipedalWalker}: $xy$ represent the top-2 principal components of the walker's hull.
    }
\label{fig:kde}
\end{wrapfigure}

\subsection{Policy behavior visualization and analysis}
\label{sec:visualization}
In this section, we mainly answer two questions: 
\begin{compactenum}[1)]
    \item How is the exploration behavior of \name compared to that of SAC, a ``flat'' RL algorithm?
    \item What is the action repetition behavior of \name in a trained control task?
\end{compactenum}

\subsubsection{Exploration behavior}
\name introduces persistent exploration along previous actions, having a better chance
of escaping the local neighborhood around a state when acting randomly.
Thus \name's exploration should yield a better state space coverage 
than SAC's does, assuming other identical conditions.
%\footnote{In fact, any method with temporal abstraction should have this exploration property. 
%
%In this section we only focus on examining our method.}.
%
To verify this, we visualize the state space coverage by SAC and \name during their initial exploration phases.

Specifically, we select two tasks \task{MountainCarContinuous} and \task{BipedalWalker} for this purpose.
For either \name or SAC, we play a random version of its policy on either task for $50$K environment frames, to simulate the initial 
exploration phase where the model parameters have not yet been altered by training.
During this period, we record all the $50$K state vectors for analysis.
For \task{MountainCarContinuous}, each state vector is 2D, representing the car's ``x-position'' and ``x-velocity'' on the 1D track.
For \task{BipedalWalker}, each state vector is 24D, where the first 4D sub-vector indicates the statistics of the walker's hull: ``angle'', ``angular velocity'', ``x-velocity'', and ``y-velocity''.
For visualization, we first extract this 4D sub-vector and form a combined dataset of $100$K vectors from both \name and SAC.
Then we apply PCA \citep{Jolliffe1986} to this dataset and project each 4D vector down to 2D.
After this, we are able to draw KDE (kernel density estimate) plots for both \task{MountainCarContinuous} and \task{BipedalWalker} 
in Figure~\ref{fig:kde}.
We see that on both tasks, a random policy of \name is able to cover more diverse states compared to SAC.
This suggests that in general, \name is better at exploration compared to a ``flat'' RL method, thanks to its ability of persistent exploration.

\subsubsection{Action repetition behavior}
\label{sec:repeat_behavior}
In theory, to achieve optimal continuous control, the best policy should always sample a new action at every step and avoid 
action repetition at all.
Thus one might assume that \name's second-stage policy shifts from frequently repeating actions in the beginning of training, to 
not repeating actions at all towards the end of training, in order to optimize the environment reward.
However, some examples in Figure~\ref{fig:repetition_examples} suggest that a significant repetition frequency
still exists in \name's top-performing evaluation episodes.
Note that because we take the mode of the switching policy during evaluation (Appendix~\ref{app:eval}), this suggests that for those repeating steps, we have $\beta^*_0 > \beta^*_1$ by the trained model.
In fact, if we evaluate 100 episodes for each of the three tasks in Figure~\ref{fig:repetition_examples} and compute the action repetition percentage (repeating steps divided by total steps), the percentages are surprisingly 89\%, 39\%, and 49\%, even though the agents are doing extremely well!
A complete list of action repetition percentages for all 14 tasks can be found in Table~\ref{tab:repeat_percent}.
Generally, \name is able to adjust the repetition frequency according to tasks, resulting in a variety of frequencies across the tasks.
More importantly, if we inspect the repeated actions, they are often not even close to action boundaries $\{a_{min},a_{max}\}$, ruling out the possibility of the policy being forced to repeat due to action clipping/squashing. 

\begin{table}[t!]
\setlength\tabcolsep{2pt}
\centering
\resizebox{0.95\textwidth}{!}{
\begin{tabular}{l|c|c|c|c|c|c|c|c|c|c|c|c|c|c}
\textbf{Tasks} & \textit{MCC} & \textit{LLC} & \textit{IDP} & \textit{HOP} & \textit{ANT} 
& \textit{WAL} & \textit{HC} & \textit{BW} & \textit{BWH} & \textit{FR} & \textit{FP} & \textit{FS} & \textit{FPP} & \textit{TOW}\\
\hline
\textbf{Action repetition percentage} & 89\% & 74\% & 26\% & 37\% & 13\% & 26\% & 1\% & 25\% & 39\% & 9\% & 55\% & 54\% & 49\% & 55\%\\
\end{tabular}}
\caption{Action repetition percentages by evaluating a trained \name model for 100 episodes on each of the 14 tasks. 
Refer to Section~\ref{sec:tasks} for the full task names.
}
\label{tab:repeat_percent}
\end{table}

We believe that there are two reasons for this surprising observation.
First, with many factors such as function approximation, noise in the estimated gradient, and stochasticity of the environment dynamics, it is hardly possible for a model to reach the theoretical upper bound of an RL objective. 
Thus for a new action and the previous action, if their estimated Q values are very similar, then \name might choose to repeat with a certain chance. 
For example in Figure~\ref{fig:repetition_examples} (c), while the robot arm is lifting the object towards the goal in the air, it can just repeat the same lifting action for 3 steps (at $t+5$), without losing the optimal return (up to some estimation error).
Similar things happen to the bipedal walker (b) when it is jumping in the air (9 steps repeated at $t+7$), and to the mountain car (a) when 
it is rushing down the hill (11 steps repeated at $t+26$).
Second, as a function approximator, the policy network $\pi_{\thetaA}$ have a limited capacity, and it is not able to represent the 
optimal policy at every state in a continuous space.
For non-critical states that can be handled by repeating previous actions, \emph{\name might learn to offload the decision making of $\pi_{\thetaA}$ onto $\beta$, and save $\pi_{\thetaA}$'s representational power for critical states}.
For example, the robot arm in Figure~\ref{fig:repetition_examples} (c) invokes a fine-grained control by $\pi_{\thetaA}$ when it's grasping the object from the table, while later does not invoke $\pi_{\thetaA}$ for lifting it.
%
%The actual reason of action repetition might be more complex, perhaps a combination of the two reasons above.

\begin{figure}[!t]
\vspace{-2ex}
    \setlength{\tabcolsep}{1pt}
    \renewcommand{\arraystretch}{0.5}
    \centering
    
    \begin{tabular}{@{}c@{}c@{}c@{}}
    
    \resizebox{0.33\textwidth}{!}{
    \begin{tabular}{@{}cccccc@{}}
    $t$ & $t+13$ & $t+21$ & $t+24$ & $t+26$ & $t+37$\\\\
    \begin{tabular}{@{}c@{}}
    \includegraphics[width=0.1\textwidth]{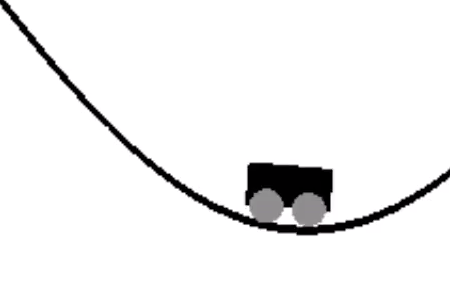}\\
    \includegraphics[width=0.1\textwidth]{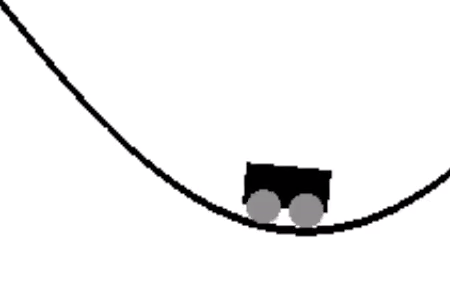}\\
    $\myvdots \ (10)$\\
    \includegraphics[width=0.1\textwidth]{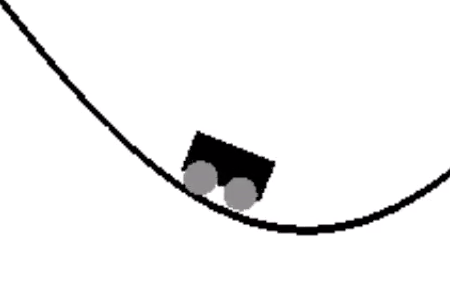}\\
    \end{tabular}
    &
    \begin{tabular}{@{}c@{}}
    \includegraphics[width=0.1\textwidth]{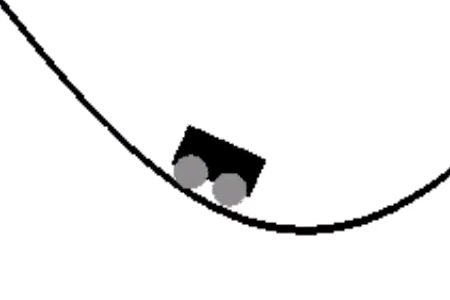}\\
    \includegraphics[width=0.1\textwidth]{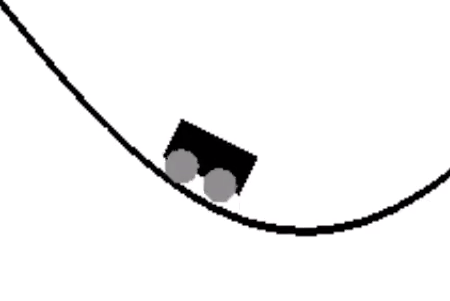}\\
    $\myvdots \ (5)$\\    
    \includegraphics[width=0.1\textwidth]{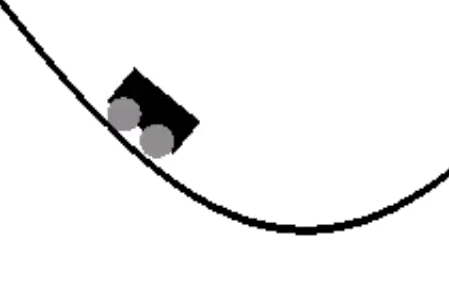}\\    
    \end{tabular}
    &
    \begin{tabular}{@{}c@{}}
    \includegraphics[width=0.1\textwidth]{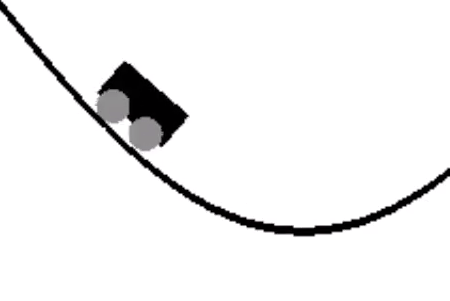}\\
    \includegraphics[width=0.1\textwidth]{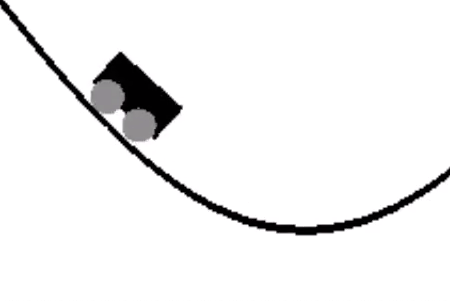}\\
    \includegraphics[width=0.1\textwidth]{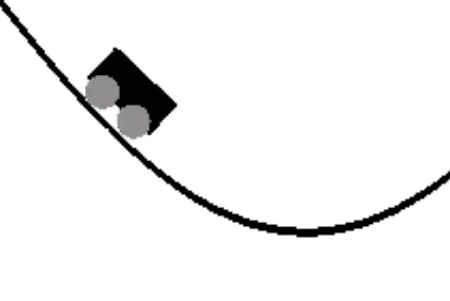}\\    
    \end{tabular}
    &
    \begin{tabular}{@{}c@{}}
    \includegraphics[width=0.1\textwidth]{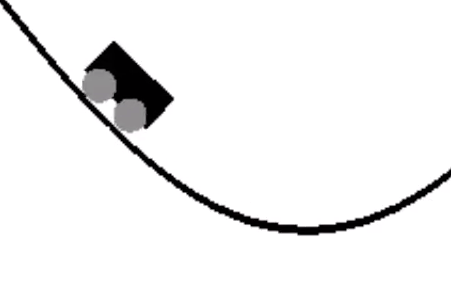}\\
    \includegraphics[width=0.1\textwidth]{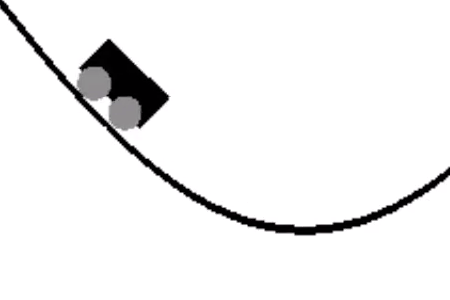}\\    
    \end{tabular}
    &
    \begin{tabular}{@{}c@{}}
    \includegraphics[width=0.1\textwidth]{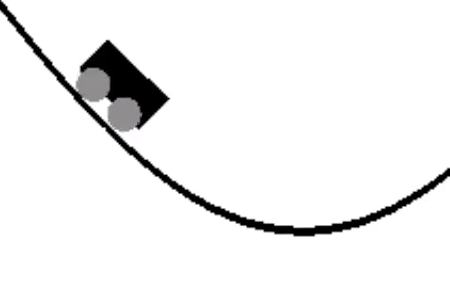}\\
    \includegraphics[width=0.1\textwidth]{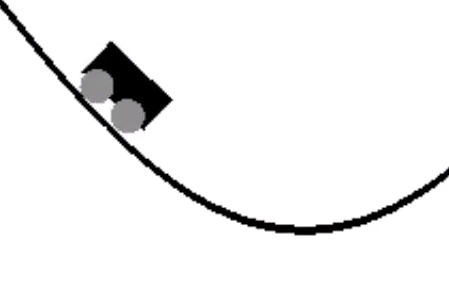}\\    
    $\myvdots \ (8)$\\    
    \includegraphics[width=0.1\textwidth]{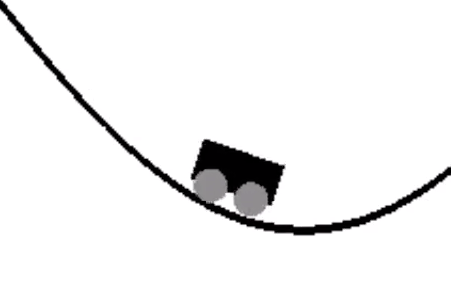}\\        
    \end{tabular}
    &
    \begin{tabular}{@{}c@{}}
    \includegraphics[width=0.1\textwidth]{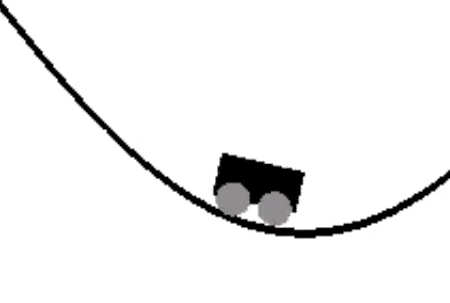}\\
    \end{tabular}\\
    \end{tabular}
    }
    
    &
    
    \resizebox{0.33\textwidth}{!}{
    \begin{tabular}{@{}cccccc@{}}
    $t$ & $t+6$ & $t+7$ & $t+16$ & $t+17$ & $t+18$\\\\
    \begin{tabular}{@{}c@{}}
    \includegraphics[width=0.1\textwidth]{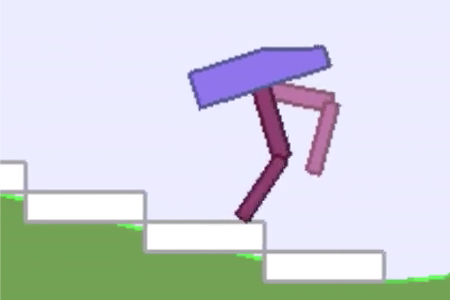}\\
    \includegraphics[width=0.1\textwidth]{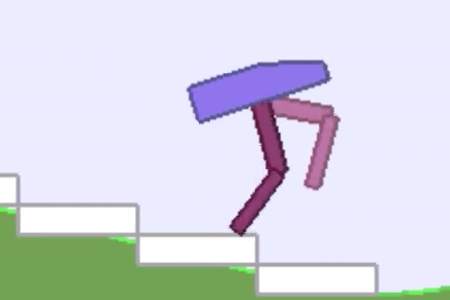}\\
    $\myvdots \ (3)$\\
    \includegraphics[width=0.1\textwidth]{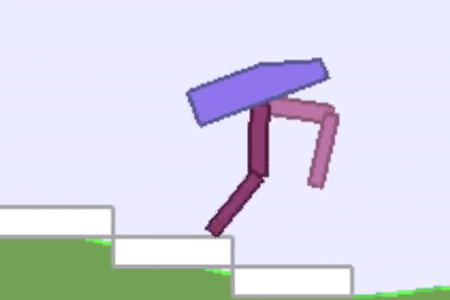}\\
    \end{tabular}
    &
    \begin{tabular}{@{}c@{}}
    \includegraphics[width=0.1\textwidth]{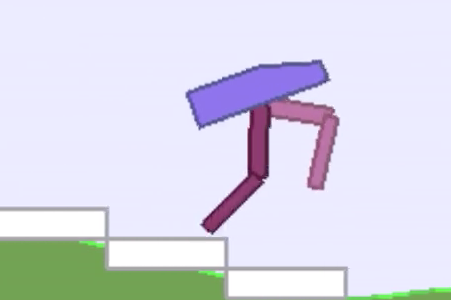}\\
    \end{tabular}
    &
    \begin{tabular}{@{}c@{}}
    \includegraphics[width=0.1\textwidth]{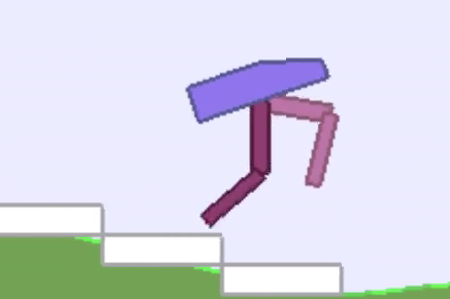}\\
    \includegraphics[width=0.1\textwidth]{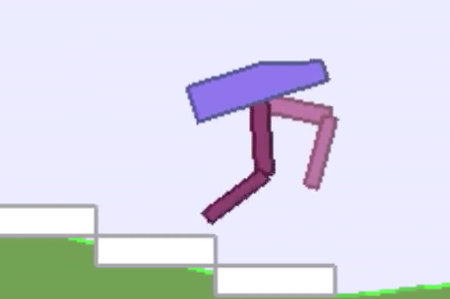}\\
    $\myvdots \ (6)$\\
    \includegraphics[width=0.1\textwidth]{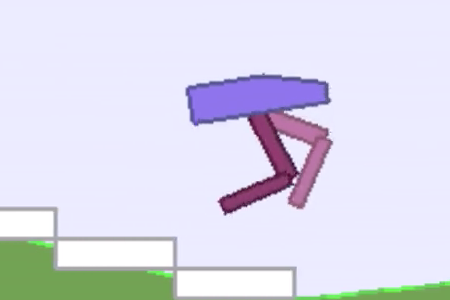}\\    
    \end{tabular}
    &
    \begin{tabular}{@{}c@{}}
    \includegraphics[width=0.1\textwidth]{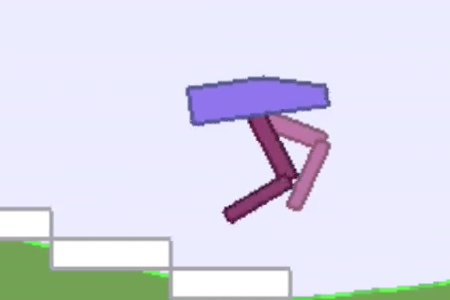}\\
    \end{tabular}
    &
    \begin{tabular}{@{}c@{}}
    \includegraphics[width=0.1\textwidth]{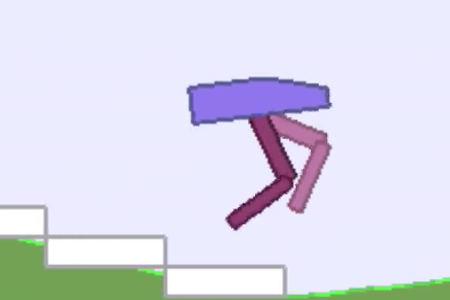}\\
    \end{tabular}
    &
    \begin{tabular}{@{}c@{}}
    \includegraphics[width=0.1\textwidth]{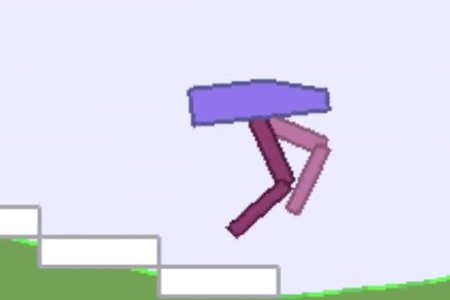}\\
    \includegraphics[width=0.1\textwidth]{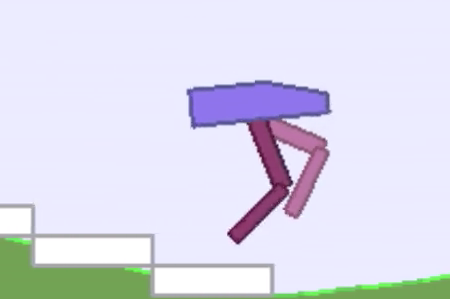}\\
    \end{tabular}\\
    \end{tabular}
    }
    
    &
    
    \resizebox{0.33\textwidth}{!}{
    \begin{tabular}{@{}cccccc@{}}
    $t$ & $t+1$ & $t+2$ & $t+3$ & $t+4$ & $t+5$\\\\
    \begin{tabular}{@{}c@{}}
    \includegraphics[width=0.1\textwidth]{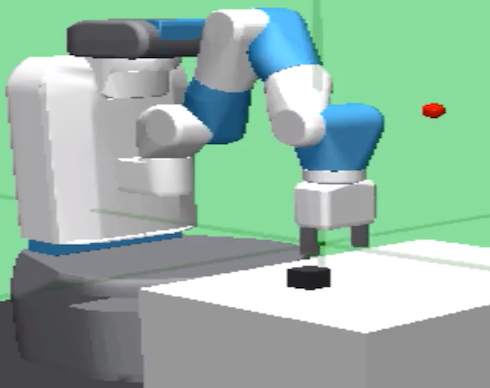}\\
    \end{tabular}
    &
    \begin{tabular}{@{}c@{}}
    \includegraphics[width=0.1\textwidth]{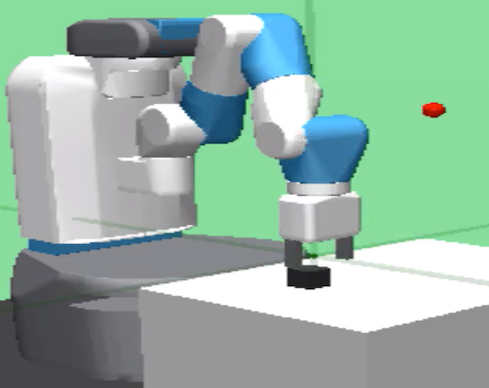}\\
    \end{tabular}
    &
    \begin{tabular}{@{}c@{}}
    \includegraphics[width=0.1\textwidth]{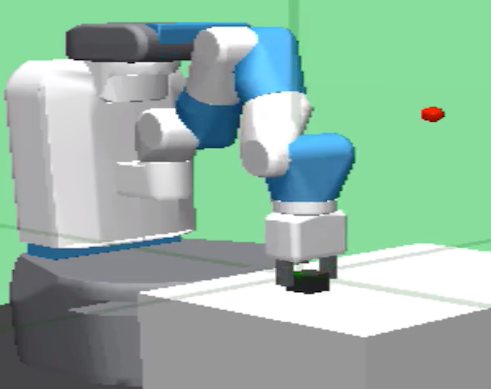}\\
    \end{tabular}
    &
    \begin{tabular}{@{}c@{}}
    \includegraphics[width=0.1\textwidth]{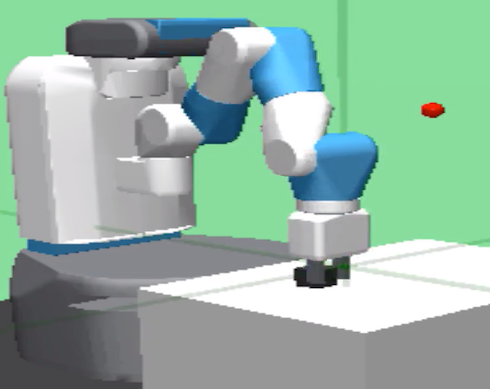}\\
    \end{tabular}
    &
    \begin{tabular}{@{}c@{}}
    \includegraphics[width=0.1\textwidth]{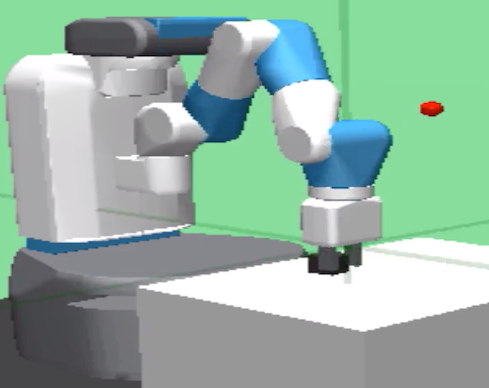}\\
    \end{tabular}
    &
    \begin{tabular}{@{}c@{}}
    \includegraphics[width=0.1\textwidth]{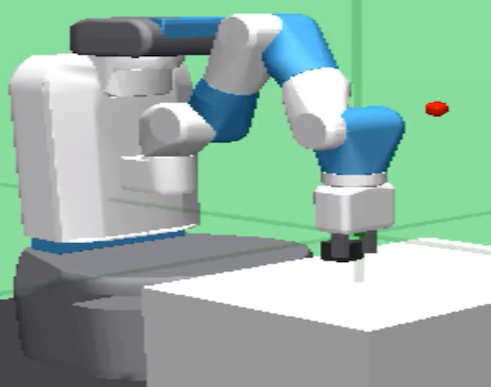}\\
    \includegraphics[width=0.1\textwidth]{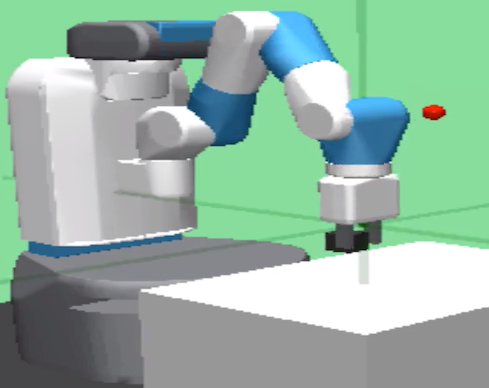}\\
    \includegraphics[width=0.1\textwidth]{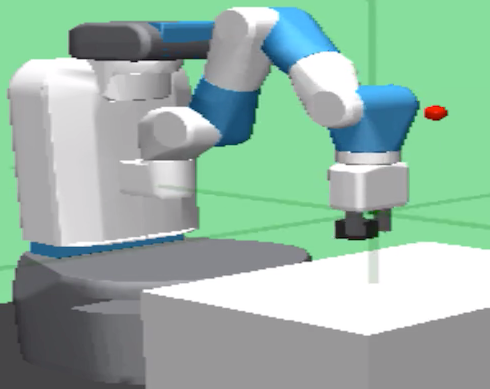}\\
    \end{tabular}\\
    \end{tabular}
    }
    \\
    (a) & (b) & (c)\\
    
    \end{tabular}
    \caption{Example frames (cropped) from top-performing evaluation episodes of \name.
    Each column contains consecutive frame(s) generated by the same action, where
    `` $\myvdots \ (n)$'' denotes $n$ similar frames omitted due to space limit.
    %
%    Each index represents the time of the first frame in a column.
    %
    (a): In \task{MountainCarContinuous} the car first backs up to build gravitational potential and rushes down;
    (b): In \task{BipedalWalkerHardcore} the bipedal walker jumps one step down; 
    (c): In \task{FetchPickAndPlace} the robot arm approaches the object and lifts it to a goal location.
    }
\label{fig:repetition_examples}
\vspace{-1ex}
\end{figure}

\section{Conclusion}
\vspace{-1ex}
We have proposed \name, a simple but effective off-policy RL algorithm that is a middle ground between ``flat'' and hierarchical RL.
\name incorporates closed-loop temporal abstraction into actor-critic by adding a second-stage policy that chooses between the
previous action and a new action output by an actor.
%
%\name encourages persistent exploration and natively supports multi-step TD backup with a repeat-through Q operator.
%
\name yielded strong empirical results on a variety of continuous control tasks, outperforming prior works that also model action repetition.
The evaluation and visualization revealed the success factors of \name: persistent exploration and a compare-through Q operator for multi-step TD backup.
We believe that our work has provided valuable insights into modeling temporal abstraction and action hierarchies for solving complex RL tasks in the future.

%\ifcsname usepreprint\endcsname
%\else
\textbf{Societal impact}\ \ This paper proposes a general algorithm to improve the efficiency of RL for continuous control. 
The algorithm can be applied to many robotics scenarios in the real world.
How to address the potentially unsafe behaviors and risks of the deployed algorithm caused to the surroundings in real-world scenarios remains an open problem and requires much consideration.
%\fi

% Acknowledgements should only appear in the accepted version.

\ifcsname usepreprint\endcsname
\section*{Acknowledgements}
The authors would like to thank Jerry Bai and Le Zhao for helpful discussions on this project, and the Horizon AI platform team for infrastructure support.
\fi

\bibliography{main}
\bibliographystyle{apalike}

\ifcsname usepreprint\endcsname
\else
%%%%%%%%%%%%%%%%%%%%%%%%%%%%%%%%%%%%%%%%%%%%%%%%%%%%%%%%%%%%
\section*{Checklist}

\iffalse
%%% BEGIN INSTRUCTIONS %%%
The checklist follows the references.  Please
read the checklist guidelines carefully for information on how to answer these
questions.  For each question, change the default \answerTODO{} to \answerYes{},
\answerNo{}, or \answerNA{}.  You are strongly encouraged to include a {\bf
justification to your answer}, either by referencing the appropriate section of
your paper or providing a brief inline description.  For example:
\begin{itemize}
  \item Did you include the license to the code and datasets? \answerYes{See Section~\ref{gen_inst}.}
  \item Did you include the license to the code and datasets? \answerNo{The code and the data are proprietary.}
  \item Did you include the license to the code and datasets? \answerNA{}
\end{itemize}
Please do not modify the questions and only use the provided macros for your
answers.  Note that the Checklist section does not count towards the page
limit.  In your paper, please delete this instructions block and only keep the
Checklist section heading above along with the questions/answers below.
%%% END INSTRUCTIONS %%%
\fi

\begin{enumerate}

\item For all authors...
\begin{enumerate}
  \item Do the main claims made in the abstract and introduction accurately reflect the paper's contributions and scope?
    \answerYes{}
  \item Did you describe the limitations of your work?
    \answerYes{} In the experiment results section
  \item Did you discuss any potential negative societal impacts of your work?
    \answerYes{} In the conclusion section.
  \item Have you read the ethics review guidelines and ensured that your paper conforms to them?
    \answerYes{}
\end{enumerate}

\item If you are including theoretical results...
\begin{enumerate}
  \item Did you state the full set of assumptions of all theoretical results?
    \answerYes{}
	\item Did you include complete proofs of all theoretical results?
    \answerYes{} Appendix
\end{enumerate}

\item If you ran experiments...
\begin{enumerate}
  \item Did you include the code, data, and instructions needed to reproduce the main experimental results (either in the supplemental material or as a URL)?
    \answerNo{} The code will reveal author identities. We will release the code and instructions if the paper is accepted.
  \item Did you specify all the training details (e.g., data splits, hyperparameters, how they were chosen)?
    \answerYes{} Appendix
	\item Did you report error bars (e.g., with respect to the random seed after running experiments multiple times)?
    \answerYes{} We reported standard deviations in the experiment.
	\item Did you include the total amount of compute and the type of resources used (e.g., type of GPUs, internal cluster, or cloud provider)?
    \answerYes{} Appendix H.3
\end{enumerate}

\item If you are using existing assets (e.g., code, data, models) or curating/releasing new assets...
\begin{enumerate}
  \item If your work uses existing assets, did you cite the creators?
    \answerYes{} See the tasks section for the simulators we use.
  \item Did you mention the license of the assets?
    \answerNo{} Licenses have been included in the original simulator papers/urls/websites.
  \item Did you include any new assets either in the supplemental material or as a URL?
    \answerNo{} Our new asset only contains our source code. We will release it if the paper is accepted, for the reason stated above.
  \item Did you discuss whether and how consent was obtained from people whose data you're using/curating?
    \answerNA{} The data/simulators are all virtual.
  \item Did you discuss whether the data you are using/curating contains personally identifiable information or offensive content?
    \answerNA{} The data/simulators are all virtual.
\end{enumerate}

\item If you used crowdsourcing or conducted research with human subjects...
\begin{enumerate}
  \item Did you include the full text of instructions given to participants and screenshots, if applicable?
    \answerNA{}
  \item Did you describe any potential participant risks, with links to Institutional Review Board (IRB) approvals, if applicable?
    \answerNA{}
  \item Did you include the estimated hourly wage paid to participants and the total amount spent on participant compensation?
    \answerNA{}
\end{enumerate}

\end{enumerate}

%%%%%%%%%%%%%%%%%%%%%%%%%%%%%%%%%%%%%%%%%%%%%%%%%%%%%%%%%%%%

\fi

%%%%%%%%%%%%%%%%%%%%%%%%%%%%%%%%%%%%%%%%%%%%%%%%%%%%%%%%%%%%%%%%%%%%%%%%%%%%%%%
%%%%%%%%%%%%%%%%%%%%%%%%%%%%%%%%%%%%%%%%%%%%%%%%%%%%%%%%%%%%%%%%%%%%%%%%%%%%%%%
% DELETE THIS PART. DO NOT PLACE CONTENT AFTER THE REFERENCES!
%%%%%%%%%%%%%%%%%%%%%%%%%%%%%%%%%%%%%%%%%%%%%%%%%%%%%%%%%%%%%%%%%%%%%%%%%%%%%%%
%%%%%%%%%%%%%%%%%%%%%%%%%%%%%%%%%%%%%%%%%%%%%%%%%%%%%%%%%%%%%%%%%%%%%%%%%%%%%%%
\clearpage

\appendix

\section{\name pseudo code}
\label{app:algorithm}

\begin{algorithm}[h!]
        \SetAlgoLined
        \textbf{Input:} $\thetaQ$, $\thetaA$, $\lambda$ (learning rate), and $\tau$ (moving average rate)\\
        \textbf{Initialize:} Randomize $\thetaQ$ and $\thetaA$, $\bar{\thetaQ}\leftarrow\thetaQ$, $\mathcal{D}\leftarrow \emptyset$\\
        \For{each training iteration}{
            \For{each rollout step} {
%                \vspace{-2ex}
                %\begin{equation*}
                    %\hat{a} \sim
                    %\Big\{ 
                    %\begin{array}{ll}
                    %    \pi_{\thetaA}(\hat{a}|s,a^-), & \text{continuous action}\\
                    %    \pi^*(\hat{a}|s,a^-)\ \  \text{(Eq.~\ref{eq:optimal_a})}, & \text{discrete action}\\
                    %\end{array}
                    %\vspace{-2ex}
                %\end{equation*}\\
                $\hat{a} \sim  \pi_{\thetaA}(\hat{a}|s,a^-)$ \algorithmiccomment{first-stage policy}\\
                $b\sim \beta^*_b$ (Eq.~\ref{eq:optimal_beta}) \algorithmiccomment{second-stage policy}\\
                $a\leftarrow a^-$ if $b=0$ else $a\leftarrow \hat{a}$\\
                $s'\sim \mathcal{P}(s'|s,a)$\\
                $\mathcal{D}\leftarrow\mathcal{D}\bigcup \{(a^-,s,b,a,s',r(s,a,s'))\}$\\
                $(s,a^-)\leftarrow (s',a)$
            }
            \For{each gradient step}{
                $\thetaQ\leftarrow \thetaQ - \lambda \Delta\thetaQ$ (gradient of Eq.~\ref{eq:policy_eval} with the compare-through $\mathcal{T}^{\pi^{\text{ta}}}$) \algorithmiccomment{policy evaluation}\\
                $\thetaA\leftarrow \thetaA + \lambda\Delta\thetaA$ (Eq.~\ref{eq:policy_gradient}) \algorithmiccomment{policy improvement}\\
                $\alpha \leftarrow \alpha - \lambda\Delta\alpha$ (gradient of Eq.~\ref{eq:alpha}, $\alpha=\alpha', \alpha''$)  \algorithmiccomment{$\alpha$ adjustment}\\
                $\bar{\thetaQ}\leftarrow\bar{\thetaQ}+\tau(\thetaQ-\bar{\thetaQ})$ \algorithmiccomment{target network update}
            }
        }
        \textbf{Output:} $\thetaQ$ and $\thetaA$\\        
        \caption{Temporally abstract actor-critic}
    \label{alg:taac}
\end{algorithm}

\section{State value of the two-stage policy}
\label{app:state_value}
Let $P(a|s,\hat{a},a^-)\triangleq \beta_0\delta(a-a^-) + \beta_1\delta(a-\hat{a})$, we have
\begin{equation*}
\begin{array}{ll}
    V^{\pi^{\text{ta}}}_{\thetaQ}(s|a^-)&=\displaystyle\int_a \pi^{\text{ta}}(a|s,a^-)Q_{\thetaQ}(s,a) \text{d}a\\
    &=\displaystyle\int_a\left[\int_{\hat{a}}\pi_{\thetaA}(\hat{a}|s,a^-)P(a|s,\hat{a},a^-) \text{d}\hat{a}\right] Q_{\thetaQ}(s,a) \text{d}a\\
    &=\displaystyle\int_{\hat{a}}\pi_{\thetaA}(\hat{a}|s,a^-)\left[\int_a P(a|s,\hat{a},a^-)Q_{\thetaQ}(s,a)\text{d}a\right]\text{d}\hat{a}\\
    &=\displaystyle\int_{\hat{a}}\pi_{\thetaA}(\hat{a}|s,a^-)\left[\beta_0 Q_{\thetaQ}(s,a^-) + \beta_1 Q_{\thetaQ}(s,\hat{a})\right]\text{d}\hat{a}\\
    &=\displaystyle\expect_{\hat{a}\sim \pi_{\thetaA}, b\sim \beta}\left[(1-b)Q_{\thetaQ}(s,a^-) + bQ_{\thetaQ}(s,\hat{a})\right].\\
    \iffalse
    \text{where}
    \begin{array}{l}
        Q_{\thetaQ}(s,a^-,\hat{a},0) \triangleq Q_{\thetaQ}(s,a^-), \\
        Q_{\thetaQ}(s,a^-,\hat{a},1) \triangleq Q_{\thetaQ}(s,\hat{a}).\\
    \end{array}\\
    \fi
\end{array}
\end{equation*}

\section{Evaluating a policy by taking its (approximate) mode}
\label{app:eval}
An entropy-augmented training objective always maintains a pre-defined entropy level of the trained policy for exploration \citep{Haarnoja2018}. 
This results in stochastic behaviors and potentially lower scores if we measure the policy's rollout trajectories.
To address this randomness issue and reflect a method's actual performance, in the experiments we evaluate a method and compute its unnormalized scores by taking the (approximate) 
mode of its policy distribution. 
Following~\citet{Haarnoja2018}, we use a squashed diagonal Gaussian to represent a continuous policy for every comparison method.  
Specifically, when sampling an action, we first sample from the unsquashed Gaussian $z\sim \mathcal{N}(\mu,\sigma^2)$, and then apply 
the squashing function $x=a\cdot tanh(z) + b$ to respect the action boundaries $[b-a,b+a]$. 
However, because of the squashing effect, it's difficult to exactly obtain the mode of this distribution. 
So in practice, to approximately get the mode, we first get the mode $\mu$ from the unsquashed Gaussian, and then directly apply the squashing function 
$\tilde{\mu}=a\cdot tanh(\mu) + b$. 
This $\tilde{\mu}$ is treated as the (approximate) mode of the Gaussian policy. 

When evaluating TAAC, in the second stage of its two-stage policy, we also take the mode of the switching policy distribution $\beta$ as $\argmax_{b\in\{0,1\}}\beta_b$.

\section{Deriving the actor gradient}
\label{app:actor-gradient}
To maximize the objective in Eq.~\ref{eq:ent_v_opt} with respect to $\beta$, one can parameterize $\beta$ and use stochastic gradient ascent to adjust its parameters, similar to \citet{Neunert2020}. 
However, for every sampled $(s,a^-,\hat{a})$, there is in fact a closed-form solution for the inner expectation over $b\sim\beta$. 

In general, suppose that we have $N$ values $X(i)\in \mathbb{R}, i=0,\ldots,N-1$. We want to find a 
discrete distribution $P$ by the objective
\begin{equation*}
\begin{array}{rl}
    &\displaystyle\max_P \sum_{i=0}^{N-1}P(i)(X(i) - \alpha \log P(i)),\\
    s.t.& \displaystyle\sum_{i=0}^{N-1}P(i)=1,\\
\end{array}
\end{equation*}
where $\alpha>0$. Using a Lagrangian multiplier $\lambda$, we convert it to an unconstrained optimization 
problem:
\[\max_{P,\lambda}\sum_{i=0}^{N-1}P(i)(X(i)-\alpha \log P(i))+\lambda(\sum_{i=0}^{N-1} P(i) - 1).\]
Taking the derivative w.r.t. each $P(i)$ and setting it to 0, we have
\[P^*(i)=\exp\left(\frac{X(i)+\lambda-\alpha}{\alpha}\right)\propto \exp\left(\frac{X(i)}{\alpha}\right),\]
where $\lambda$ is calculated to ensure $\sum_{i=0}^{N-1}P^*(i)=1$.
Furthermore, let $Z=\sum_i\exp(\frac{X(i)}{\alpha})$ be the normalizer.
The resulting maximized objective is 
\begin{equation}
\label{eq:log_z}
    \displaystyle\sum_{i=0}^{N-1}P^*(i)(X(i)-\alpha(\frac{X(i)}{\alpha}-\log Z))
    =\displaystyle\sum_{i=0}^{N-1}P^*(i)\alpha\log Z
    =\displaystyle\alpha\log Z.
\end{equation}

To derive $\beta^*$ for Eq.~\ref{eq:ent_v_opt} given any $(s,a^-)\sim\mathcal{D}, \hat{a}\sim \pi_{\thetaA}$, 
we set $X(0)=Q_{\thetaQ}(s,a^-)$ and $X(1)=Q_{\thetaQ}(s,\hat{a})$. Then $\beta^*$ can be found
as below:
\begin{equation}
\label{eq:optimal_beta}
\beta_b^* \propto \exp\left(\frac{(1-b)Q_{\thetaQ}(s,a^-) + bQ_{\thetaQ}(s,\hat{a})}{\alpha}\right).
\end{equation}
Since this is a global maximum solution given any sampled $(s,a^-,\hat{a})$, $\beta^*$ is guaranteed to be no worse than any
parameterized policy. Putting $\beta^*$ back into Eq.~\ref{eq:ent_v_opt}, we are able to simplify $V_{\thetaQ}^{\pi^{\text{ta}}}$ (referring to Eq.~\ref{eq:log_z}) as
\begin{equation}
\label{eq:ent_v_simple}
    \displaystyle\expect_{\hat{a}\sim \pi_{\thetaA}}
    \alpha \left[ \log\sum_{b=0}^1 \exp\left(\frac{(1-b)Q_{\thetaQ}(s,a^-) + bQ_{\thetaQ}(s,\hat{a})}{\alpha}\right)
            - \log \pi_{\thetaA}(\hat{a}|s,a^-)
    \right].
\end{equation}
Then we apply the re-parameterization trick $\hat{a}=f_{\thetaA}(\epsilon,s,a^-), \epsilon\sim \mathcal{N}(0,I)$.
Approximating the gradient w.r.t. $\thetaA$ with a single sample of $\epsilon$, we get Eq.~\ref{eq:policy_gradient}.

%To derive $\pi^*$ for Eq.~\ref{eq:ent_v_simple} given any $(s,a^-)\sim\mathcal{D}$, we set 
%$X(\hat{a})=\alpha \log \sum_{b=0}^1\exp(\frac{Q_{\thetaQ}(s,a^-,\hat{a},b)}{\alpha})$. Then $\pi^*$ can be found as 
%$P^*$ above.

\begin{figure}[!t]
    \centering
    \includegraphics[width=\textwidth]{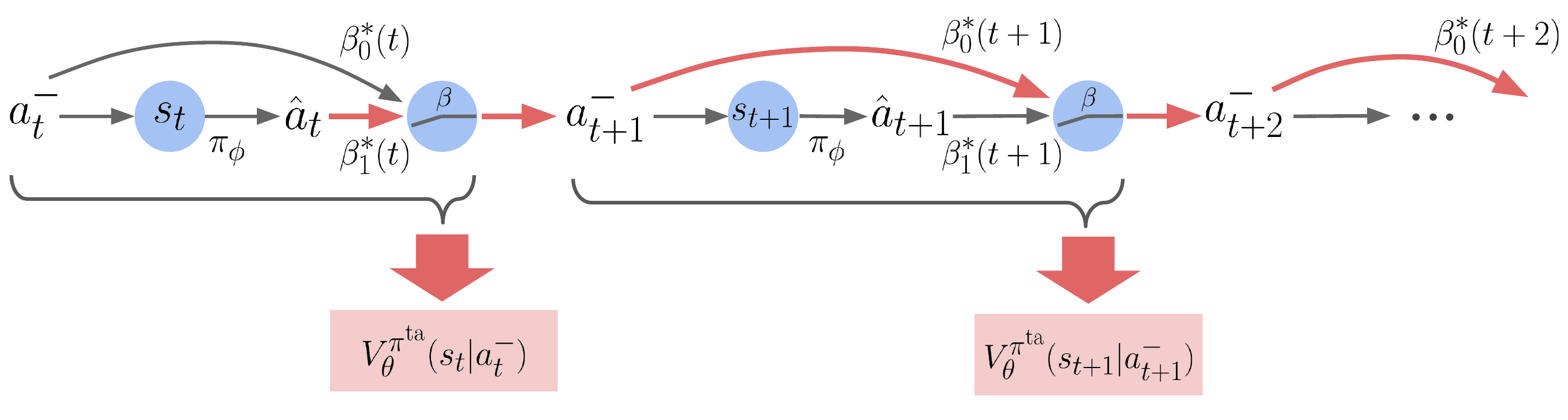}
    \caption{Rollout computational graph of $\pi^{\text{ta}}$ starting from step $t$.
    Rectangles denote negative losses to maximize.
    The red edges denote the gradient paths (reversed direction) for $\hat{a}_t$ when maximizing the sum of all V values: $\sum_{m=0}^{M-1} V_{\theta}^{\pi^{\text{ta}}}(s_{t+m}|a_{t+m}^-)$.
    }
\label{fig:rollout_graph}
\end{figure}

\section{Multi-step actor gradient}
\label{app:multi-step-actor-grad}
Let us first consider on-policy training as a simplified setting when computing the actor gradient for our two-stage policy $\pi^{\text{ta}}$.
Suppose the unroll length is $M$, which means each time we unroll the current policy $\pi^{\text{ta}}$ for $M$ steps, do a gradient update with the collected data, and unroll with the updated policy for the next $M$ steps, and so on.
In this case, the rollout computational graph is illustrated in Figure~\ref{fig:rollout_graph}.
Let $\beta_b^*(t+m)=\beta^*(b|s_{t+m},\hat{a}_{t+m},a^-_{t+m})$ be the optimal $\beta$ policy at step $t+m$ and define $w_{t+m}=\beta_1^*(t)\prod_{m'=1}^m\beta^*_0(t+m')$ to be the probability of $V_{\thetaQ}^{\pi^{\text{ta}}}(s_{t+m}|a_{t+m}^-)$ adopting $\hat{a}_t$ via action repetition.
The overall V value over the $M$ steps to be maximized is
\[\mathcal{V}\triangleq\sum_{m=0}^{M-1}V_{\thetaQ}^{\pi^{\text{ta}}}(s_{t+m}|a_{t+m}^-).\]
Ignoring the entropy term, its gradient w.r.t. $\hat{a}^t$ is 
\begin{equation}
\label{eq:multi-step-actor-grad}
\frac{\partial \mathcal{V}}{\partial \hat{a}^t}=\sum_{m=0}^{M-1}w_{t+m}\frac{\partial Q_{\thetaQ}(s_{t+m},\hat{a}_t)}{\partial \hat{a}_t},
\end{equation}
where the weights $w_{t+m}$ correspond to different red partial paths staring from $\hat{a}^t$ and ending at different negative losses in Figure~\ref{fig:rollout_graph}.

The above computational graph assumes that when unrolling the current $\pi^{\text{ta}}$, we are able to interact with the environment to obtain states $\{s_{t+1},s_{t+2},\ldots\}$.
This is true for on-policy training but not for off-policy training.
In the latter case, if we directly use the sampled sequence $\{s_t,s_{t+1},s_{t+2},\ldots\}$ from the replay buffer in Eq.~\ref{eq:multi-step-actor-grad}, the resulting gradient will suffer from off-policyness.
Thus in practice, we truncate the gradient to the first step, $\frac{\partial\mathcal{V}}{\partial \hat{a}_t}\approx w_t\frac{\partial Q_{\thetaQ}(s_t,\hat{a}_t)}{\partial \hat{a}_t}$.
We believe that this truncation is a simple but good approximation, for the two reasons:
\begin{compactenum}[a)]
    \item Because Eq.~\ref{eq:multi-step-actor-grad} is defined for a sampled state-action trajectory, $\frac{\partial Q_{\thetaQ}(s_{t+m},\hat{a}_t)}{\partial \hat{a}_t}$ has a much higher sample variance as $m$ increases.
    \item The weights $w_{t+m}$ decrease exponentially so the influence of $\hat{a}_t$ on future $\frac{\partial Q_{\thetaQ}}{\partial \hat{a}_t}$ quickly decays.
\end{compactenum}
Empirically, the truncated one-step gradient yields good results in our experiments.

\section{Different temperatures for $\beta$ and $\pi_{\thetaA}$}
\label{app:different-temps}
We use two different temperatures $\alpha'$ and $\alpha''$ for weighting the entropy of $\beta$ and $\pi_{\thetaA}$
respectively, to have a finer control of their entropy terms.
Accordingly, the objective in Eq.~\ref{eq:ent_v_opt} changes to
\begin{equation*}
    \displaystyle\expect_{(s,a^-)\sim\mathcal{D},\hat{a}\sim \pi_{\thetaA}, b\sim\beta}
    \left[ 
            (1-b)Q_{\thetaQ}(s,a^-) + bQ_{\thetaQ}(s,\hat{a})
            - \alpha'\log \beta_b - \alpha''\log \pi_{\thetaA}(\hat{a}|s,a^-)
    \right].
\end{equation*}
%
%With this change, the convergence proof in Appendix~\ref{app:proof} still applies since its analysis is independent of the 
%entropy terms. 

Revisiting Section~\ref{sec:policy_improve}, several key formulas are updated to reflect this change.
Eq.~\ref{eq:optimal_beta} is updated to
\begin{equation}
\beta_b^* \propto \exp\left(\frac{(1-b)Q_{\thetaQ}(s,a^-) + bQ_{\thetaQ}(s,\hat{a})}{\alpha'}\right).
\end{equation}
Eq.~\ref{eq:policy_gradient} is updated to
\begin{equation*}
\begin{array}{l}
\Delta\phi\triangleq\displaystyle\left(\beta^*_1\frac{\partial Q_{\thetaQ}(s,\hat{a})}{\partial \hat{a}} - \alpha''\frac{\partial \log\pi_{\thetaA}(\hat{a}|s,a^-)}{\partial \hat{a}}\right)
\frac{\partial f_{\thetaA}}{\partial \thetaA}
- \displaystyle\alpha'' \frac{\partial \log \pi_{\thetaA}(\hat{a}|s,a^-)}{\partial \thetaA}.\\
\end{array}
\end{equation*}
%
%Eq.~\ref{eq:optimal_a} is updated to 
%%
%\begin{equation*}
%\pi^*(\hat{a}|s,a^-)\propto \left[\sum_{b=0}^1\exp(\frac{Q_{\thetaQ}(s,a^-,\hat{a},b)}{\alpha'})\right]^{\frac{\alpha'}{\alpha''}}.
%\end{equation*}
%
%Finally, we adjust $\alpha'$ and $\alpha''$ separately according to their own entropy targets $\mathcal{H}'$ and 
%$\mathcal{H}''$ with two separate objectives similar to Eq.~\ref{eq:alpha}.

\section{SAC-Krep}
\label{app:sac-mixed}
Following \citet{Delalleau2019} we extend the original SAC algorithm to support a hybrid of discrete and continuous actions~\footnote{An implementation of SAC with hybrid actions is available at \url{https://github.com/HorizonRobotics/alf/blob/pytorch/alf/algorithms/sac_algorithm.py}.}, to 
implement the baseline SAC-Krep~\citep{sharma2017learning,Biedenkapp2021} in Section~\ref{sec:comparison_methods}.
We denote the discrete and continuous actions by $b$ ($1\le b \le B$) and $a$, respectively.
Let the joint policy be $\pi(a,b|s)=\pi_{\thetaA}(a|s)\pi(b|s,a)$, namely, 
the joint policy is decomposed in a way that it outputs a continuous action, followed by a discrete action conditioned on 
that continuous action.
Let $Q_{\thetaQ}(s,a,b)$ be the parameterized expected return of taking action $(a,b)$ at state $s$.
Then the entropy-augmented state value is computed as
\begin{equation*}
V_{\thetaQ}^{\pi}(s)=\expect_{(a,b)\sim\pi} 
\Big[
Q_{\thetaQ}(s,a,b) -\alpha''\log\pi_{\thetaA}(a|s) - \alpha'\log\pi(b|s,a)
\Big].
\end{equation*}
Similar to Section~\ref{sec:policy_improve}, we can derive an optimal closed-form $\pi^*(b|s,a)$ given any $(s,a)$, and then optimize the continuous policy $\pi_{\thetaA}(a|s)$ similarly to Eq.~\ref{eq:policy_gradient}.

For policy evaluation, in the case of SAC-Krep, $b$ represents how many steps $a$ will be executed without being interrupted.
Thus the objective of learning $Q_{\thetaQ}$ is
\begin{equation*}
    \begin{array}{l}
        \displaystyle \Min_{\thetaQ} \expect_{(s_t,a_t,b_t,s_{t+b_t})\sim \mathcal{D}} \left[Q_{\thetaQ}(s_t,a_t,b_t)- \mathcal{B}^{\pi}Q_{\bar{\thetaQ}}(s_t,a_t,b_t)\right]^2, \\ 

        \text{with}\ \mathcal{B}^{\pi}Q_{\bar{\thetaQ}}(s_t,a_t,b_t)=\displaystyle\sum_{t'=t}^{t+b_t-1}\gamma^{t'-t}r(s_{t'},a_{t'},s_{t'+1}) + \gamma^{b_t} V^{\pi}_{\bar{\thetaQ}}(s_{t+b_t}),\\
    \end{array}
\end{equation*}
Namely, the Q value is bootstrapped by $b$ steps. 
We instantiate the Q network by having the continuous action $a$ as an input in addition to $s$, and let the network output $B$ heads, 
each representing $Q(s,a,b)$.

\section{SAC-Hybrid}
\label{app:sac-hybrid}
Following H-MPO~\citep{Neunert2020} we define a factored policy of a newly sampled continuous action $\hat{a}$ and a binary switching action $b$:
\[\pi((\hat{a},b)|s,a^-)=\pi_{\thetaA_a}(\hat{a}|s,a^-)\pi_{\thetaA_b}(b|s,a^-),\]
where the observation consists of state $s$ and previous action $a^-$.
Note that a big difference between this formulation with either \name or SAC-Krep (Appendix~\ref{app:sac-mixed}) is that $\hat{a}$ and $b$ are independent.
That is, the decision of ``repeat-or-act`` is made in parallel with the new action. 
The entropy-augmented state value is computed as 
\[V_{\thetaQ}^{\pi}((s,a^-))=\expect_{\hat{a}\sim\pi_{\thetaA_a},b\sim\pi_{\thetaA_b}}\Big[
Q_{\thetaQ}((s,a^-),(\hat{a},b)) -\alpha''\log\pi_{\thetaA_a}(\hat{a}|s,a^-) - \alpha'\log\pi_{\thetaA_b}(b|s,a^-)
\Big],\]
and $\thetaA_a$ and $\thetaA_b$ can be optimized by gradient ascent. 
Finally, the Bellman operator for policy evaluation is
\[\mathcal{B}^{\pi}Q_{\bar{\thetaQ}}((s,a^-),(\hat{a},b))=r(s,a,s') + \gamma V_{\bar{\thetaQ}}^{\pi}((s',a)),\]
where $a=(1-b)a^-+b\hat{a}$ is the action output to the environment.
Similar to SAC-Krep, to instantiate the Q network, we use $(s,a^-,\hat{a})$ as the inputs and let the network output two heads for $b=0$ and $b=1$.
Compared to \name's Q formulation (Eq.~\ref{eq:policy_eval}), clearly SAC-Hybrid's Q has to handle more input/output mappings for the same transition dynamics, 
which makes the policy evaluation less efficient.

\begin{table*}[!t]
\resizebox{\textwidth}{!}{
    \begin{tabular}{l|l|l|l|l|l}
    \multirow{2}{*}{Category} & \multirow{2}{*}{Task} & \multirow{2}{*}{Gym environment name} & \multirow{2}{*}{Observation space} & \multirow{2}{*}{Action space} & Reward
    \\
    & & & & & normalization \\
    \hline
    \multirow{3}{*}{\textbf{SimpleControl}} & \task{MountainCarContinuous} & \texttt{MountainCarContinuous-v0} & $\mathbb{R}^2$ & $[-1,1]^1$ & \multirow{3}{*}{$[-5,5]$} \\
    & \task{LunarLanderContinuous} & \texttt{LunarLanderContinuous-v2} & $\mathbb{R}^8$ & $[-1,1]^2$ & \\
    & \task{InvertedDoublePendulum} & \texttt{InvertedDoublePendulum-v2} & $\mathbb{R}^{11}$ & $[-1,1]^1$ & \\
    \hline
    \multirow{4}{*}{\textbf{Locomotion}} & \task{Hopper} & \texttt{Hopper-v2} & $\mathbb{R}^{11}$ & $[-1,1]^3$ & \multirow{4}{*}{$\times$}\\
    & \task{Ant} & \texttt{Ant-v2} & $\mathbb{R}^{111}$ & $[-1,1]^8$ & \\ \cline{3-5}
    & \task{Walker2d} & \texttt{Walker2d-v2} & \multirow{2}{*}{$\mathbb{R}^{17}$} & \multirow{2}{*}{$[-1,1]^6$} & \\
    & \task{HalfCheetah} & \texttt{HalfCheetah-v2} & & & \\
    \hline
    \multirow{2}{*}{\textbf{Terrain}} & \task{BipedalWalker} & \texttt{BipedalWalker-v2} & \multirow{2}{*}{$\mathbb{R}^{24}$} & \multirow{2}{*}{$[-1,1]^4$} & \multirow{2}{*}{$[-1,1]$} \\
    & \task{BipedalWalkerHardcore} & \texttt{BipedalWalkerHardcore-v2} & & & \\
    \hline
    \multirow{4}{*}{\textbf{Manipulation}} & \task{FetchReach} & \texttt{FetchReach-v1} & $\mathbb{R}^{13}$ & \multirow{4}{*}{$[-1,1]^4$} & \multirow{4}{*}{$[-1,1]$} \\ \cline{3-4}
    & \task{FetchPush} & \texttt{FetchPush-v1} & \multirow{3}{*}{$\mathbb{R}^{28}$} & & \\
    & \task{FetchSlide} & \texttt{FetchSlide-v1} & & & \\
    &\task{FetchPickAndPlace} & \texttt{FetchPickAndPlace-v1} & & & \\
    \hline
    \multirow{7}{*}{\textbf{Driving}} & \multirow{7}{*}{\task{Town01}} & \multirow{7}{*}{\texttt{Town01}} & ``camera'': $\mathbb{R}^{128\times 64\times 3}$,
    & \multirow{7}{*}{$[-1,1]^4$}  & \multirow{7}{*}{$[-5,5]$} \\
    & & & ``radar'': $\mathbb{R}^{200 \times 4}$, & & \\
    & & & ``collision'': $\mathbb{R}^{4 \times 3}$, & & \\
    &&& ``IMU'': $\mathbb{R}^7$, & & \\
    &&& ``goal'': $\mathbb{R}^3$, & & \\
    &&& ``velocity'': $\mathbb{R}^3$, &&\\
    &&& ``navigation'': $\mathbb{R}^{8\times 3}$ &&\\
    &&& ``prev action'': $[-1,1]^{4}$ &&\\
    \end{tabular}
}
\caption{The 14 tasks with their environment details.
Note that reward clipping is performed after reward normalization (if applied).
Except \task{Town01}, the input observation is a flattened vector.
}
\label{tab:tasks}
\end{table*}

\section{Task details}
\label{app:tasks}
All 14 tasks are wrapped by the OpenAI Gym \citep{Brockman2016} interface.
All of them, except \task{Town01}, are very standard and follow their 
original definitions.
The environment of \task{Town01} is customized by us with various map options using a base map called ``Town01'' provided by the CARLA 
simulator \citep{Dosovitskiy17}, which we will describe in detail later.
We always scale the action space of every task to $[-1,1]^A$, where $A$ is the action dimensionality defined by the task environment.
The observation space of each task is unchanged, except for \task{Town01}.
Note that we use MuJoCo 2.0 \citep{Todorov2012} for simulating \textbf{Locomotion} and \textbf{Manipulation} \footnote{A different version of MuJoCo may result in different observations, rewards, and incomparable environments; see \url{https://github.com/openai/gym/issues/1541}.}.
A summary of the tasks is in Table~\ref{tab:tasks}.

\subsection{Reward normalization}
\label{app:reward_norm}
We normalize each task's reward using a normalizer that maintains adaptive exponential moving averages of the reward and its second moment.
Specifically, let $\xi$ be a pre-defined update speed ($\xi=8$ across all experiments), and $L$ be the total number of times the normalizer statistics has been updated so far, then 
for the incoming reward $r$, the mean $m_1$ and second moment $m_2$ are updated as 
\begin{equation*}
    \begin{array}{ll}
        \eta_L & = \frac{\xi}{L+\xi},\\
        m_1 & \leftarrow (1-\eta_L)m_1 + \eta_L r,\\
        m_2 & \leftarrow (1-\eta_L)m_2 + \eta_L r^2,\\
        L & \leftarrow L + 1,\\
    \end{array}
\end{equation*}
with $L=m_1=m_2=0$ as the initialized values.
Basically, the moving average rate $\eta_L$ decreases according to $\frac{1}{L}$. 
With this averaging strategy, one can show that by step $L$, the weight for the reward encountered at step $l \le L$ is roughly in proportional 
to $(\frac{l}{L})^{(\xi-1)}$.
Intuitively, as $L$ increases, the effective averaging window expands because the averaging weights are computed by the changing 
ratio $\frac{l}{L}$.
Finally, given any reward $r'$, it is normalized as 
\[\min(\max(\frac{r'-m_1}{\sqrt{m_2-m_1^2}}, -c), c),\]
where $c>0$ is a constant set to either 1 or 5, according to which value produces better performance for SAC on a task.
We find that the suite of \textbf{Locomotion} tasks is extremely sensitive to reward definition, and thus do not apply reward normalization to it.
The normalizer statistics is updated only when rewards are sampled from the replay buffer.
Note that for a task, the same reward normalization (if applied) is used by all 8 evaluated methods with no discrimination.

\subsection{Town01}
\begin{wrapfigure}[20]{r}{0.4\textwidth}
    \centering
    \includegraphics[width=0.3\textwidth]{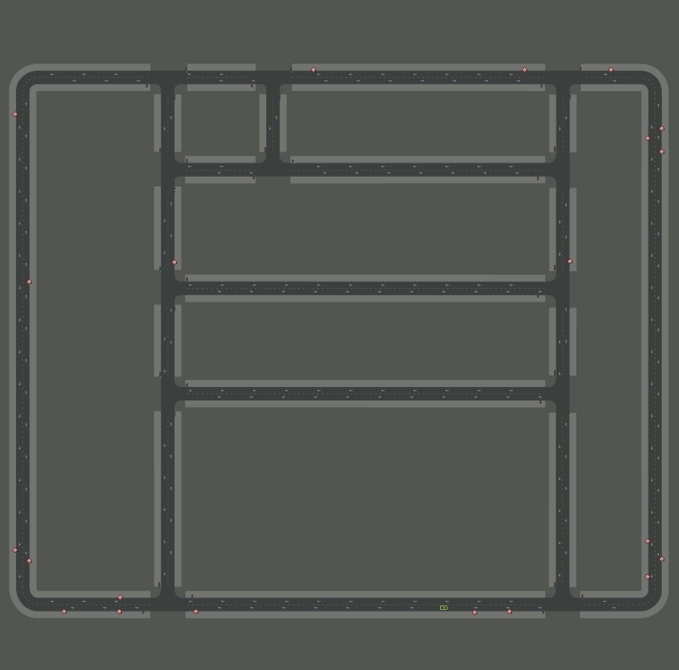}
    \caption{The layout of the map ``Town01'' (picture from \url{https://carla.readthedocs.io}). 
    The actual map is filled with other objects such as buildings, trees, pedestrians, and traffic lights to make
    it a realistic scene of a small town. 
    }
    \label{fig:town01}
\end{wrapfigure}

Our \task{Town01} task is based on the ``Town01'' map \citep{Dosovitskiy17} that consists of 12 T-junctions (Figure~\ref{fig:town01}).
The map size is roughly $400 \times 400$ \si[per-mode = symbol]{\metre\squared}.
At the beginning of each episode, the vehicle is first randomly spawned at a lane location.
Then a random waypoint is selected on the map and is set as the destination for the vehicle.
The maximal episode length (time limit) is computed as
\[N_{\text{frames}}=\frac{L_{\text{route}}}{S_{\text{min}}\cdot \Delta t}\]
where $L_{\text{route}}$ is the shortest route length calculated by the simulator, $S_{\text{min}}$ is the average minimal speed expected for a meaningful driving, and $\Delta t$ is the simulation step time. We set $S_{\text{min}}=5$\si[per-mode = symbol]{\metre\per\s} and $\Delta t=0.1$\si[per-mode = symbol]{\s} through the experiment.
An episode can terminate early if the vehicle reaches the destination, or gets stuck at collision for over a certain amount of time.
We customize the map to include 20 other vehicles and 20 pedestrians that are programmed by the simulator's built-in AI to act in the scenario.
We use the default weather type and set the day length to 1000 seconds.

The action space of the vehicle is 4 dimensional: (``throttle'', ``steer'', ``brake'', ``reverse''). 
We customize the observation space to include 8 different multi-modal inputs:
\begin{compactenum}
    \item ``camera'': a monocular RGB image ($128\times 64\times 3$) that shows the road condition in front of the vehicle;
    \item ``radar'': an array of 200 radar points, where each point is represented by a 4D vector;
    \item ``collision'': an array of 4 collisions, where each collision is represented by a 3D vector;
    \item ``IMU'': a 7D IMU measurement vector of the vehicle's status;
    \item ``goal'': a 3D vector indicating the destination location;
    \item ``velocity'': the velocity of the vehicle relative to its own coordinate system;
    \item ``navigation'': an array of 8 future waypoints on the current navigation route, each waypoint is a 3D vector in the         coordinate system of the vehicle;
    \item ``prev action'': the action taken by the vehicle at the previous time step.
\end{compactenum}

Since the observation space of \task{Town01} is huge, we apply normalization to all input sensors for more efficient 
training.
We normalize each input sensor vector in a similar way of reward normalization.
After normalization, the vector is element-wisely clipped to $[-5,5]$.

To train the vehicle, we define the task reward by 4 major components:
\begin{compactenum}
    \item ``distance'': a shaped reward that measures how much closer the vehicle is to the next navigation waypoint after one time step;
    \item ``collision'': if a collision is detected, then the vehicle gets a reward of $-\min(20, 0.5\cdot\max(0, \bar{R}))$, where $\bar{R}$ 
        is the accumulated episode reward so far;
    \item ``red light'': if a red light violation is detected, then the vehicle gets a reward of $-\min(20, 0.3\cdot\max(0, \bar{R}))$, 
        where $\bar{R}$ is the accumulated episode reward so far;
    \item ``goal'': the vehicle gets a reward of $100$ for reaching the destination.        
\end{compactenum}
The overall reward at a time step is computed as the sum of the above 4 rewards.
This reward definition ensures that SAC obtains reasonable performance in this task.

\subsection{Network structure}
The model architecture of all compared methods is identical: each trains an actor network and a critic network~\footnote{SAC-Krep actually has an extra discrete Q network that models the values of repeating $1,2,\ldots,N$ steps. It has the same structure with the critic network for the continuous action, but with multiple output heads.}.
Following \citet{fujimoto2018}, the critic network utilizes two replicas to reduce positive bias in the Q function.
We make sure that the actor and critic network always have the same structure (but with different weights) except for the final output layer.

Below we review the network structure designed for each task category, 
shared between the actor and critic network, and shared among all 8 evaluated methods.
No additional network or layer is owned exclusively by any method.
\begin{compactenum}[$-$]
    \item \textbf{SimpleControl}: two hidden layers, each of size 256.
    \item \textbf{Locomotion}: two hidden layers, each of size 256.
    \item \textbf{Terrain}: two hidden layers, each of size 256.
    \item \textbf{Manipulation}: three hidden layers, each of size 256.
    \item \textbf{Driving}: We use an encoder to combine multi-modal sensor inputs. 
    The encoder uses a mini ResNet \citep{He2016} of 6 bottleneck blocks (without BatchNorm) to encode an RGB image into a latent embedding of size 256, 
    where each bottleneck block has a kernel size of 3, filters of $(64, 32, 64)$, and a stride of 2 (odd block) or 1 (even block).
    The encoder then flattens any other input and projects it to a latent embedding of size 256.
    All the latent embeddings are averaged and input to an FC layer of size 256 to yield a single encoded vector that summarizes the input 
    sensors.
    Finally, the actor/critic network is created with one hidden layer of size 256, with this common encoded vector as input.
    We detach the gradient when inputting the encoded vector to the actor network, and only allow the critic network to learn it.
\end{compactenum}
We use ReLU for all hidden activations.

\section{Experiment details}
\label{app:experiments}
\subsection{Entropy target calculation}
When computing an entropy target, instead of directly specifying a floating number which is usually unintuitive, we 
calculate it by an alternative parameter $\delta$.
If the action space is continuous, then suppose that it has $K$ dimensions, and every dimension 
is bounded by $[m,M]$.
We assume the entropy target to be the entropy of a continuous distribution 
whose probability uniformly concentrates on a slice of the support $\delta(M-m)$ with $P=\frac{1}{\delta(M-m)}$.
Thus the entropy target is calculated as 
\begin{equation}
\label{eq:target_entropy}
    \displaystyle -K\int_m^M P(a)\log P(a) \text{d}a 
    = \displaystyle -K\log \frac{1}{\delta(M-m)}
    = \displaystyle K\left[\log\delta + \log(M-m)\right].
\end{equation}
For example, by this definition, an entropy target of -1 per dimension used by \citet{Haarnoja2018} is equivalent to setting 
$\delta=0.184$ here with $M=1$ and $m=-1$.
If the action space is discrete with $K > 1$ entries,
we assume the entropy target to be the entropy of a discrete distribution that 
has one entry of probability $1-\delta$, with $\delta$ uniformly distributed over the other $K-1$ entries.
Thus the entropy target is calculated as 
\begin{equation}
\label{eq:target_entropy_discrete}
-\delta\log\frac{\delta}{K-1}-(1-\delta)\log(1-\delta).
\end{equation}
We find setting $\delta$ instead of the direct entropy target is always more intuitive in practice.

\begin{table*}[!t]
\definecolor{Gray}{gray}{0.8}
\newcolumntype{a}{>{\columncolor{Gray}}c}
    \centering
    \resizebox{\textwidth}{!}{
    \begin{tabular}{@{}l|c|c|c|a|a@{}}
        \multirow{2}{*}{Hyperparameter} &  \multirow{2}{*}{\textbf{SimpleControl}} 
        &  \multirow{2}{*}{\textbf{Terrain}} &  \multirow{2}{*}{\textbf{Driving}} & \textbf{Manipulation} & \textbf{Locomotion}\\
        & &  & & \citep{plappert2018multigoal} & \citep{Haarnoja2018}\\
        \hline
        Learning rate & $10^{-4}$ & $5\times 10^{-4}$ & & $10^{-3}$ & $3\times 10^{-4}$ \\
        Reward discount & $0.99$ & & & $0.98$ & \\
        Number of parallel actors for rollout & $1$ & $32$ & $4$ & $38$ & \\
        Replay buffer size per actor & $10^5$ &  & & $2\times 10^4$ & $10^6$\\
        Mini-batch size & $256$  & $4096$ &$64$ & $4864$ & \\
        Entropy target $\delta$ (Eq.~\ref{eq:target_entropy}) & $0.1$ &  & & $0.2$ & $0.184$\\
        Target Q smoothing coefficient & $5\times 10^{-3}$ & & & $5\times 10^{-2}$ & \\
        Target Q update interval & 1 & & & 40 & \\
        Training interval (env frames) per actor & $1$ & $5$ & $10$ & $50/40$ & \\
        Total environment frames for rollout & $10^5$  & $5\times 10^6$& $10^7$ & $10^7$ & $10^6$ \\
    \end{tabular}    
    }
    \caption{The hyperparameter values of SAC on 5 task categories.
    The two shaded columns \textbf{Manipulation} and \textbf{Locomotion}  use exactly the same hyperparameter values from the original papers, 
    and we list them for completeness.
    An empty cell in the table means using the same hyperparameter value as the corresponding one of \textbf{SimpleControl}.
    The training interval  of \textbf{Manipulation} ($50/40$) means updating models 40 times in a row for every 50 environment steps (per actor), 
    which also follows the convention set by \citet{plappert2018multigoal}.
    }
    \label{tab:hyperparameters}
\end{table*}

\subsection{Hyperparameters}
We use Adam \citep{kingma2017adam} with $\beta_1=0.9$, $\beta_2=0.999$, and $\epsilon=10^{-7}$ to train each method.
Below we first describe the hyperparameter values of the vanilla SAC baseline.
These values are selected by referring to either previously published ones or our typical options for SAC runs.
For \textbf{Locomotion} and \textbf{Manipulation}, we directly adopt the hyperparameter values from 
the original papers.
Among the remaining 3 task categories (\textbf{SimpleControl}, \textbf{Terrain}, and \textbf{Driving}), 
several hyperparameter values vary due to task differences (Table~\ref{tab:hyperparameters}).
This variance also serves to test if our comparison result generalizes under different training settings.

The hyperparameter values of the other 7 evaluated methods are the same as shown in Table~\ref{tab:hyperparameters}, 
but with extra hyperparameters if required by a method.
Note that for open-loop action repetition methods like SAC-Nrep and SAC-Krep, the counting of environment frames includes frames generated 
by repeated actions.
For the repeating hyperparameter $N$ in Section~\ref{sec:comparison_methods}, we set it to 3 for \textbf{SimpleControl}, \textbf{Locomotion} and \textbf{Manipulation}, and 
to 5 for \textbf{Terrain} and \textbf{Driving}. 
For SAC-EZ, we set $\mu$ of the zeta distribution to 2 following \citet{dabney2020}, and linearly decay $\epsilon$ from $1$
to $0.01$ over the course of first $\frac{1}{10}$ of the training.
$\epsilon$ is then kept to be $0.01$ till the end of training.
Finally, for SAC-Hybrid, \name-1td, \name-Ntd, and \name, the discrete action
requires its own entropy target (Appendix~\ref{app:different-temps}) computed by Eq.~\ref{eq:target_entropy_discrete}.
We set $\delta$ in that equation to $0.05$ in all task categories.
%
%Also,  in \textbf{Locomotion} we enforce a maximal number of 5 repeating steps during rollout while still allowing an arbitrary number of repeating steps 
%during training and evaluation, for the three TASAC variants.
In \textbf{Locomotion}, we clip the advantage $Q_{\theta}(s,\hat{a})-Q_{\theta}(s,a^-)$ to $[0,+\infty)$ when computing $\beta^*$ for \name-1td, \name-Ntd, and \name.
This clipping biases the agent towards sampling new actions.

\subsection{Computational resources}
The required computational resources for doing all our experiments are moderate.
We define a job group as a (\texttt{method}, \texttt{task\_category}) pair, \eg\ (\name, \textbf{Locomotion}).
We used three RTX 2080Ti GPUs at the same time for training the jobs within one group simultaneously.
We evenly distributed a job group (each job in the group represents a (\texttt{task}, \texttt{random\_seed}) combination, \eg\ (\textit{Ant}, \textit{seed0})) across the three GPUs.
Finally, all job groups were launched on a cluster.
Among the groups, (\name, \textbf{Driving}) took the longest training time which was roughly 36 hours, while (SAC, \textbf{SimpleControl}) took the shortest time which was about 2 hours. 
The rest of job groups mostly finished within 12 hours each.

\section{More experimental results}
\label{app:more-results}

\begin{figure}[!h]
    \centering
    \resizebox{\textwidth}{!}{
    \begin{tabular}{@{}c@{}c@{}c@{}}
        \includegraphics[width=0.33\textwidth]{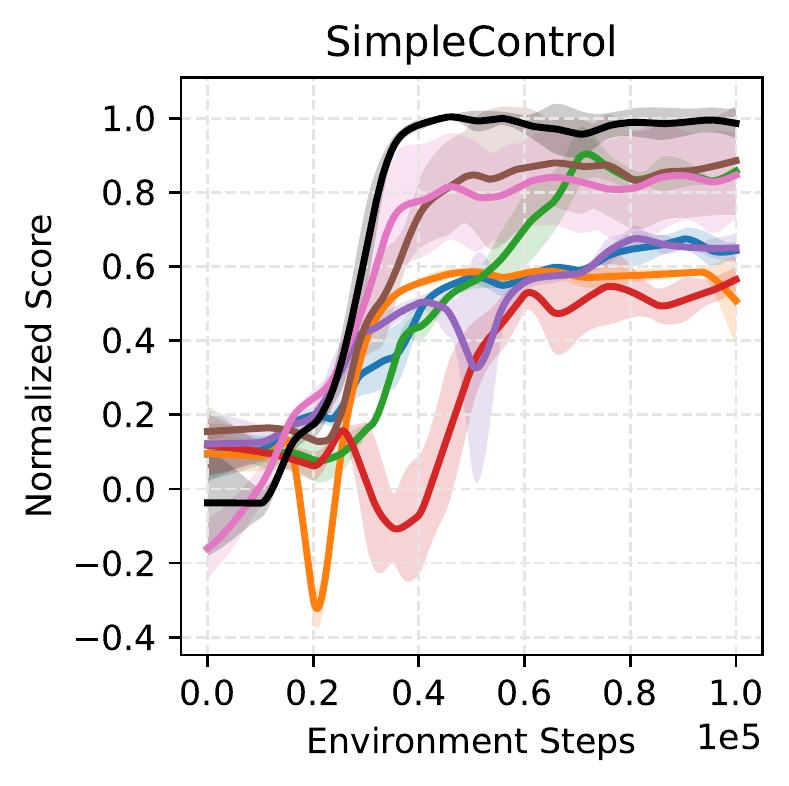}
        &\includegraphics[width=0.33\textwidth]{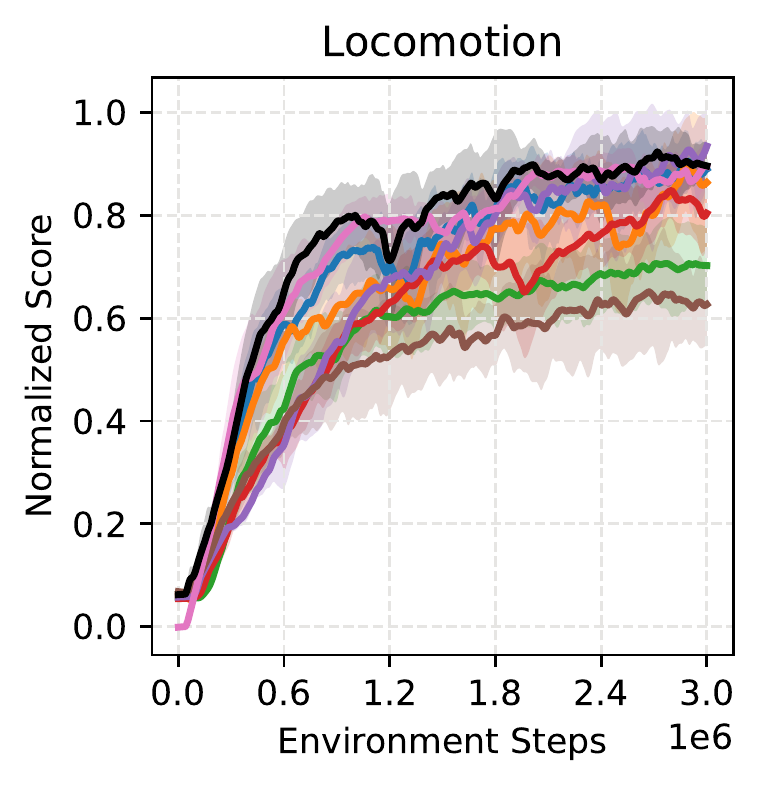}
        &\includegraphics[width=0.33\textwidth]{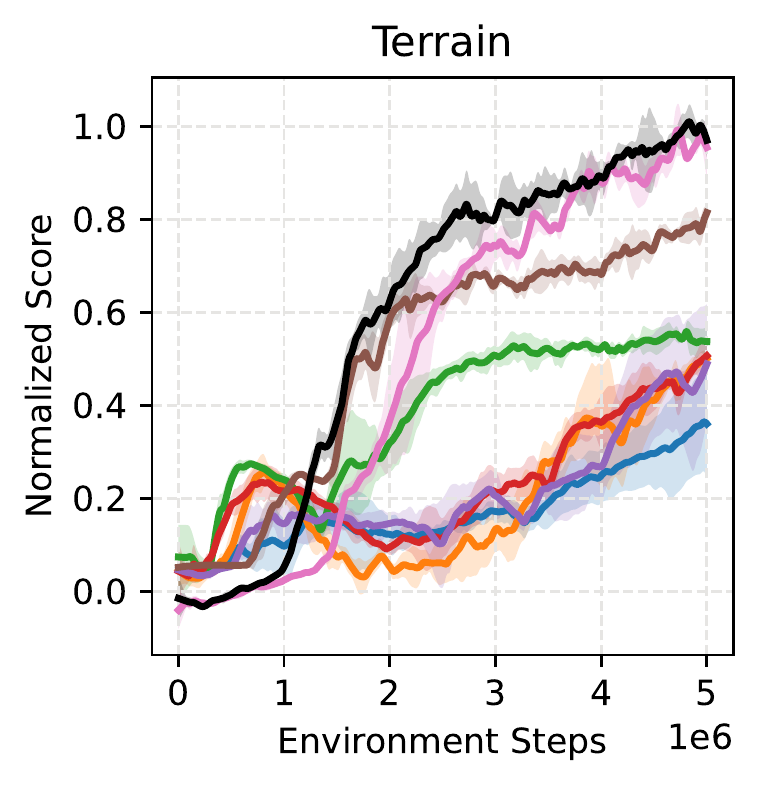}\\
        \includegraphics[width=0.33\textwidth]{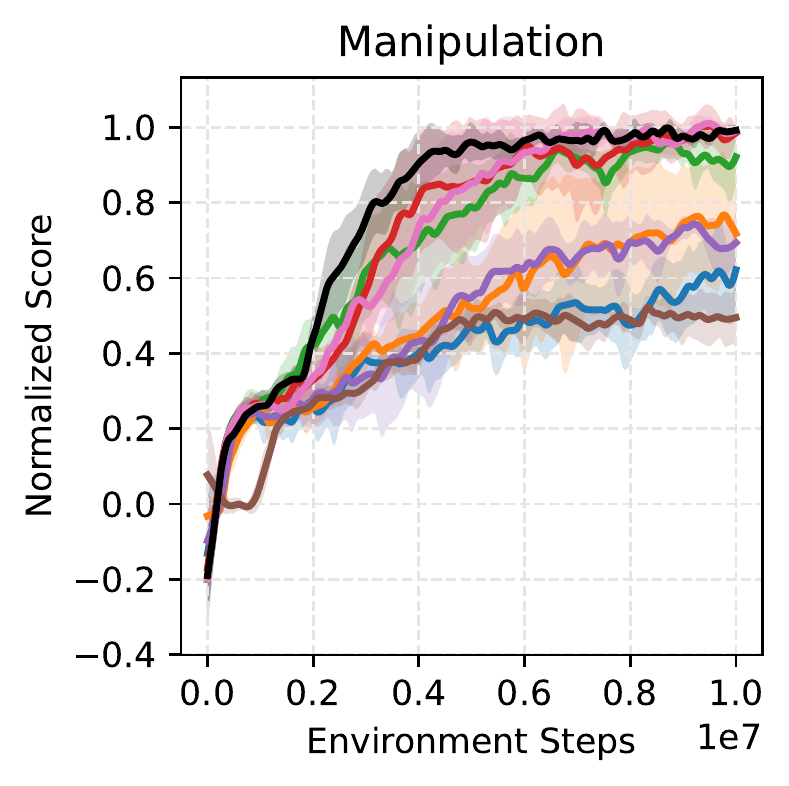}
        &\includegraphics[width=0.33\textwidth]{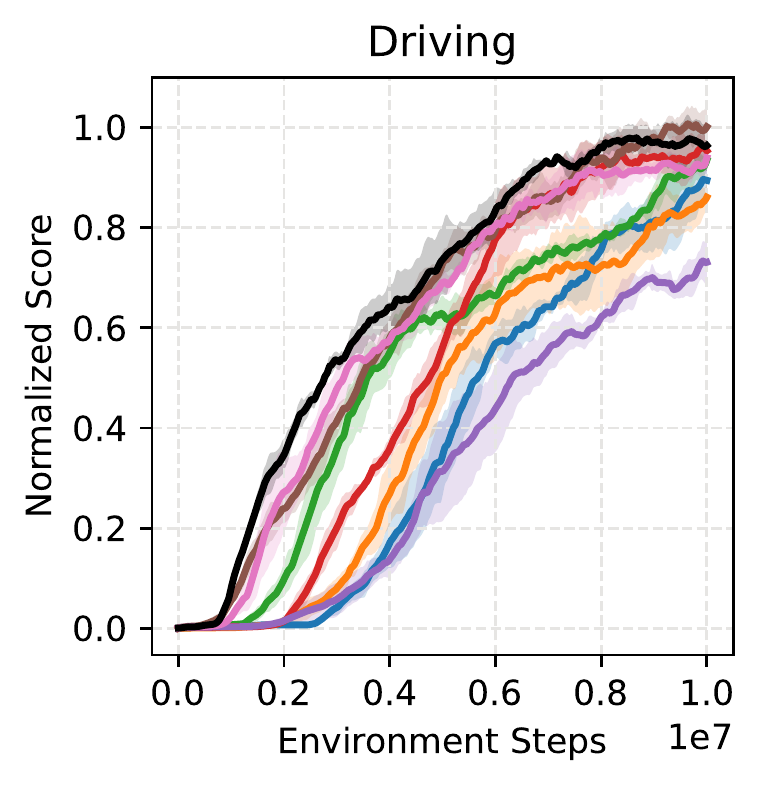}
        &\includegraphics[viewport =-20 -50 180 130, clip, width=0.33\textwidth]{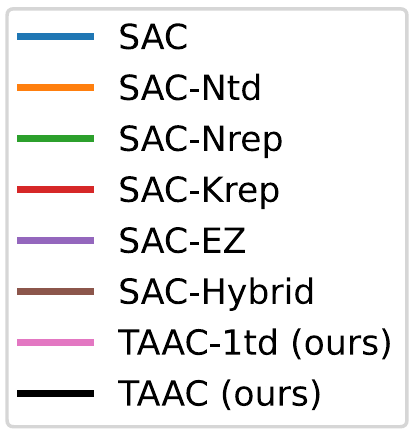} \\
    \end{tabular}
    }
    \caption{The n-score curves of the 5 task categories. 
    Each curve is a mean of a method's n-score curves on the tasks within a task category, where 
    the method is run with 3 random seeds for each task.
    }
\label{fig:curves}
\end{figure}

\begin{figure*}[!t]
    \centering
    \begin{tabular}{@{}c@{}c@{}c@{}c@{}}
    \includegraphics[width=0.29\textwidth]{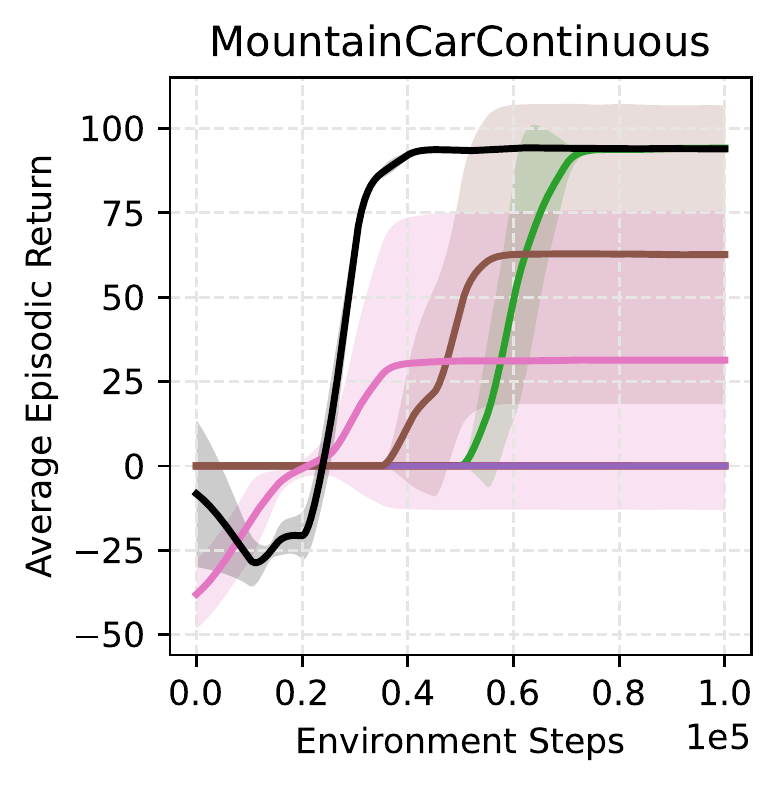}
    &\includegraphics[width=0.3\textwidth]{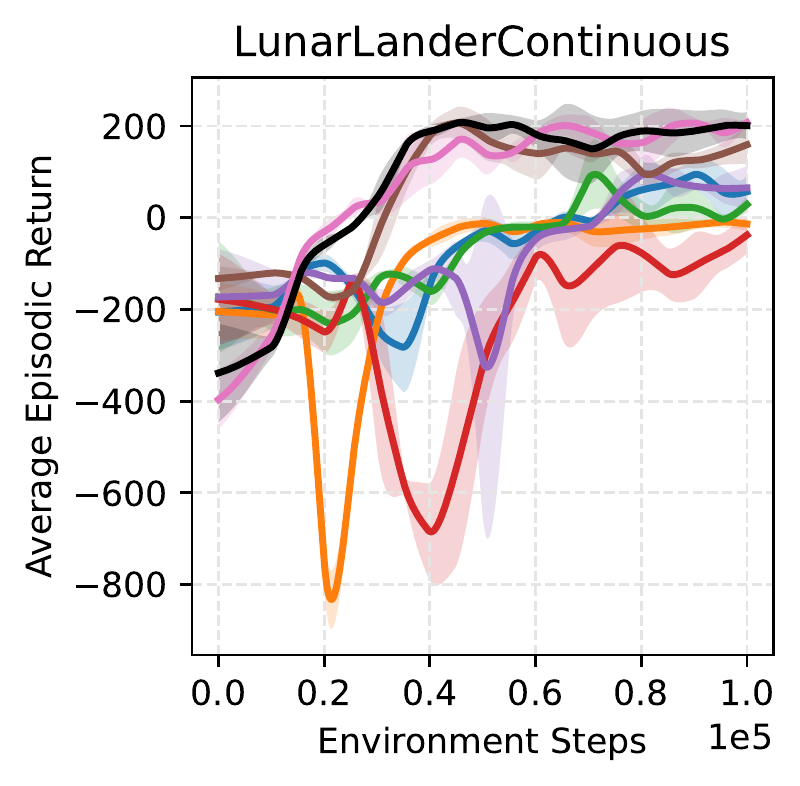}
    &\includegraphics[width=0.3\textwidth]{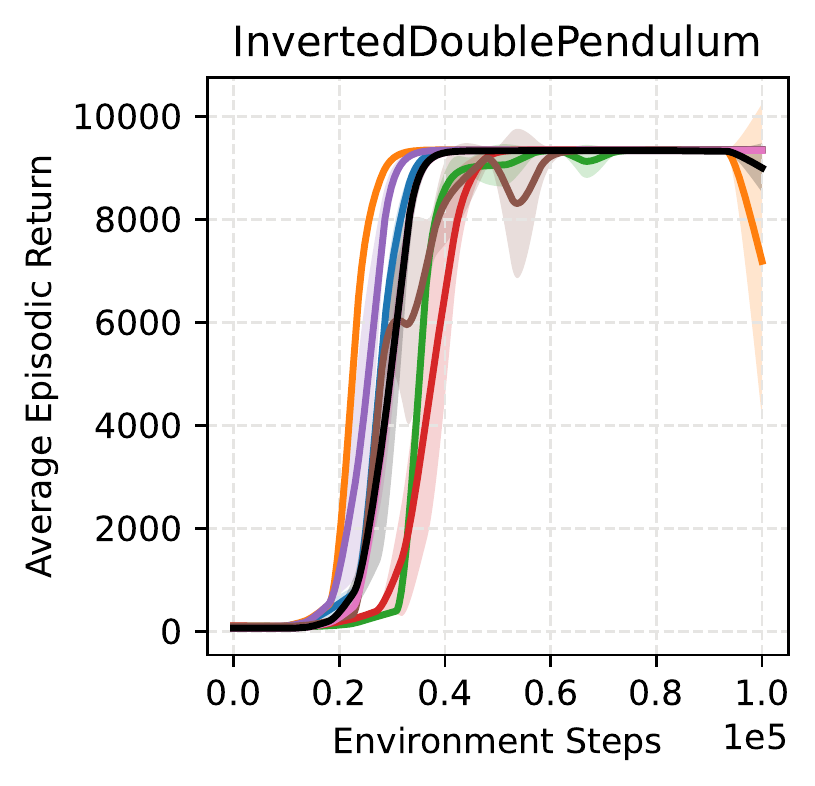}\\
    \includegraphics[width=0.3\textwidth]{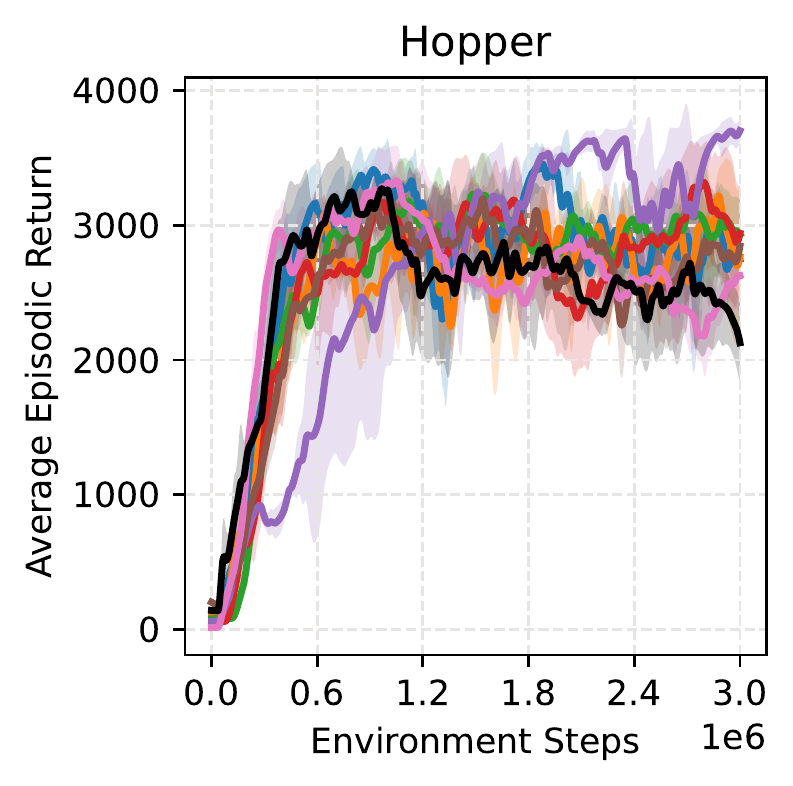}
    &\includegraphics[width=0.3\textwidth]{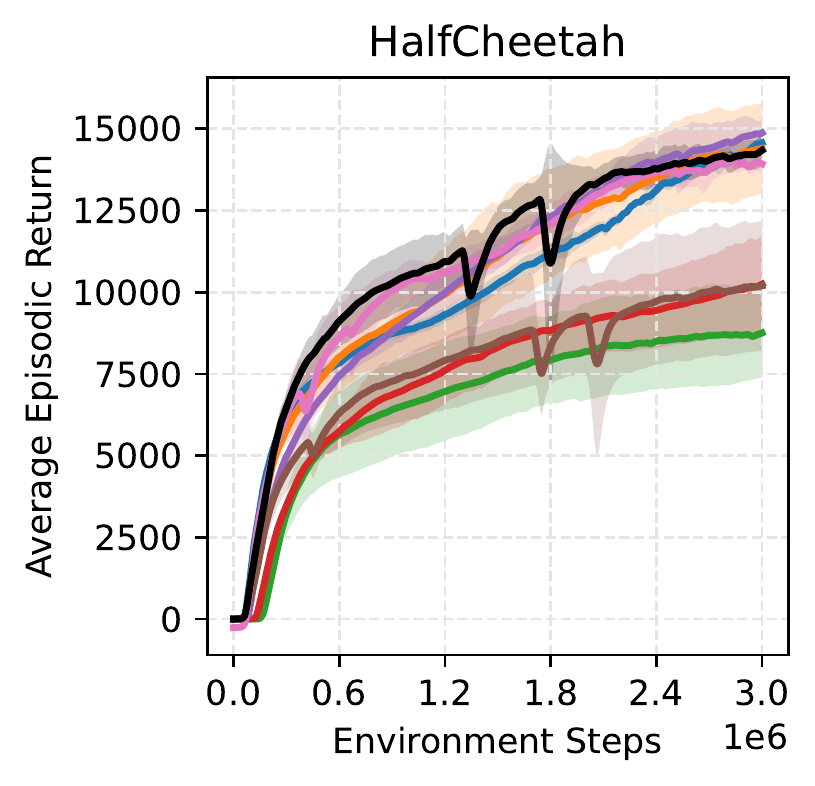}
    &\includegraphics[width=0.3\textwidth]{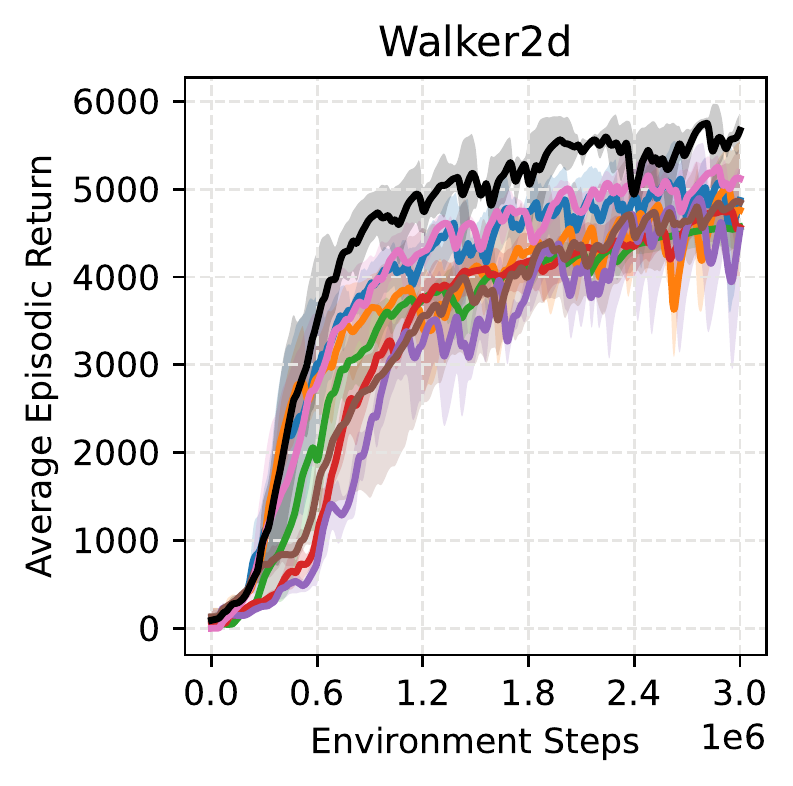}\\
    \includegraphics[width=0.3\textwidth]{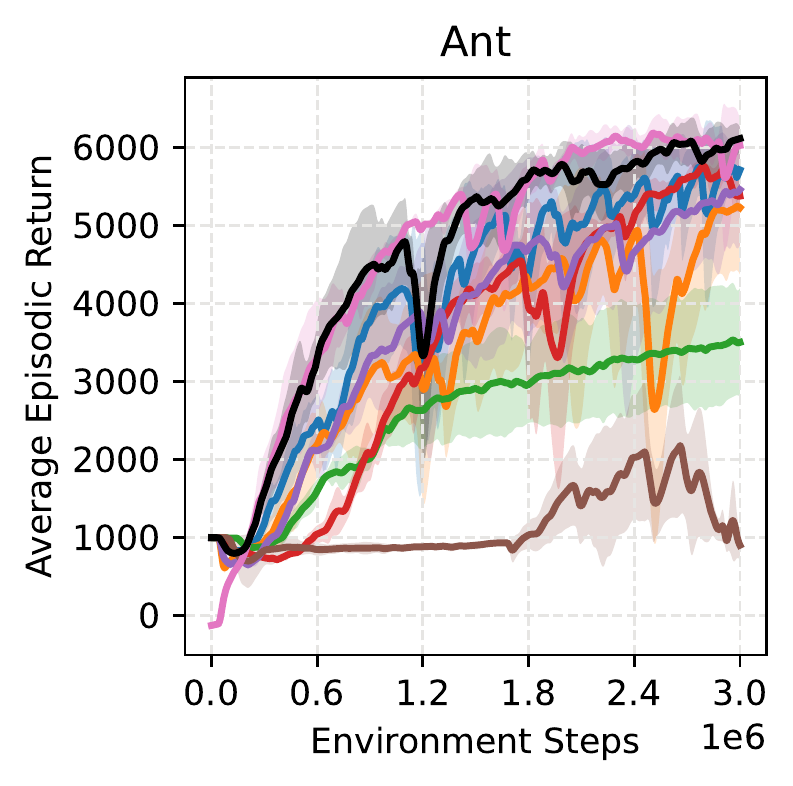}
    &\includegraphics[width=0.3\textwidth]{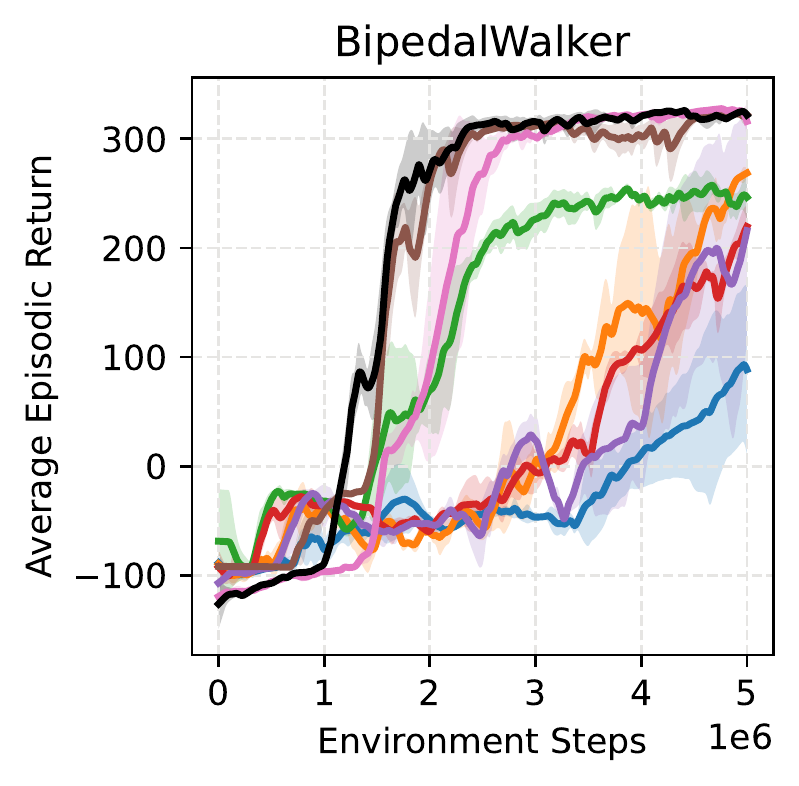}
    &\includegraphics[width=0.3\textwidth]{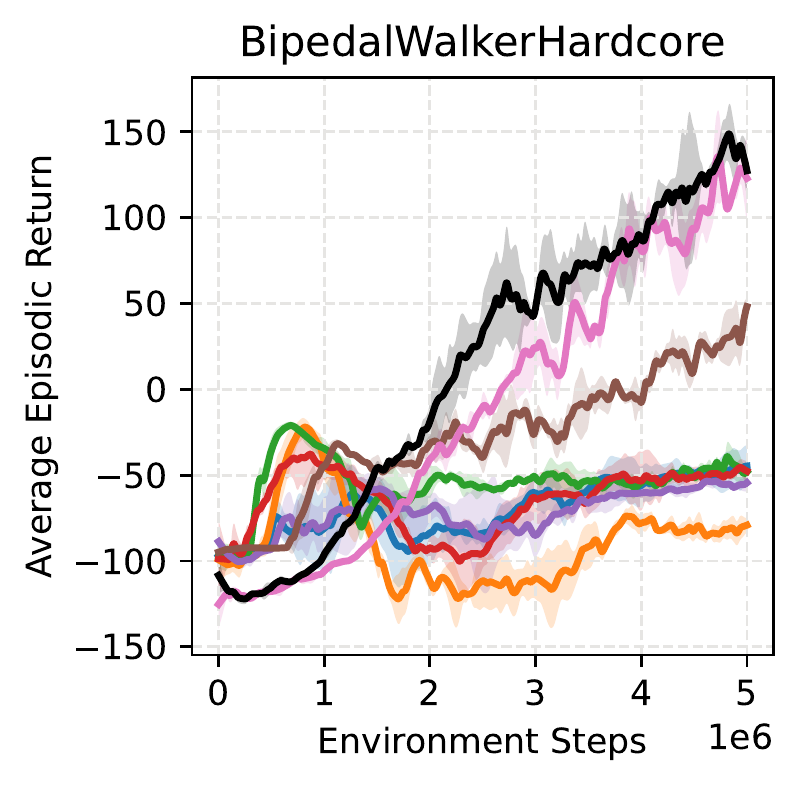}\\
    \includegraphics[width=0.3\textwidth]{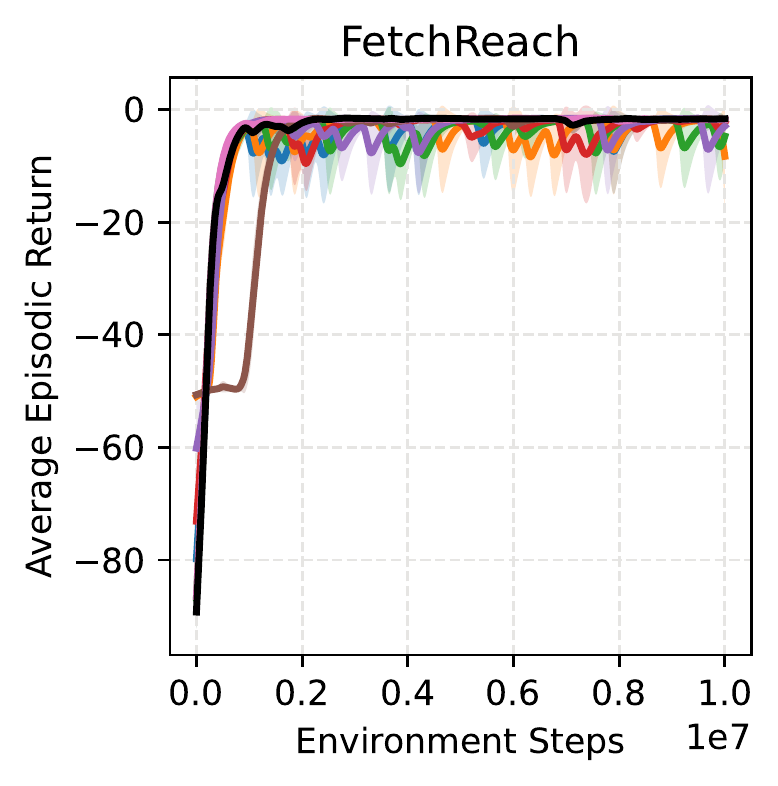}
    &\includegraphics[width=0.3\textwidth]{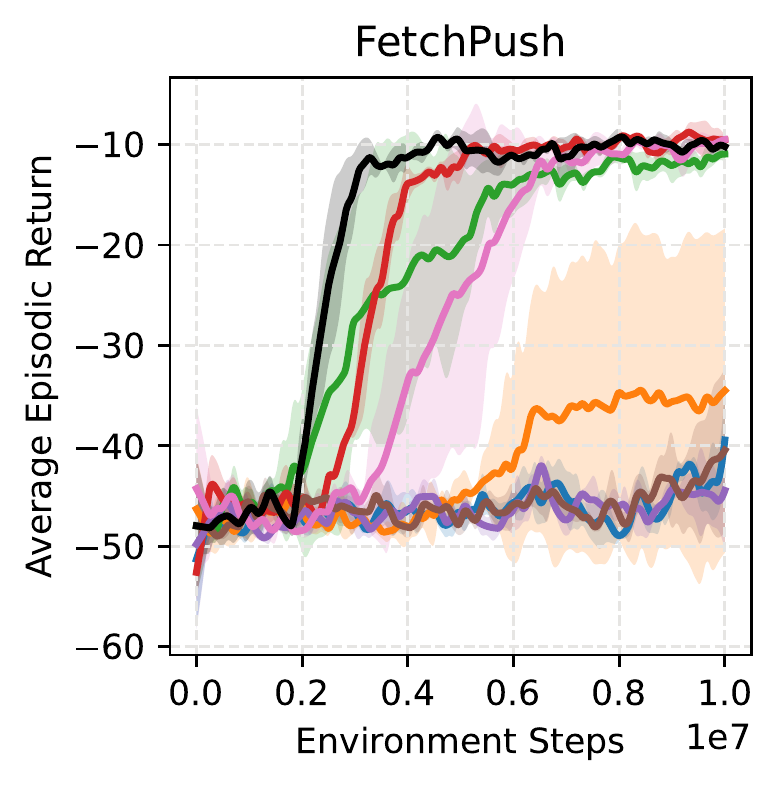}
    &\includegraphics[width=0.3\textwidth]{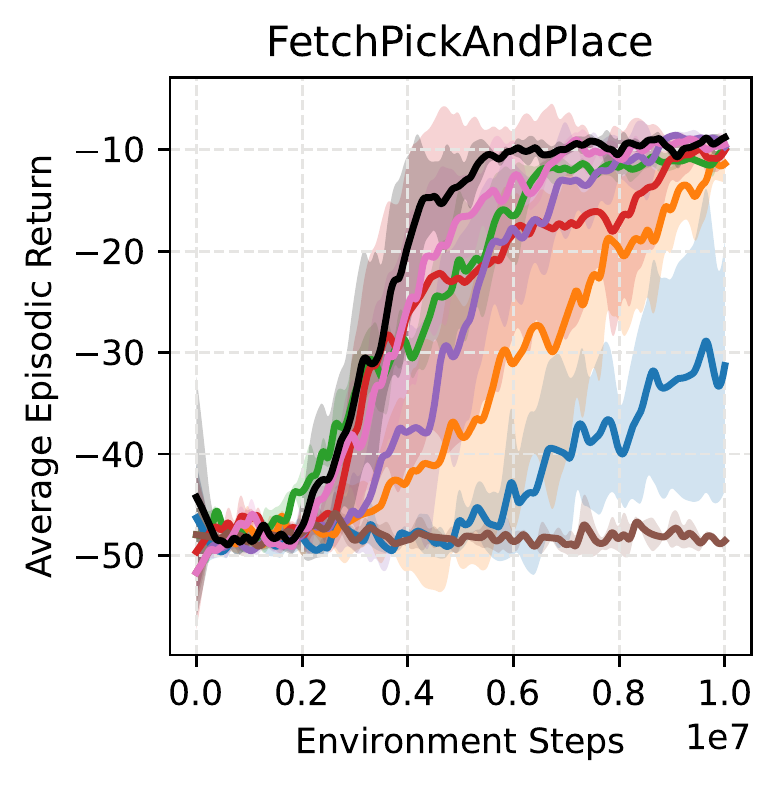}
    \\
    \includegraphics[width=0.3\textwidth]{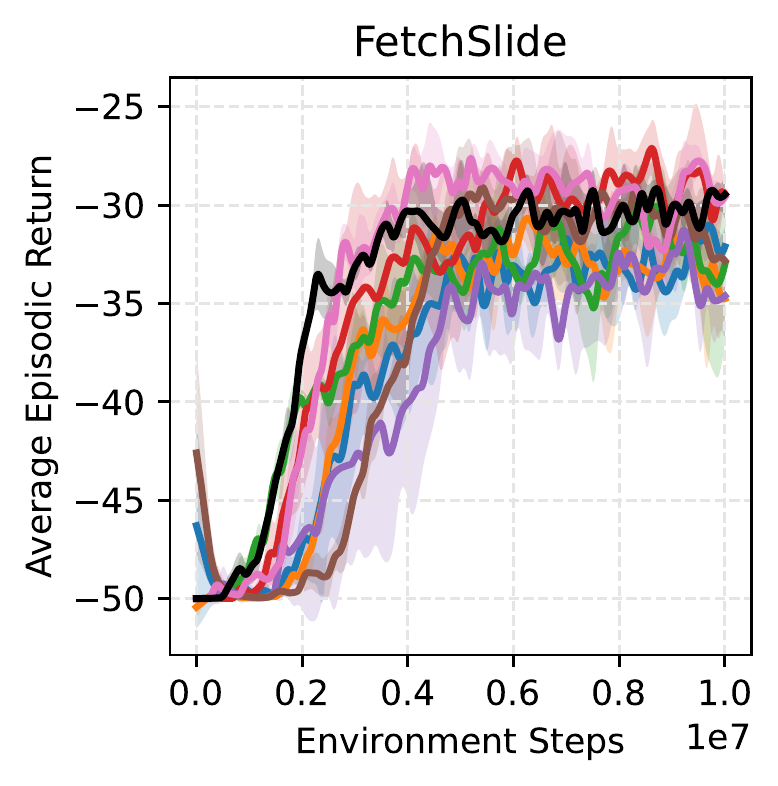}
    &\includegraphics[width=0.3\textwidth]{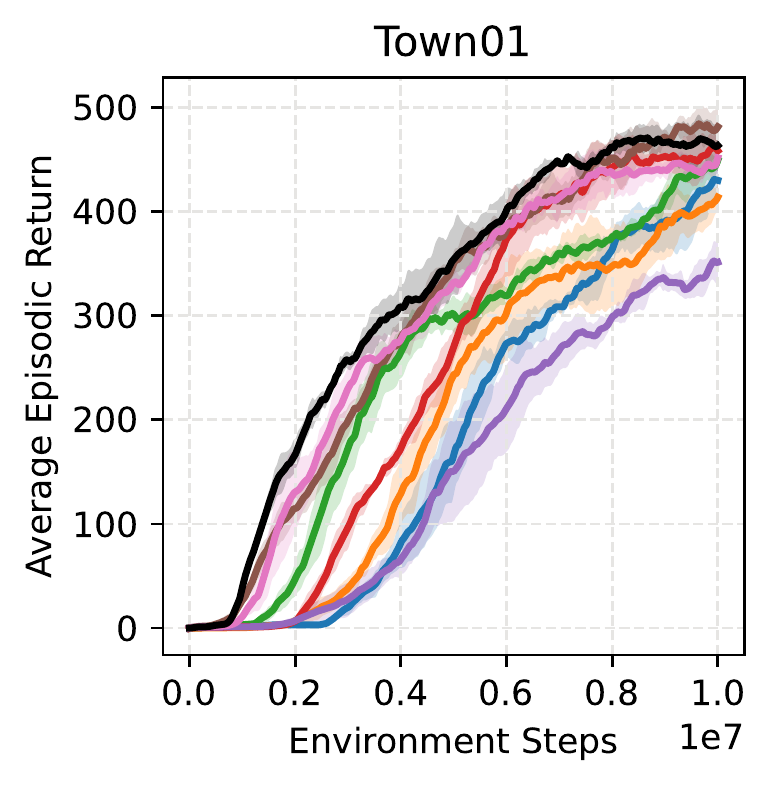}
    %&\dbox{\includegraphics[width=0.24\textwidth]{images/curves/Town01_Average_Episodic_Return_15M.pdf}}    
    &\includegraphics[viewport =-20 -50 190 130, clip, width=0.3\textwidth]{images/curves/legend_curves.pdf}
    \\
    \end{tabular}
    \caption{The unnormalized reward curves of the 14 tasks. 
    Each curve is averaged over 3 random seeds, and the shaded area around it represents the standard deviation. 
    %
    %The last subfigure in the dashed box runs to 15M steps to verify if SAC and SAC-Nrep have a chance of outperforming \name with 
    %more environment steps on \task{Town01}.
    }
\label{fig:unnormalized-curves}
\end{figure*}

For the 8 comparison methods in Section~\ref{sec:comparison_methods}, we list their n-score curves of all 5 task categories in Figure~\ref{fig:curves}, and their unnormalized score curves of all 14 tasks in Figure~\ref{fig:unnormalized-curves}.
The score curves are smoothed using exponential moving average to reduce noises.

%
\iffalse
We notice that the scores of SAC and SAC-Nrep are still climbing considerably and approaching that of \name at the end 
of training in \task{Town01}.
%
To rule out the possibility of them outperforming \name eventually, we train SAC, SAC-Nrep, and \name from scratch again to 
15M environment steps (50\% more steps).
%
The updated score curves are shown in the dashed box at the bottom.
%
We see that \name still performs better with more training time.
\fi
%
%Figure~\ref{fig:n10} contains the n-score curves for the experiment in Section~\ref{sec:off-policy-exp}, to show how off-policyness introduced by \name-Ntd's TD backup operator degrades its training performance as $N$ increases.
%
%Since \name uses an unbiased compare-through operator, its final n-score and n-AUC are both much better.

\iffalse
\begin{figure*}[!t]
    \centering
    \begin{tabular}{@{}c@{}c@{}c@{}c@{}}
        \includegraphics[width=0.24\textwidth]{images/curves/Terrain_Normalized_Score_N10.pdf}
        &\includegraphics[width=0.24\textwidth]{images/curves/Manipulation_Normalized_Score_N10.pdf}
        & \includegraphics[width=0.24\textwidth]{images/curves/Driving_Normalized_Score_N10.pdf}
        & \includegraphics[viewport =-10 -70 190 110, clip, width=0.25\textwidth]{images/curves/N10_legend.pdf} \\
    \end{tabular}
    \caption{The n-score curves of \name-Ntd and \name on three task categories \textbf{Terrain}, \textbf{Manipulation}, and \textbf{Driving}, with $N=10$.
    %
    \name-Ntd becomes worse (due to a biased Q operator) as $N$ increases while \name does not.
    %
    Each curve is an aggregation of a method's performance on the tasks within a task category, where 
    the method is run with 3 random seeds for each task.
    %
    Shaded areas represent standard deviation.}
\label{fig:n10}
\end{figure*}
\fi

\clearpage
\onecolumn
\section{Proof of multi-step policy evaluation convergence}

\label{app:policy_eval_proof}
We now prove that the compare-through Q operator $\mathcal{T}^{\pi}$ in Section~\ref{sec:policy_eval} is unbiased, namely,  $Q^{\pi}$ is a unique fixed point of $\mathcal{T}^{\pi}$ (policy evaluation convergence).
Assuming a tabular setting, value functions and policies are no longer parameterized and can be enumerated over all states and actions.

\subsection{Definition}
\begin{figure}[!t]
    \centering
    \includegraphics[width=0.9\textwidth]{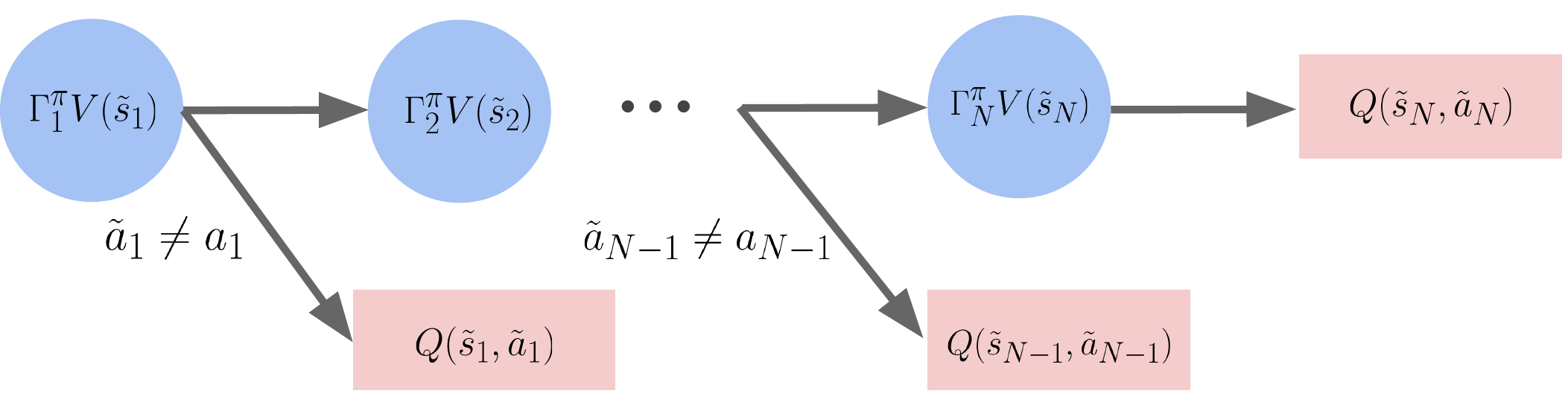}
  \caption{The stochastic binary tree defined by $\mathcal{T}^{\pi}$. 
  Circles are inner nodes and rectangles are leaves. 
  Whenever the sampled action $\tilde{a}_n$ is not equal to the rollout action $a_n$, a tree path terminates. 
  }
\label{fig:binary_tree}
\end{figure}

We first present a formal definition of $\mathcal{T}^{\pi}$.
Suppose that we consider $N$-step ($N\ge 1$) TD learning.
Each time we sample a historical trajectory $(s_0,a_0,s_1,r_{s_0,a_0,s_1}, \ldots, s_N, r_{s_{N-1},a_{N-1},s_N})$ of $N+1$ steps from the replay buffer to update $Q(s_0,a_0)$.
For convenience, we define a sequence of auxiliary operators $\Gamma^{\pi}_n$ for the $V$ values backup recursively as
\begin{equation}
\label{eq:v-operator}
    \begin{array}{rl}
        \Gamma^{\pi}_{N}V(\tilde{s}_N)=&\displaystyle\expect_{\pi(\tilde{a}_N|\tilde{s}_N)}Q(\tilde{s}_N,\tilde{a}_N),\\
        \Gamma^{\pi}_{n}V(\tilde{s}_n)=&\displaystyle\expect_{\pi(\tilde{a}_n|\tilde{s}_n)}\left[\mathbbm{1}_{\tilde{a}_n\ne a_n} \underbrace{Q(\tilde{s}_n,\tilde{a}_n)}_{\text{``stop''}}
        +\mathbbm{1}_{\tilde{a}_n=a_n}\underbrace{\displaystyle\expect_{\color{blue}\mathcal{P}(\tilde{s}_{n+1}|\tilde{s}_n,\tilde{a}_n)}\left[{\color{blue}r_{\tilde{s}_n,\tilde{a}_n,\tilde{s}_{n+1}}}+\gamma\Gamma_{n+1}^{\pi}V(\tilde{s}_{n+1})\right]}_{\text{``expand''}}\right],\\
        & \text{for}\ 1\le n \le N-1,\\
    \end{array}
\end{equation}
and based on which we define 
\begin{equation}
\label{eq:operator}
\mathcal{T}^{\pi}Q(s_0,a_0)=\expect_{\color{blue}\mathcal{P}(\tilde{s}_1|s_0,a_0)}\left[{\color{blue}r_{s_0,a_0,\tilde{s}_1}}+\gamma \Gamma_1^{\pi}V(\tilde{s}_1)\right]
\end{equation}
as the final operator to update $Q(s_0,a_0)$.
Intuitively, the above recursive transform defines a stochastic binary tree, where $\Gamma_n^{\pi}V(\cdot)$ are inner nodes and $Q(\cdot)$ are leaves (Figure~\ref{fig:binary_tree}).
The branching of an inner node (except the last $\Gamma_N^{\pi}V(\tilde{s}_N)$) depends on the indicator function $\mathbbm{1}_{\tilde{a}_n\ne a_n}$.
From the root $\Gamma_1^{\pi}V(\tilde{s}_1)$ to a leaf, the maximum path length is $N+1$ (when all $\tilde{a}_n=a_n$) and the minimum path length is 2 (when $\tilde{a}_1\ne a_1$).

To actually estimate $\mathcal{T}^{\pi}Q(s_0,a_0)$ during off-policy training without access to the environment for {\color{blue}$\mathcal{P}$} and {\color{blue}$r$}, we use the technique introduced in Section~\ref{sec:policy_eval} to sample a path from the root to a leaf on the binary tree, by re-using the historical trajectory as much as possible.
Specifically, we first set $\tilde{s}_1=s_1$ and $r_{s_0,a_0,\tilde{s}_1}=r_{s_0,a_0,s_1}$ as in the typical 1-step TD learning setting.
%on the historical state sequence $(s_1, s_2, \ldots, s_N)$, we first sample actions from the current $\pi$ as $(\tilde{a}_1, \tilde{a}_2, \ldots, \tilde{a}_N)$.
%
Starting from $n=1$, we sample $\tilde{a}_n\sim \pi(\cdot|s_n)$ and compare $\tilde{a}_n$ with $a_n$. 
If they are equal, we continue to set $\tilde{s}_{n+1}=s_{n+1}$ and $r_{\tilde{s}_n,\tilde{a}_n,\tilde{s}_{n+1}}=r_{s_n,a_n,s_{n+1}}$.
%p
We repeat this process until $\tilde{a}_n\ne a_n$. 
In a word, 
\begin{equation*}
    \mathcal{T}^{\pi}Q(s_0,a_0)\approx r_{s_0,a_0,s_1} + \gamma r_{s_1,a_1,s_2} + \ldots + \gamma^n Q(s_n,\tilde{a}_n),\  n=\min \left(\{n| \tilde{a}_n \ne a_n\} \cup \{N\}\right).
\end{equation*}

Usually for a continuous policy $\pi$, $\mathbbm{1}_{\tilde{a}_n\ne a_n}$ is 1 with a probability of 1 because two sampled actions are always unequal.
So $\mathcal{T}^{\pi}$ will stop expanding at $s_1$ and it seems no more than just a normal Bellman operator for 1-step TD learning.
However, if $\pi$ is specially structured and has a way of generating two identical actions in a continuous space, then it has the privilege of entering deeper tree branches for multi-step TD learning.
For example, \name is indeed able to generate $\tilde{a}_n$ identical to the rollout action $a_n$ if $\tilde{b}_k=b_k=0$, for all $1 \le k \le n$.
In this case, $\tilde{a}_n=a_n=a_0$.

Here we note that the above point estimate of $\mathcal{T}^{\pi}$ can also be written as $\mathcal{T}^{\pi}Q(s_0,a_0) \approx Q(s_0,a_0) + \Delta Q(s_0,a_0)$, where 
\begin{equation*}
\begin{array}{l}
\Delta Q(s_0,a_0)=\displaystyle\sum_{n=0}^{N-1}\gamma^n\left(\prod_{i=0}^n \mathbbm{1}_{a_i=\ta_i}\right)\Big[
r_{s_n,a_n,s_{n+1}} + \gamma Q(s_{n+1},\ta_{n+1}) - Q(s_n,\ta_n) \Big].\\
\end{array}
\end{equation*}
Thus it shares a very similar form with Retrace~\citep{Munos2016}, except the traces are now binary values defined by action comparison.

\subsection{Convergence proof}
Given any historical trajectory $\tau=(s_0,a_0,s_1,r_{s_0,a_0,s_1}, \ldots, s_N, r_{s_{N-1},a_{N-1},s_N})$ from an arbitrary behavior policy, 
we first verify that $Q^{\pi}$ is a fixed point of $\mathcal{T}^{\pi}$.
When $Q=Q^{\pi}$ in Eq.~\ref{eq:v-operator}, we have $\Gamma_N^{\pi}V(\tilde{s}_n)=\expect_{\pi(\tilde{a}_N|\tilde{s}_n)}Q^{\pi}(\tilde{s}_n,\tilde{a}_N)=V^{\pi}(\tilde{s}_n)$.
Now assuming $\Gamma_{n+1}^{\pi}V=V^{\pi}$, we have
\begin{equation*}
    \begin{array}{rl}
    \Gamma^{\pi}_{n}V(\tilde{s}_n)&=\displaystyle\expect_{\pi(\tilde{a}_n|\tilde{s}_n)}\left[\mathbbm{1}_{\tilde{a}_n\ne a_n}\cdot Q^{\pi}(\tilde{s}_n,\tilde{a}_n)
            +\mathbbm{1}_{\tilde{a}_n=a_n}\displaystyle\expect_{\mathcal{P}(\tilde{s}_{n+1}|\tilde{s}_n,\tilde{a}_n)}\left[r_{\tilde{s}_n,\tilde{a}_n,\tilde{s}_{n+1}}+\gamma V^{\pi}(\tilde{s}_{n+1})\right]\right]\\
    &=\displaystyle\expect_{\pi(\tilde{a}_n|\tilde{s}_n)}\left[\mathbbm{1}_{\tilde{a}_n\ne a_n} Q^{\pi}(\tilde{s}_n,\tilde{a}_n)
            +\mathbbm{1}_{\tilde{a}_n=a_n} Q^{\pi}(\tilde{s}_n,\tilde{a}_n)\right]\\
    &=\displaystyle\expect_{\pi(\tilde{a}_n|\tilde{s}_n)}Q^{\pi}(\tilde{s}_n,\tilde{a}_n)\\
    &=V^{\pi}(\tilde{s}_n).\\
    \end{array}
\end{equation*}        
Thus finally we have $\mathcal{T}^{\pi}Q^{\pi}(s_0,a_0)=\expect_{\mathcal{P}(\tilde{s}_1|s_0,a_0)}[r_{s_0,a_0,\tilde{s}_1}+\gamma V^{\pi}(\tilde{s}_1)]=Q^{\pi}(s_0,a_0)$ for any $(s_0,a_0)$.

To prove that $Q^{\pi}$ is the unique fixed point of $\mathcal{T}^{\pi}$, we verify that $\mathcal{T}^{\pi}$ is a contraction mapping on the infinity norm space of $Q$.
Suppose we have two different Q instantiations $Q$ and $Q'$, we would like prove that after applying $\mathcal{T}^{\pi}$ to them, $\Vert Q-Q' \Vert_{\infty}$ becomes strictly smaller than before.
Let $\Delta=\Vert Q-Q' \Vert_{\infty}$ be the current infinity norm, \textit{i.e.}, $\Delta=\max_{s,a}|Q(s,a)-Q'(s,a)|$.
Then we have 
\begin{equation*}
    \begin{array}{rl}
    \Vert \Gamma_N^{\pi}V-\Gamma^{\pi}_N V' \Vert_{\infty}&=\displaystyle\max_s|\Gamma_N^{\pi}V(s)-\Gamma_N^{\pi}V'(s)|\\
    &=\displaystyle\max_s\left|\expect_{\pi(\cdot|s)}(Q(s,\cdot)-Q'(s,\cdot))\right|\\
    &\le \displaystyle\max_s\expect_{\pi(\cdot|s)}\left|Q(s,\cdot)-Q'(s,\cdot)\right|\\
    &\le \displaystyle\max_s \expect_{\pi(\cdot|s)}\Delta\\
    &=\Delta,\\
    \end{array}
\end{equation*}
and for $1 \le n \le N-1$ recursively
\begin{equation*}
    \begin{array}{rl}
    \Vert \Gamma_n^{\pi}V-\Gamma^{\pi}_n V' \Vert_{\infty}&=\displaystyle\max_s|\Gamma_n^{\pi}V(s)-\Gamma_n^{\pi}V'(s)|\\
    &=\displaystyle\max_s \left|\expect_{\pi(a|s)}\left[\mathbbm{1}_{a\ne a_n}(Q(s,a)-Q'(s,a)) + \mathbbm{1}_{a=a_n}\gamma\expect_{\mathcal{P}(s'|s,a)}[\Gamma_{n+1}^{\pi}V(s')-\Gamma_{n+1}^{\pi}V'(s')]\right] \right|\\
    &\le \displaystyle\max_s \expect_{\pi(a|s)}\left[\mathbbm{1}_{a\ne a_n}\left|Q(s,a)-Q'(s,a)\right| + \mathbbm{1}_{a=a_n}\gamma\expect_{\mathcal{P}(s'|s,a)}\left|\Gamma_{n+1}^{\pi}V(s')-\Gamma_{n+1}^{\pi}V'(s')\right|\right] \\
    & \le \displaystyle\max_s \expect_{\pi(a|s)}\left[\mathbbm{1}_{a\ne a_n}\Delta + \mathbbm{1}_{a=a_n}\gamma\Delta \right]\\
    & \le \displaystyle\max_s \expect_{\pi(a|s)}\left[\mathbbm{1}_{a\ne a_n}\Delta + \mathbbm{1}_{a=a_n}\Delta \right]\\
    &= \displaystyle\max_s \expect_{\pi(a|s)}\Delta\\
    &=\Delta.\\
    \end{array}
\end{equation*}
Finally,
\begin{equation*}
    \begin{array}{rl}
        \Vert \mathcal{T}^{\pi}Q-\mathcal{T}^{\pi}Q' \Vert_{\infty}&=\displaystyle\max_{s,a}|\mathcal{T}^{\pi}Q(s,a)-\mathcal{T}^{\pi}Q'(s,a)|\\
        &=\displaystyle\max_{s,a}\left|\gamma\expect_{\mathcal{P}(\cdot|s,a)}[\Gamma_1^{\pi}V(\cdot)-\Gamma_1^{\pi}V'(\cdot)] \right|\\
        &\le \gamma \displaystyle\max_{s,a}\expect_{\mathcal{P}(\cdot|s,a)}|\Gamma_1^{\pi}V(\cdot)-\Gamma_1^{\pi}V'(\cdot)|\\
        &\le \gamma \displaystyle\max_{s,a}\expect_{\mathcal{P}(\cdot|s,a)}\Delta\\
        &=\gamma \Delta\\
        &=\gamma \Vert Q-Q'\Vert_{\infty}.\\
    \end{array}
\end{equation*}
Since the discount factor $0<\gamma<1$, we have proved that $\mathcal{T}^{\pi}$ is a contraction mapping.

Importantly, this contraction holds for any historical trajectory $\tau$, even though this trajectory differs each time for an operator transform.
Namely, every operator transform step will bring $Q$ and $Q'$ closer, \emph{regardless of the actual value of the historical trajectory referred to}.
Combining this with $Q^{\pi}$ being a fixed point of $\mathcal{T}^{\pi}$, we have shown that any $Q$ will converge to $Q^{\pi}$ if we repeatedly apply $\mathcal{T}^{\pi}$ to it.

%%%%%%%%%%%%%%%%%%%%%%%%%%%%%%%%%%%%%%%%%%%%%%%%%%%%%%%%%%%%%%%%%%%%%%%%%%%%%%%
%%%%%%%%%%%%%%%%%%%%%%%%%%%%%%%%%%%%%%%%%%%%%%%%%%%%%%%%%%%%%%%%%%%%%%%%%%%%%%%

\end{document}